%% file: main_aistats.tex
\documentclass[twoside]{article}
\usepackage[english]{babel}
\usepackage[T1]{fontenc}
\usepackage{lmodern}

\usepackage[accepted]{aistats2022}
\usepackage[round]{natbib}
\input{math_commands.tex}

\usepackage{multirow}
\usepackage{graphicx}
\usepackage{subcaption}
\usepackage{amssymb}
\usepackage{mathtools}
\usepackage{hyperref}       % hyperlinks
\usepackage{makecell}
\usepackage{xspace}

\usepackage{caption}
\usepackage{colortbl}

\newcommand{\xhdr}[1]{\vspace{0em}\noindent{{\bf #1.}}}
\newcommand{\hide}[1]{}
\newcommand{\na}{~~~~~~~N/A~~~~~~~~}
\newcommand{\naapp}{N/A}
\newcommand{\ie}{\textit{i.e., }}
\newcommand{\eg}{\textit{e.g., }}
\newcommand{\std}[1]{\scriptsize{$\pm$#1}}
\newcommand{\deriv}[2]{\frac{\partial{#1}}{\partial{#2}}}

\newcolumntype{a}{>{\columncolor{Gray}}c}
\newcolumntype{b}{>{\columncolor{white}}c}

\newcommand{\Expmethods}{Explanation methods\xspace}
\newcommand{\expmethod}{explanation method\xspace}
\newcommand{\expmethods}{explanation methods\xspace}

\newcommand{\methods}{methods\xspace}
\newcommand{\reliable}{reliable\xspace}
\newcommand{\reliability}{reliability\xspace}
\newcommand{\localreg}{local region\xspace}
\newcommand\ccg[1]{\cellcolor{gray!35}{#1}} % for cells in second column % and gray colored rows
\newcommand\ccr[1]{\cellcolor{red!25}{#1}} % for cells in second column % and gray colored rows
\newcommand\ccb[1]{\cellcolor{blue!25}{#1}} % for cells in second column % and gray colored rows
\newcommand\ccy[1]{\cellcolor{yellow!35}{#1}} % for cells in second column % and gray colored rows
\usepackage{booktabs} % To thicken table lines
\usepackage{color, colortbl}
\definecolor{LightCyan}{rgb}{0.88,1,1}
\usepackage[dvipsnames,table,xcdraw]{xcolor}

\hypersetup{
colorlinks = true,
linkcolor = Blue,
anchorcolor = blue,
citecolor = Blue,
filecolor = cyan,
menucolor = ForestGreen,
runcolor = cyan,
urlcolor = ForestGreen}

\begin{document}
% If your paper is accepted and the title of your paper is very long,
% the style will print as headings an error message. Use the following
% command to supply a shorter title of your paper so that it can be
% used as headings.
%
%\runningtitle{I use this title instead because the last one was very long}

% If your paper is accepted and the number of authors is large, the
% style will print as headings an error message. Use the following
% command to supply a shorter version of the authors names so that
% they can be used as headings (for example, use only the surnames)
%
%\runningauthor{Surname 1, Surname 2, Surname 3, ...., Surname n}

\twocolumn[

\aistatstitle{Probing GNN Explainers: A Rigorous Theoretical and Empirical Analysis of GNN Explanation Methods}

\aistatsauthor{Chirag Agarwal\And Marinka Zitnik*\And Himabindu Lakkaraju*}

\aistatsaddress{Harvard University \And  Harvard University \And Harvard University} ]

\begin{abstract}
\input{000abstract}

\end{abstract}

\section{INTRODUCTION}
\label{sec:intro}
\input{010intro}
\section{RELATED WORK}
\label{sec:related}
\input{020related}

\section{THEORETICAL ANALYSIS OF GNN EXPLANATION METHODS}
\label{sec:properties}
\input{030problem}

\section{EMPIRICAL  ANALYSIS OF GNN EXPLANATION METHODS}
\label{sec:exp}
\input{050experiment}

\section{CONCLUSIONS}
\label{sec:conclusion}
\input{060conclusion}

\subsubsection*{Acknowledgements}
\input{070acknowledgement}

% \clearpage
\bibliographystyle{plainnat}
\bibliography{references.bib}

\input{supp_aistats}

\end{document}

%% file: math_commands.tex
\usepackage{amsmath,amsfonts,bm}

\def\eqref#1{equation~\ref{#1}}
\def\1{\bm{1}}

\DeclareMathAlphabet{\mathsfit}{\encodingdefault}{\sfdefault}{m}{sl}
\SetMathAlphabet{\mathsfit}{bold}{\encodingdefault}{\sfdefault}{bx}{n}

\DeclareMathOperator*{\argmax}{arg\,max}

%% file: 000abstract.tex
% As Graph Neural Networks (GNNs) are increasingly employed in real-world applications, it becomes critical to ensure that the stakeholders understand the rationale behind their predictions. While several GNN \expmethods have been recently proposed to explain their decisions, little to no work exists on providing a common framework for analyzing their reliability.
%As Graph Neural Networks (GNNs) are increasingly being employed in real-world applications, several GNN \expmethods have been proposed in recent literature to ensure that stakeholders understand the rationale behind GNN predictions. However, the lack of a common theoretical framework for understanding how well these \expmethods perform limits their use in practice.
\looseness=-1
As Graph Neural Networks (GNNs) are increasingly being employed in critical real-world applications, several methods have been proposed in recent literature to explain the predictions of these models. 
%it becomes critical to ensure that the relevant stakeholders understand the rationale behind the predictions of these models. 
%While several GNN \expmethods have been proposed in recent literature, 
However, there has been little to no work on systematically analyzing the reliability of these methods. Here, we introduce the first-ever theoretical analysis of the reliability of state-of-the-art GNN \expmethods. More specifically, we theoretically analyze the behavior of various state-of-the-art GNN \expmethods with respect to several desirable properties (e.g., faithfulness, stability, and fairness preservation) and establish upper bounds on the violation of these properties. We also empirically validate our theoretical results using extensive experimentation with nine real-world graph datasets. Our empirical results further shed light on several interesting insights about the behavior of state-of-the-art GNN \expmethods. %:

%% file: 010intro.tex
% \chirag{1. What is the problem?\\
% 2. Why is it interesting and important?\\
% 3. Why is it hard? (E.g., why do naive approaches fail?)\\
% 4. Why hasn't it been solved before? (Or, what's wrong with previous proposed solutions? How does mine differ?)\\
% 5. What are the key components of my approach and results? Also include any specific limitations.\\}
%%% DONT DELETE BELOW - HIMA%%
%, and develop strategies for further model refinement. 
%While the effectiveness of GNNs can be attributed to their ability to learn node representations by recursively incorporating information from neighboring nodes via complex non-linear transformations~\cite{}, this complexity makes it rather challenging to understand or interpret their predictions~\cite{}. 
%%%%%%%%%%%%%%%%%%%%%%%%%%%%%%%%%%%%%%%%%%%%
\looseness=-1
Graph Neural Networks (GNNs) have emerged as powerful tools for effectively representing graph structured data, such as social, information, chemical, and biological networks. As these models are increasingly being employed in critical applications (\eg drug repurposing~\citep{zitnik2018modeling}, crime forecasting~\citep{jin2020addressing}), it becomes essential to ensure that the relevant stakeholders can understand and trust their functionality~\citep{ying2019gnnexplainer}. Only if the stakeholders have a clear understanding of the behavior of these models, they can evaluate when and how much to rely on these models, and detect potential biases or errors in them. To this end, several approaches have been proposed in recent literature to explain the predictions of GNNs~\citep{baldassarre2019explainability,faber2020contrastive,huang2020graphlime,lucic2021cf,luo2020parameterized,pope2019explainability,schlichtkrull2020interpreting,vu2020pgm,ying2019gnnexplainer}. 
Based on the techniques they employ, these approaches can be broadly characterized into perturbation-based~\citep{luo2020parameterized,schlichtkrull2020interpreting,ying2019gnnexplainer}, gradient-based~\citep{Simonyan14a,sundararajan2017axiomatic}, and surrogate-based~\citep{huang2020graphlime,vu2020pgm} methods~\citep{yuan2020explainability}.
\looseness=-1
While several classes of GNN \expmethods have been proposed in recent literature, there is little to no understanding as to which of these approaches are more effective than the others and/or if some of these approaches are better suited for certain kinds of real-world applications.
This lack of understanding not only limits the applicability of GNN \expmethods in practice but also hinders the progress of research in graph XAI. More specifically, without such a deeper understanding, stakeholders in real-world settings may not be able to determine which approaches to employ, and researchers in the field may expend a lot of resources studying ineffective solutions.  %In addition, this could potentially hinder the progress of graph XAI as researchers in the field may expend a lot of resources on ineffective solutions.  %This lack of understanding not only limits the applicability of GNN \expmethods in practice but also hampers the progress of research in graph XAI. 
This lack of understanding mainly stems from the fact that there is very little work on systematically analyzing the reliability of various classes of state-of-the-art GNN \expmethods. 

%While several \expmethods have been previously proposed, there are no works on theoretically analyzing the reliability of GNN explanations and benchmarking the performance of state-of-the-art GNN \expmethods. Existing explanation evaluation strategies are heuristically designed for specific techniques~\cite{sanchez2020evaluating,yuan2020explainability,faber2021comparing} where qualitative evaluations are subjective and time-consuming, and quantitative measures are accuracy based that do not consider other model properties. These limitations emphasize the need for a generic and comprehensive theoretical framework that can evaluate the reliability of all GNN \expmethods.
% \chirag{to put things into common ground; cite benchmarks effort in vision and language}
% \chirag{mention about the previous attempts}
% \looseness=-1
Few recent works have focused on empirically evaluating GNN \expmethods~\citep{yuan2020explainability}. For instance,~\cite{sanchez2020evaluating} focused on evaluating methods that output attributions (i.e., highlight input features influential to model predictions). They outlined various properties a GNN explanation method should satisfy --- e.g., accuracy, faithfulness, and stability. Using these metrics, they empirically evaluated only gradient-based GNN \expmethods (e.g., SmoothGrad, GradCAM). More recently,~\cite{faber2021comparing} highlighted the pitfalls of using arbitrary ground truth patterns in the data when evaluating GNN explanations as there may be a mismatch between these patterns and the GNN itself. They introduced three benchmark datasets to alleviate the aforementioned pitfalls. %, and argued against evaluating GNN explanations by comparing them with arbitrary ground truth patterns in the data because there may often be a mismatch between these patterns and the GNN itself. 
While these works make initial attempts at empirically evaluating GNN explanations, the metrics outlined are neither generalizable nor exhaustive. For example, most of the proposed metrics rely on the availability of ground truth explanations, thus severely limiting the kinds of datasets that can be used during evaluation. Those metrics do not account for fairness properties of explanations which are critical to applications like crime forecasting~\citep{aivodji2019fairwashing}. Further, these works do not focus on theoretically analyzing the reliability or effectiveness of state-of-the-art GNN explanation methods. 

\xhdr{Present work} In this work, we introduce the first ever theoretical analysis of the reliability of state-of-the-art GNN \expmethods. More specifically, we analyze the behavior of various GNN \expmethods w.r.t. several key desirable properties such as faithfulness (i.e., faithfully mimicking the predictions of the underlying model), stability (to small changes in the input), and \emph{fairness preservation} (i.e., preserving the (un)fairness of the underlying model). While we leverage existing notions of faithfulness and stability outlined in prior literature, we introduce the notion of fairness preservation for GNN explanations for the first time in this work. %This property becomes particularly critical 
As GNNs are increasingly being deployed in domains such as criminal justice and  financial lending,
%. In such settings
it becomes critical to ensure that the GNN explanations preserve the fairness properties of the underlying GNN models. 
For instance, if a GNN model is biased against a protected group (e.g., violates the notion of statistical parity), then the corresponding explanation should reflect that. 

% \looseness=-1
We formalize the above properties such that they do not rely on the availability of ground truth explanations and are therefore more generalizable to different domains and datasets. We then leverage these formalisms to establish theoretical upper bounds on the violation of the aforementioned properties for various state-of-the-art GNN \expmethods~(Sec.~\ref{sec:properties}). To carry out our theoretical analysis, we leverage the notion of Lipschitz continuity (Theorems 2-7) and employ concepts from information theory and probability theory such as data processing inequalities (Theorems 1-8) and total variation distance (Theorem 8).
%Our theoretical analysis establishes tight upper bounds on the violation of the aforementioned properties for several state-of-the-art GNN \expmethods~(Sec.~\ref{sec:properties}; Theorems 1-8). 
We also perform an extensive empirical evaluation with GNN \expmethods on nine real-world datasets and multiple learning tasks (i.e., node classification, link prediction and graph classification). Our empirical results validate our theoretical bounds and also unearth some critical insights about the behavior of state-of-the-art GNN \expmethods, %where on average across all datasets, we observe that: 
% \hima{NEED TO FINALIZE INSIGHTS}
% on average, accounting for graph structure does not really add significant value in terms of our key properties
% 2. We are seeing that the random baselines (particularly random edge baseline) performs on par or sometimes even better than the other SOTA methods. For instance, random edge baseline turns out to be one of the best performing algorithms when it comes to faithfulness.
% 3. GraphMask algorithms outperforms all other algos and baselines when it comes to preserving counterfactual fairness
1) Gradient-based methods exhibit poor performance \textit{w.r.t.}~faithfulness and stability, but perform quite well \textit{w.r.t.}~counterfactual fairness; perturbation-based methods, on the other hand, exhibit the best performance at preserving group fairness (Fig.~\ref{app:fig_category_compare}), 2) Random baselines (particularly random edge baseline) perform either on par or sometimes even better than state-of-the-art GNN \expmethods (Fig.~\ref{app:fig_method_compare}) \textit{w.r.t.}~faithfulness and group fairness preservation, and 3) on average across all datasets and all our key properties, explanations comprising of graph structures perform slightly better than those that comprise of node features (Fig.~\ref{app:fig_node_edge_compare}).

%% file: 020related.tex
% \chirag{push the related works to main tex but only include benchmarking and graphxai; and end it with the main contribution}
% \hima{CHIRAG: change all cite to citep in this section and beyond!!}
This paper builds upon a wealth of previous research at the intersection of \expmethods, graph neural networks, and systematic evaluation of explanations. 

\looseness=-1
\xhdr{Explanation methods for GNNs}
GNNs specify non-linear transformation functions that map graph structures (nodes, edges or entire graphs) into compact vector embeddings~\citep{li2021representation}. 
% This is typically achieved through a process that iteratively propagates neural messages along edges of a graph and aggregates them at nodes~\citep{gilmer2017neural}. 
A variety of GNN architectures have been designed ~\citep{pareja2020evolvegcn,yun2019graph,zitnik2018modeling}, and recent research has focused on developing methods to explain GNN predictions~\citep{baldassarre2019explainability,pope2019explainability,ying2019gnnexplainer,huang2020graphlime,luo2020parameterized,vu2020pgm,schlichtkrull2020interpreting,chen2021molecule,han2010explainable}. 
Early methods developed graph analogs of \emph{gradient-based methods} from computer vision literature, including gradient heatmaps~\citep{Simonyan14a} and integrated gradients~\citep{sundararajan2017axiomatic}. 
% \hima{why are these called perturbation methods then? there is no connection between perturbation methods and them finding small subgraphs; even other methods find small subgraphs}
Recently, \emph{perturbation-based methods} \citep{ying2019gnnexplainer,luo2020parameterized,schlichtkrull2020interpreting} explain GNN predictions by observing the change in model predictions \textit{w.r.t.} different input perturbations to study node and edge importance.
% finding small subgraphs that are most influential for the prediction \textit{w.r.t.} different perturbations of the input graph. 
% To this end, GraphMASK~\citep{schlichtkrull2020interpreting} identifies edges in each GNN layer that can be dropped without affecting the final prediction. Similarly, GNNExplainer \citep{ying2019gnnexplainer} learns real-valued edge and node attribute masks that select influential subgraphs. 
Finally, \emph{surrogate-based methods}~\citep{huang2020graphlime,vu2020pgm} fit an interpretable model to local neighborhoods of the node such that the model captures the GNN's behavior in the local vicinity of target nodes. See Appendix~\ref{app:sec_exp_methods} for a detailed overview of the \expmethods.

\looseness=-1
%\chirag{Make it generic from and change it to something like evaluation of GNN \expmethods\dots general overview of the efforts.. give one example and refer to section 3 for detail\dots humble the tone while discussing related works and introduction \dots}
%\chirag{cite benchmarks from other modalities in the related works \dots cite some theoretical papers from generic xai and use that to motivate our work}
%\chirag{TALK ABOUT EXISTING WORKS FROM GENERIC XAI and THE GENERALIZATION AND EXHAUSTIVENESS OF GRAPHXAI evaluations}
%\marinka{I'm finalizing this.}
\looseness=-1
\xhdr{Evaluation of GNN \expmethods} 
%
% Recent work has shed light on the downsides of post-hoc explanation techniques.
% For instance, \citep{rudin2019stop} argued that post-hoc explanations unreliable as they are not necessarily faithful to the underlying models. % and can present spurious correlations.
% Further, empirical studies of explanations~\citep{adebayo2018sanity,slack2019can,lakkaraju2020how,rudin2019stop,dombrowski2019explanations, ghorbani2019interpretation} revealed several vulnerabilities of post-hoc explanations, including their fragility by demonstrating that explanations can change drastically when the input is perturbed~\citep{alvarez2018robustness}. 
% To this end, \citep{levine2019certifiably,pmlr-v119-chalasani20a,garreau2020looking} have made first steps towards analysing robustness and other properties of some post-hoc explanation techniques.
% Few works have analysed robustness and other properties of post-hoc explanation techniques \citep{levine2019certifiably,pmlr-v119-chalasani20a,garreau2020looking}.
%A Learning Theoretic Perspective on Local Explainability\citep{li2021learning} 
%an empirical study of deep neural network explanation\citep{jeyakumar2020can}
%
Empirical studies of deep neural network explanations evaluated methods and designed benchmarks for image, text, audio, time series, and sensory datasets~\citep{jeyakumar2020can,liu2021synthetic,arya2019one,fauvel2020performance,deyoung2019eraser,amparore2021trust}. 
% \hima{Chirag, please take a look at intro and use some of that language here about what is wrong with existing evaluation metrics. Don't copy paste but try to reword}
However, due to the relative infancy of GNN explainability as a field, rigorous analyses of GNN \expmethods are very limited. While few works such as~\cite{sanchez2020evaluating,yuan2020explainability,faber2021comparing} make initial attempts at empirically evaluating GNN \expmethods, the metrics outlined are neither generalizable nor exhaustive. For instance, most of the proposed metrics rely on the availability of ground truth explanations, thus severely limiting the kinds of datasets that can be used for evaluation. The proposed metrics also do not account for fairness properties of explanations which are critical to applications such as crime forecasting~\citep{aivodji2019fairwashing}. Furthermore, despite few preliminary attempts at theoretical analysis of generic XAI techniques such as LIME and SmoothGrad~\citep{chen2018learning,agarwal2021toward,garreau2020explaining}, no theoretical analysis of GNN \expmethods has been attempted.

%% file: 030problem.tex
% \hima{This section should MOTIVATE as well as formally define all our key properties}
%\hima{ all your bounds are on the negative aspect i.e., unfairness, instability but your theorem statements say fairness and stability. This needs to be fixed!}
\looseness=-1
In this section, we theoretically analyze the reliability of various state-of-the-art GNN \expmethods. We first outline and formalize the key desirable properties that capture the reliability of a given GNN explanation, namely, \emph{faithfulness}, \emph{stability}, and \emph{fairness preservation}.  More specifically, we posit that a reliable GNN explanation should  \emph{faithfully} mimic the predictions of the underlying GNN model, preserve critical model characteristics such as \emph{(un)fairness} of the underlying model, and exhibit \emph{stability} to small input perturbations. While we adopt existing notions of faithfulness and stability outlined in prior literature, we introduce and define the notion of fairness preservation for GNN explanations for the first time in this work. We then leverage these formalisms to derive upper bounds on the violation of the aforementioned properties for several GNN \expmethods.
\xhdr{Notation: Graphs and GNNs}
Let $\mathcal{G}{=}(\mathcal{V}, \mathcal{E}, \mathbf{X})$ denote an undirected and unweighted graph comprising of a set of nodes $\mathcal{V}$, a set of edges $\mathcal{E}$, and a set of node feature vectors $\mathbf{X}{=}\{\mathbf{x}_{1}, \dots, \mathbf{x}_{N}\}$ corresponding to nodes in $\mathcal{V}$, where $\mathbf{x}_{u}{\in}\mathbb{R}^{M}$.
Let $N{=}|\mathcal{V}|$ denote the number of nodes in the graph and $\mathbf{A}\in \mathbb{R}^{N \times N}$ be the adjacency
matrix, where element $\mathbf{A}_{ij}{=}1$ if nodes $i$ and $j$ are connected by some edge in $\mathcal{E}$, and $\mathbf{A}_{ij}{=}0$ otherwise.
We use $\mathcal{N}_{u}$ to denote the 1-hop neighbors of node $u$ excluding itself.
Without loss of generality, we focus on the node classification task and use $f$ to denote a GNN model trained to predict node labels. Note that the metrics we consider in this work can also be applied to other graph machine learning tasks (e.g., link prediction and graph prediction) as we demonstrate in Sec.~\ref{sec:exp}. The GNN model's prediction for node $u$ is given by $\mathbf{\hat{y}}_{u}{=}f(\mathcal{G}_{u})$, where $\mathcal{G}_{u}$ is the computation graph for node $u$ and $\mathbf{\hat{y}}_{u}{\in}[0, 1]^C$.
% subgraph containing node $u$ and its 1-hop neighbors and $\mathbf{\hat{y}}_{u}{\in}[0, 1]^C$ 
% \marinka{See my comment in Slack. This is not true. $f$ never operates on 1-hop neighborhood of $u$}. 
The associated adjacency matrix and node attributes for $\mathcal{G}_{u}$ are denoted by $\mathbf{A}_{u}{\in}\{0, 1\}^{N\times N}$ (an element in this matrix has the value $1$ if it corresponds to an edge connecting node $u$ and some node in $\mathcal{N}_u$; otherwise it is set to $0$) and $\mathbf{X}_{u}{=}\{\mathbf{x}_{i} | i \in \{u, \mathcal{N}_{u}\}\}$ respectively.
Also, $\hat{y}_{u}{=}\argmax_{c} \mathbf{\hat{y}}_{u}$ is the predicted label, where $\hat{y}_{u} \in \{0, 1, \dots, C-1\}$ and $C$ is the number of classes.

\looseness=-1
\xhdr{Notation:~GNN explanations}
In this work, we focus on instance level explanations which are the most popular class of explanations studied in GNN literature. Instance level explanations, as the name suggests, explain model predictions associated with individual entities (e.g., nodes) in the graph (and do not capture the global behavior of the entire GNN model). %   It produces explanations in terms of nodes, edges, and/or node features that support the model's predictions.
% It produces explanations as one or several training instances (\ie nodes, edges, and/or node attributes) that support the model's predictions.  %where an \expmethod generates an explanation for a node $u$'s prediction returned by the GNN model $f$.
% Let $G_{u}$ denote the subgraph associated with node $u$ which fully determines information (\ie node features and edge connectivity) needed by $f$ to predict $u$'s label.
%
For instance, the explanation $\mathbf{E}_{u}$ corresponding to node $u$ comprises of a subset of node features and a subset of edges that influence the prediction of node $u$, i.e., $\hat{y}_u$. %E(f, \mathcal{G}_{u})$
% For explaining the predictions for each node $u$, the explanation algorithm generates $\mathbf{E}_{u}$ as its corresponding explanation. 
In particular, the explanation $\mathbf{E}_{u}$ consists of a discrete node feature mask $\mathbf{r}_{u}{\in}\{0, 1\}^{M}$ and/or a discrete edge mask $\mathbf{R}_{u}{\in}\{0, 1\}^{N{\times}N}$. An element in $\mathbf{r}_{u}$ or $\mathbf{R}_{u}$ takes the value $1$ if the corresponding node feature or edge (respectively) influences the prediction (and is therefore important), and is set to $0$ otherwise.
%as discrete masks identifying important node attributes for explaining $\hat{y}_{u}$. 
We use $t(\mathbf{E}_{u}, \mathcal{G}_{u})$ to denote a masking function which \emph{zeroes out} all those node features and incident edges of $u$ that are not deemed as important by the explanation $\mathbf{E}_{u}$, i.e., $t(\mathbf{E}_{u}, \mathcal{G}_{u})$ updates the node attributes and adjacency matrix corresponding to $\mathcal{G}_{u}$ as follows: $\mathbf{X}_{u}{=} \{\mathbf{x}_{u} \circ \mathbf{r}_{u}\} \cup \{ \mathbf{x}_i  | i \in \mathcal{N}_{u}\}$ and $\mathbf{A}_{u}{=}\mathbf{R}_{u} \circ \mathbf{A}_{u}$. Finally, we use $\hat{y}^{E}_{u}{=}f(t(\mathbf{E}_{u}, \mathcal{G}_{u}))$ to denote the prediction label output by $f$ for node $u$ when the masked subgraph is provided as the input.

\subsection{Theoretical Guarantees on Faithfulness}
\label{sec:def_faith}
\looseness=-1
% \hima{editing here}
A reliable explanation should highlight the key node/edge features that the underlying GNN leverages to make a prediction. In a GNN’s neural message-passing scheme, every node has access to a local view of the graph created by propagating neural messages (embeddings) along edges in the node’s local neighborhood~\citep{vignac2020building}.
% Thus, an explanation metric should leverage this local neighborhood to evaluate the explanation for the node’s prediction. 
% However, existing metrics \citep{sanchez2020evaluating,yuan2020explainability} fail to exploit this key aspect of a graph.
% \ie they do not leverage local neighborhoods around nodes when defining evaluation metrics for local explanations.
% addresses all the above challenges and
Following \cite{pope2019explainability} and \cite{yuan2020explainability}, we evaluate the faithfulness of the explanation corresponding to any given node $u$ by leveraging both its features as well as the incident edges.
% We consider a set of small perturbations to the attribute values and incident edges (\ie local 1-hop neighborhood) of a given node and then compute the difference in the model predictions across all the perturbations.  (\ie local 1-hop neighborhood)
Formally, we say that an explanation $\mathbf{E}_{u}$ corresponding to a node $u$ is faithful if it accurately captures the behavior of the underlying model $f$ in the \localreg around $u$. To operationalize this, we first generate the \localreg around $u$ by constructing a set $\mathcal{K}$ of nodes comprising of node $u$ and its perturbations. These perturbations are generated by making infinitesimally small changes to its node features and/or rewiring the edges incident on node $u$ with a small probability. Next, we obtain the model predictions as well as explanation $\mathbf{E}_{u}$'s predictions for node $u$ and all its perturbations. Note that the \emph{explanation's predictions} can be obtained by first using the mapping function $t$, which takes an explanation mask and applies it to any given node and its subgraph (See Notation on GNN explanations above) to generate a new masked subgraph, which is then passed as input to the model $f$ to obtain a prediction. The average difference between the model and explanation predictions for all the nodes in $\mathcal{K}$ will provide us with an estimate of how \emph{unfaithful} the explanation $\mathbf{E}_{u}$ is. The smaller this estimate, the more faithful the explanation $\mathbf{E}_{u}$.

\xhdr{Definition 1 (Faithfulness)}
\textit{Given a set $\mathcal{K}$ comprising of a node $u$ and its perturbations, an explanation $\mathbf{E}_{u}$ corresponding to node $u$ is said to be faithful if:}
\begin{equation}
    % \frac{1}{\mathcal{K}+1}
    \frac{1}{|\mathcal{K}|}\sum_{u'\in\mathcal{K}}||f(\mathcal{G}_{u'}) - f(t(\mathbf{E}_{u}, \mathcal{G}_{u'}))||_{2} \leq \delta,
    \label{eq:faith}
\end{equation}
\vskip -0.15in
where $\mathcal{G}_{u'}$ denotes a subgraph of node $u'$ and $\delta$ is an infinitesimally small constant. Note that the left hand side of Eqn.~\ref{eq:faith} is a measure of unfaithfulness of the explanation $\mathbf{E}_{u}$. So, higher values indicate higher degree of unfaithfulness. 

Now we derive upper bounds on unfaithfulness of explanations output by GNN \expmethods.

% \hima{let's not stuff weights into $\gamma$ here and in other places!}
\xhdr{Theorem 1} \textit{Given a node $u$ and a set $\mathcal{K}$ of node perturbations, the unfaithfulness (Sec.~\ref{sec:def_faith}, Eqn.~\ref{eq:faith}) of its explanation $\mathbf{E}_{u}$ can be bounded as follows:}
\begin{equation}
    \frac{1}{|\mathcal{K}|}\sum_{u'\in\mathcal{K}}||f(\mathcal{G}_{u'}) - f(t(\mathbf{E}_{u}, \mathcal{G}_{u'})) ||_{2} \leq \gamma~\frac{(1{+}|\mathcal{K}|)}{|\mathcal{K}|}~||\Delta||_{2}, \nonumber
    %\label{eq:unfaithful}
\end{equation}
where $f(\mathcal{G}_{u'}){=}\mathbf{\hat{y}}_{u'}$ are softmax predictions using original graph attributes, $f(t(\mathbf{E}_{u}, \mathcal{G}_{u'})){=}\mathbf{\hat{y}}_{u'}^{E}$ are softmax predictions using important attributes identified by $\mathbf{E}_{u}$, $\gamma$ is the product of the Lipschitz constants for GNN's activation function and layer weights, and $\Delta$ represents the embedding difference for node $u$ when we exclude unnecessary nodes/node features/edges as identified by an explanation.

\textit{Proof Sketch.~} In Theorem 1, the bound estimates an explanation's unfaithfulness and is tight if the Lipschitz constant of the activation functions and $\ell_{p}$-norm of the GNN weights are bounded. The theorem demonstrates that the bound is dependent not only on the output explanation but also on the weights of the GNN and is small when the difference between the embeddings using $\mathbf{E}_{u}$ (see notations) is small or for GNN layers having smaller Lipschitz constant values.
% Intuitively, the upper bound depends on the model properties (captured by the Lipschitzness of the weight matrices and activation functions of GNN  layers) and is tight for layers having smaller Lipschitz constant values.
% embedding for node $u$ where we exclude the unimportant features as identified by an explanation.

We show that unfaithfulness of a node feature explanation (like GraphLIME) is bounded by $\gamma_{11}\frac{1{+}|\mathcal{K}|}{|\mathcal{K}|}||(\mathbf{1}{-}\mathbf{r}_{u})\circ\mathbf{x}_{u}||_{2}$, where $\gamma_{11}$ is the product of the Lipschitz constant for GNN's activation function, weights of the last classification layer and self-attention weight of node $u$ across all GNN layers, and $\Delta$ for GraphLIME is $(\mathbf{1}{-}\mathbf{r}_{u})\circ\mathbf{x}_{u}$ representing the difference in node $u$'s features when we exclude the unimportant node features identified by the explanation. Similarly, for an edge-level explanations (like GraphMASK), the unfaithfulness is bounded by $\gamma_{12}\frac{(1{+}|\mathcal{K}|)}{|\mathcal{K}|}||\Delta_{\mathbf{x}_{v}}||_{2}$, where $\gamma_{12}$ is similar to $\gamma_{11}$ but uses weights associated with $u$'s immediate neighbors instead of self-attention weight and $\Delta_{\mathbf{x}_{v}}$ is the difference between embeddings of $u$'s neighbors where we exclude unnecessary edges as identified by the GraphMASK explanation. See Appendix~\ref{app:local_group_faith} for more details.
% edges that are not part of the explanation (\ie unnecessary edges as identified by the explanation).

\subsection{Theoretical Guarantees on Stability}
\label{sec:def_stable}

% \chirag{Just refer to the stability from previous work}
% \looseness=-1
Another key trait of a reliable explanation is that it should exhibit stability, i.e.,  infinitesimally small perturbations to an instance (which do not affect its model prediction) should not change its explanation drastically~\citep{lakkaraju2020robust,yuan2020explainability}.
% \xhdr{Limitations of existing stability definitions for GNN Explainers} Stability has been well studied in other areas of machine learning (ML) including adversarial ML and the broader XAI literature~\cite{alvarez2018towards,lakkaraju2020robust}. In the context of GNNs, previous stability definition captures the differences between the explanations for an original and perturbed instance \textit{w.r.t.} the ground truth explanation \cite{sanchez2020evaluating} or use the graph analog of \citet{lakkaraju2020robust} by arbitrarily deleting nodes/edges/node features \cite{yuan2020explainability}.
% \xhdr{Proposed stability} 
To this end, we use the definition of stability outlined in \citet{yuan2020explainability}. %Note that this notion of stability has been well studied in other areas of machine learning (ML) including adversarial ML and the broader explainable ML literature~\cite{sundararajan2017axiomatic}.  
An explanation $\mathbf{E}_{u}$ for node $u$'s prediction is considered stable if the explanations corresponding to $u$ (\ie $\mathbf{E}_{u}$) and its perturbation $u'$ (denoted by $\mathbf{E}_{u'}$) are similar. Here, the perturbation $u'$ is generated in the same way as above by rewiring the edges incident on $u$ with a small probability and/or making small changes to its node features. 
%and, similar to faithfulness, perturb the feature vectors and/or incident edges of a given node once by rewiring the edges incident on node $u$.
% with a small probability and/or making infinitesimally small changes to its node attributes, preserving the underlying data distribution. 
%We then say that an explanation $\mathbf{E}_{u}$ is stable if the explanations corresponding to $u$ (\ie $\mathbf{E}_{u}$) and its perturbed counterpart $u'$ (\ie $E_{u'}$) are similar. 
% Note that the perturbation strategy coupled with the fact that our metric do not require access to ground truth explanations makes it more general and reliable compared to prior works.
% Note that the perturbation $u'$ here is generated in the same way as above by rewiring the edges incident on node $u$ with a small probability and/or making infinitesimally small changes to its node attributes. 
%Similarly, the explanations' predictions are obtained using the mapping function $t$ as discussed above in Section~\ref{sec:def_faith}.
%With respect to GNNs, an \expmethod is stable if it produces the same explanation for graphs $\mathcal{G}_{u}$ and $\mathcal{G}_{u'}$, where $\mathcal{G}_{u'}$ is obtained by modifying $\mathcal{G}_{u}$ by perturbing its node features and/or rewiring its incident edges, and the black-box model $f$ outputs the same prediction for $\mathcal{G}_{u}$ and $\mathcal{G}_{u'}$.

\xhdr{Definition 2 (Stability)}\textit{ Given a node $u$ and its perturbation $u'$, an explanation $\mathbf{E}_{u}$ corresponding to node $u$ is said to be stable if:}
% it identifies node attributes (features or incident edges) that minimizes the error $\delta$ between the explanations generated for the original and the perturbed graph, \ie}
\begin{equation}
    % \arg\min_{E(f, \mathcal{G}'_{u})} (f(\mathcal{G}'_{u}) - f(E(f, \mathcal{G}'_{u})))^{2}
    \mathcal{D}\big(\mathbf{E}_{u}, \mathbf{E}_{u'}\big) \leq \delta,
%    ~~\text{if}~~ f(\mathcal{G}_{u}) = f(\mathcal{G}_{u'}),
    % \mathcal{D}\big(~E(f, \mathcal{G}_{u}), E(f, \mathcal{G}'_{u})~\big) \leq \delta
    % ~~\text{if}~~ f(\mathcal{G}_{u}) = f(\mathcal{G}'_{u}),
    \label{eq:stable}
\end{equation}
% \vskip -0.15in
where $\mathbf{E}_{u'}$ is the explanation for $u'$, $\mathcal{D}(\cdot)$ computes distance between two explanations, and $\delta$ is an infinitesimally small constant. The left side of Eqn.~\ref{eq:stable} measures instability of the explanation $\mathbf{E}_{u}$ and higher values indicate higher instability.

% \looseness=-1
Now, we take a representative \expmethod for each class of gradient-based (VanillaGrad~\citep{Simonyan14a}), perturbation-based (GraphMASK~\citep{schlichtkrull2020interpreting}), and surrogate-based (GraphLIME~\citep{huang2020graphlime}) methods and derive bounds on the instability of their explanations.

\xhdr{Theorem 2 (VanillaGrad)} \textit{Given a non-linear activation function $\sigma$ that is Lipschitz continuous, the instability (Sec.~\ref{sec:def_stable}, Eqn.~\ref{eq:stable}) of explanation $\mathbf{E}_{u}$ returned by VanillaGrad method can be bounded as follows:}
\begin{equation}
    ||\triangledown_{\mathbf{x}_{u'}}f-\triangledown_{\mathbf{x}_{u}}f||_{p} \leq \gamma_{3}||\mathbf{x}_{u'}-\mathbf{x}_{u}||_{p},
    \label{eq:grad_stable}
\end{equation}
% \vskip -0.1in
where $\gamma_{3}$ is the product of the $\ell_{\text{p}}$-norm of the prediction difference between the original and perturbed node, the weight of the final classification layer, and GNN's weight matrices.
% where $\gamma_{3}$ is the product of the $\ell_{\text{p}}$-norm of the prediction difference between the original and perturbed node and $\ell_{\text{p}}$-norm of the weight of other GNN layers. % 's weight matrices.
% is a constant, 
% $\mathbf{x}_{u}$ is node $u$'s feature vector, and $\mathbf{x}_{u'}$ is the perturbed node feature vector.

\textit{Proof Sketch.~} Using data processing inequalities, we prove that the $\ell_p$-norm of the difference between the gradient explanations generated using the original and perturbed node features is Lipschitz continuous. In Theorem 2, we demonstrate that the instability of a VanillaGrad explanation is upper bounded by the instability of the underlying GNN model, e.g., VanillaGrad explanation has higher instability for GNNs with higher $\ell_{p}$-norm layer weights. See Appendix~\ref{app:gradient_instability} for more details.
% Using data processing inequalities, we prove that the difference between the gradient explanations using the original and perturbed node features can be represented as a Lipschitz function. Further, in Theorem 2, we show that an explanation has higher instability if the norm of the weights of GNN layers are large. See Appendix~\ref{app:gradient_instability} for more details.

\looseness=-1
\xhdr{Theorem 3 (GraphMASK)} \textit{Given concatenated embeddings of node $u$ and $v$, the instability (Sec.~\ref{sec:def_stable}, Eqn.~\ref{eq:stable}) of explanation $\mathbf{E}_{u}$ returned by GraphMASK method can be bounded as follows:}
\begin{equation}
    ||\mathbf{z}^{l}_{u', v} - \mathbf{z}^{l}_{u, v}||_{2} \leq \gamma_{4}^{l}~||\mathbf{q}^{l}_{u',v} - \mathbf{q}^{l}_{u,v}||_{2},
    \label{eq:gmask_stable}
\end{equation}
where $\mathbf{z}^{l}_{u, v}$ is the GraphMASK explanation indicating whether an edge connecting node $u$ and $v{\in}\mathcal{N}_{u}$ in layer $l$ can be dropped or not, $\mathbf{q}^{l}_{u,v}$ is the concatenated embeddings for node $u$ and $v{\in}\mathcal{N}_{u}$ at layer $l$, and $\gamma_{4}^{l}$ denotes the Lipschitz constant which is a product of the $\ell_2$-norm of the weights in the $l$-th layer and the Lipschitz constants for the layer's normalization and softplus activation function.

\textit{Proof Sketch.} We prove that a GraphMASK's explanation for an edge at layer $l$ is Lipschitz continuous and $\gamma_{4}^{l}$ is the product of the Lipschitz constants of the layer's normalization function and the $\ell_{2}$-norm of the weight matrices of the erasure function. Intuitively, the instability of GraphMASK explanation is bounded by the difference between the GNN's embedding for the original and perturbed node and the Lipschitz constant, i.e., GraphMASK explanation have higher instability if the $\ell_2$ difference between the concatenated embeddings $\mathbf{q}^{l}_{u,v}$ for the original and perturbed nodes is high. Details are in Appendix~\ref{app:graphmask_instability}.

\xhdr{Theorem 4 (GraphLIME)} \textit{Given the centered Gram matrices for the original and perturbed node features, the instability (Sec.~\ref{sec:def_stable}, Eqn.~\ref{eq:stable}) of explanation $\mathbf{E}_{u}$ returned by GraphLIME method can be bounded as:}
\begin{equation}
    ||\beta^{'}_{k} - \beta_{k}||_{F} \leq \gamma_{2}~\cdot~ \text{tr}((\frac{1}{\mathbf{e}^{T}\mathbf{W}^{-1}\mathbf{e}})^{-1}-\mathbf{I}),
\end{equation}
where $\beta^{'}_{k}$ and $\beta_{k}$ are attribute importance generated by GraphLIME for the perturbed and original node features, $\gamma_{2}$ is the trace of the Gram matrix for the original graph and its predictions, 
% a noise-independent constant, 
$\mathbf{e}$ is an all-one vector, and $\mathbf{W}$ is a matrix comprising of the noise added to the graph.

\looseness=-1
\textit{Proof Sketch.} We prove that GraphLIME's instability is bounded by the trace of the Gram matrix for the output label and the perturbed Gram matrix due to noise added in the input graph. In Theorem 4, we first derive the closed-form of the attribute importance coefficient $\beta$ and then use it derive the upper bounds for the GraphLIME's instability. The theorem demonstrates that the bounds are tighter when the trace of the Gram matrix for the original and perturbed graphs are bounded, i.e., GraphLIME explanation have higher instability if the $\ell_{2}$-norm of the Gram matrix is high. Details are in Appendix~\ref{app:graphlime_instability}.

\subsection{Theoretical Guarantees on Fairness Preservation} % Preservation}
\label{sec:def_fair}
% \xhdr{3) Fairness}
\looseness=-1
As GNNs are increasingly employed in critical application domains, such as financial lending and criminal justice, it becomes crucial to ensure that the GNN explanations preserve the fairness properties and capture the biases of the underlying model. For instance, if a model is biased against a protected group, then its explanations should reflect that. This will help both model developers and practitioners in recognizing and addressing these prejudices. Analogously, if a model is fair, then its explanations should reflect that. To this end, we introduce and consider two notions of fairness for GNN \expmethods, namely, \textit{counterfactual fairness preservation} and \textit{group fairness preservation}.
%For instance, if a model is biased against a protected group, then an explanation should reflect that as it will help both model developers and practitioners to address these prejudices. %To this end, we present the first theoretical analysis of fairness preservation in GNN explanation methods and discuss both \textit{counterfactual} and \textit{group} fairness preservation.

\xhdr{a) Counterfactual Fairness Preservation} An explanation $\mathbf{E}_{u}$ preserves counterfactual fairness if the explanations corresponding to $u$ (\ie $\mathbf{E}_{u}$) and its sensitive feature perturbation $u^s$ (denoted by $\mathbf{E}_{u^s}$) are similar (dissimilar) if their model predictions are similar (dissimilar). Note that the sensitive feature perturbation $u^{s}$ is generated by flipping/modifying the sensitive feature $s$ in the node feature vector of $u$ (denoted by $\mathbf{x}_{u}$) while keeping everything else constant.

\xhdr{Definition 3 (Counterfactual Fairness Preservation)} \textit{Given a node $u$ and its sensitive feature perturbation $u^s$, an explanation $\mathbf{E}_{u}$ is said to preserve counterfactual fairness if:
}
\begin{align}
    \mathcal{D}\big(\mathbf{E}_{u}, \mathbf{E}_{u^{s}}\big) \propto f(\mathcal{G}_u) - f(\mathcal{G}_{u^{s}}),
    \label{eq:count_fair}
\end{align}
where $\mathbf{E}_{u^{s}}$ is explaining $u^{s}$'s prediction, $\mathcal{G}_u$ and $\mathcal{G}_{u^s}$ are subgraphs associated with $u$ and $u^{s}$, respectively. The left hand side of Eqn.~\ref{eq:count_fair} is a measure of counterfactual fairness mismatch of the explanation $\mathbf{E}_{u}$. %So, higher values indicate that the explanation is not preserving counterfactual fairness.

Analogous to the stability analysis (Sec.~\ref{sec:def_stable}), we now derive the bounds for counterfactual fairness mismatch of VanillaGrad, GraphMASK, and GraphLIME \expmethods. 

\xhdr{Theorem 5 (VanillaGrad)} \textit{Given a non-linear activation function $\sigma$ that is Lipschitz continuous, the counterfactual fairness mismatch (Sec.~\ref{sec:def_fair}, Eqn.~\ref{eq:count_fair}) of an explanation $\mathbf{E}_{u}$ returned by VanillaGrad method can be bounded as follows:}
\begin{equation}
    ||\triangledown_{\mathbf{x}_{u^{s}}}f-\triangledown_{\mathbf{x}_{u}}f||_{p} \leq \gamma_{3},
    \label{eq:counter_grad}
\end{equation}
where $\mathbf{x}_{u}$ is node $u$ features, $\mathbf{x}_{u^{s}}$ is the generated counterfactual by flipping $\mathbf{x}_{u}$'s sensitive feature, and $\gamma_{3}$ is similar to that in Theorem 2.

\textit{Proof Sketch.~} The $\ell_{p}$ distance between $\mathbf{x}_{u^{s}}$ and $\mathbf{x}_{u}$ is one in Eqn.~\ref{eq:counter_grad} as all the individual node attributes are the same except the sensitive attribute which is flipped (either from $0{\to}1$ or $1{\to}0$). The right term in Eqn.~\ref{eq:counter_grad} is similar to Eqn.~\ref{eq:grad_stable}.
Hence, using $||\mathbf{x}_{u'}{-}\mathbf{x}_{u}||_{p}{=}1$ in Eqn.~\ref{eq:grad_stable}, we derive the equation in Theorem 5. The theorem demonstrates that the counterfactual fairness mismatch for a VanillaGrad explanation is bounded by the $\ell_{p}$-norm of the weights of GNN layers and the prediction difference between the original and counterfactual node.

\xhdr{Theorem 6 (GraphMASK)} \textit{Given concatenated embeddings for node $u$ and $v$, the counterfactual fairness mismatch (Sec.~\ref{sec:def_fair}, Eqn.~\ref{eq:count_fair}) of an explanation $\mathbf{E}_{u}$ returned by GraphMASK method can be bounded as:}
\begin{equation}
    ||\mathbf{z}^{l}_{u^{s}, v} - \mathbf{z}^{l}_{u, v}||_{2} \leq \gamma_{4}^{l}~||\mathbf{q}^{l}_{u^{s},v} - \mathbf{q}^{l}_{u,v}||_{2},
    \label{eq:gmask_counter}
\end{equation}
\looseness=-1
where $\mathbf{z}^{l}_{u^{s}, v}$ is the GraphMASK explanation indicating whether an edge between node $u^{s}$ and $v{\in}\mathcal{N}_{u^{s}}$ in layer $l$ can be dropped or not, $\mathbf{q}^{l}_{u^{s},v}$ is the concatenated embeddings for node $u^{s}$ and $v{\in}\mathcal{N}_{u^{s}}$ at layer $l$, and $\gamma_{4}^{l}$ is the same constant as defined in Theorem 3.

\textit{Proof Sketch.~} A counterfactual node $u^{s}$ is generated by flipping one sensitive attribute from $\mathbf{x}_{u}$. 
% The left term in Eqn.~\ref{eq:gmask_counter} is similar to Eqn.~\ref{eq:gmask_stable} of Theorem 3. 
The term $\mathbf{q}^{l}_{u^{s},v}$ denotes the concatenated embedding for node $u^{s}$ and $v\in\mathcal{N}_{u^{s}}$ at layer $l$. Note, for the first layer, the right term of Eqn.~\ref{eq:gmask_counter} simplifies to just $\gamma_{4}^{l}$ as the $\ell_{p}$ between the $\mathbf{x}_{u^{s}}$ and $\mathbf{x}_{u}$ is one, i.e.,  $||\mathbf{x}_{u^{s}}{-}\mathbf{x}_{u}||_{p}{=}1$. The Theorem states that GraphMASK explanation have higher counterfactual fairness mismatch if the product of the $\ell_2$ difference between the concatenated embeddings $\mathbf{q}^{l}_{u,v}$ and the Lipschitz constant is high.

\xhdr{Theorem 7 (GraphLIME)} \textit{Given the centered Gram matrices for the original and counterfactual node attributes, the counterfactual fairness mismatch (Sec.~\ref{sec:def_fair}, Eqn.~\ref{eq:count_fair}) of an explanation $\mathbf{E}_{u}$ returned by GraphLIME method can be bounded as follows:}
\begin{equation}
    ||\beta^{s}_{k} - \beta_{k}||_{F} \leq \gamma_{2}~\cdot~ \text{tr}((\frac{1}{\mathbf{e}^{T}\mathbf{\bar{W}}^{-1}\mathbf{e}})^{-1}-\mathbf{I}),
\end{equation}
where $\beta^{s}_{k}$ and $\beta_{k}$ are GraphLIME's attribute importance for the $k$-th feature of $u^{s}$ and $u$, respectively, and $\gamma_{2}$ is similar to that in Theorem 4.
% where $\mathbf{E}_{u}$ is a node mask produced using the attribute importance $\beta^{'}_{k}$ and $\beta^{s}_{k}$ is the attribute importance generated by GraphLIME for the $k$-th attribute of node $u^{s}$.

\textit{Proof Sketch.~} Let us consider the $k$-th node feature as a binary sensitive attribute where $s{\in}\{0, 1\}$. In Theorem 7, we obtain a matrix $\mathbf{W}$ where $\eta$ in $\mathbf{W}$ will be either $1$ or $-1$, i.e., $\eta{=}1$ when flipping the sensitive attribute from $0{\to}1$, and $\eta{=}{-}1$ when flipping from $1{\to}0$. This does not change the positive semidefinite and invertible property of $\mathbf{W}$ as all diagonal elements are still 1 (invertible) and the off-diagonal elements are exponential (always positive). We denote this modified matrix as $\mathbf{\bar{W}}$.
The proof is similar to Theorem 4.

% \looseness=-1
\xhdr{b) Group Fairness Preservation} The notion of group fairness preservation has been quantified using different metrics, such as statistical parity (SP) \citep{dwork12:fairness}, equality of opportunity~\citep{hardt16:equality} etc. Here, we focus on SP which ensures that the probability of a positive outcome is independent of the sensitive features. Formally, an explanation $\mathbf{E}_{u}$ preserves group fairness if it accurately captures the SP of the underlying GNN $f$ in the \localreg around $u$. To operationalize this, we first construct a set $\mathcal{K}$ of nodes comprising of the original node $u$, and its perturbations. Next, we obtain the model and explanation $\mathbf{E}_{u}$'s predictions for all nodes in set $\mathcal{K}$. Note, the perturbations, the model and explanation predictions are obtained in the same way as described in Sec.~\ref{sec:def_faith}. We denote the model predictions and explanation predictions using $\mathbf{\hat{y}}_{\mathcal{K}}=\{\hat{y}_{1}, \hat{y}_{2}, \dots, \hat{y}_{|\mathcal{K}|}\}$ and $\mathbf{\hat{y}}^{\mathbf{E}_{u}}_{\mathcal{K}}=\{\hat{y}^{\mathbf{E}_{u}}_{1}, \hat{y}^{\mathbf{E}_{u}}_{2}, \dots, \hat{y}^{\mathbf{E}_{u}}_{|\mathcal{K}|}\}$ respectively. Now, if the SP computed using the two vectors $\mathbf{\hat{y}}_{\mathcal{K}}$ and $\mathbf{\hat{y}}^{\mathbf{E}_{u}}_{\mathcal{K}}$ are similar, then the explanation $\mathbf{E}_{u}$ is said to preserve group fairness. The statistical parity estimates for $\mathbf{\hat{y}}_{\mathcal{K}}$ can be computed as  $\text{SP}(\mathbf{\hat{y}}_{\mathcal{K}}){=}|\Pr(\hat{y}_{u'}{=}1|s{=}0){-}\Pr(\hat{y}_{u'}{=}1|s{=}1)|$, where the probabilities are computed over all the nodes in $\mathcal{K}$. Finally, $\text{SP}(\mathbf{\hat{y}}^{\mathbf{E}_{u}}_{\mathcal{K}})$ estimate can be computed analogously using explanation $\mathbf{E}_{u}$'s predictions.

\xhdr{Definition 4 (Group Fairness Preservation)}\textit{ Given a set $\mathcal{K}$ of node $u$ and its perturbations, an explanation $\mathbf{E}_{u}$ preserves group fairness if: 
}
\begin{equation}
    |~\text{SP}(\mathbf{\hat{y}}_{\mathcal{K}}) - \text{SP}(\mathbf{\hat{y}}_{\mathcal{K}}^{\mathbf{E}_u})~| \leq \delta,
    \label{eq:group_fair}
\end{equation}
where the left hand side term of the inequality in Eqn.~\ref{eq:group_fair} is a measure of group fairness mismatch of the explanation $\mathbf{E}_{u}$. So, higher values indicate that the explanation is not preserving group fairness.
% Here, we analyze the group fairness (Sec.~\ref{sec:lg-unfairness-theory}) and counterfactual fairness (Sec.~\ref{sec:counter-unfairness-theory}) properties of state-of-the-art GNN \expmethods and derive their corresponding bounds.
Next, we derive bounds for the graph fairness mismatch.

\xhdr{Theorem 8} \textit{Given a node $u$, a sensitive feature $s$, and a set $\mathcal{K}$ comprising of node $u$ and its perturbations, the group fairness mismatch (Sec.~\ref{sec:def_fair}, Eqn.~\ref{eq:group_fair}) of an explanation $\mathbf{E}_{u}$ can be bounded as follows:}
\begin{equation}
    |~\text{SP}(\mathbf{\hat{y}}_{\mathcal{K}}){-}\text{SP}(\mathbf{\hat{y}}_{\mathcal{K}}^{\mathbf{E}_{u}})~| \leq \sum_{\mathclap{s \in \{0, 1\}}}|\text{Err}_{D_{s}}(f(t(\mathbf{E}_{u},\mathcal{G}_{u'})){-}f(\mathcal{G}_{u'}))|, \nonumber
    %\label{eq:lg_unfairness}
\end{equation}
% \marinka{Chirag, see Slack. $D$ vs $D_s$. }
where $\text{SP}(\mathbf{\hat{y}}_{\mathcal{K}})$ and $\text{SP}(\mathbf{\hat{y}}^{\mathbf{E}_{u}}_{\mathcal{K}})$ are statistical parity estimates, $D$ is the joint distribution over node features $\mathbf{x}_{u'}$ in $\mathcal{G}_{u'}$ and their respective labels $\mathbf{y}_{u'}$ for $\forall u' \in \mathcal{K}$, $D_{s}$ is $D$ conditioned on the value of the sensitive feature $s$, and $\text{Err}_{D_{s}}(\cdot)$ is the model error under $D_{s}$.
% $\text{Err}_{D}(\cdot)=\mathbb{E}_{D}[\mathbf{y}_{u'}-f(\mathcal{G}_{u'})]$ is the model error under the joint distribution $D$.

%where $\text{SP}(\mathbf{\hat{y}}_{\mathcal{K}})$ and $\text{SP}(\mathbf{\hat{y}}^{\mathbf{E}_{u}}_{\mathcal{K}})$ are statistical parity estimates, $D$ is the joint distribution over node features $\mathbf{x}_{u'}$ in $\mathcal{G}_{u'}$ and their respective labels $\mathbf{y}_{u'}$ for $\forall u' \in \mathcal{K}$, $D_{s}$ is the conditional distribution of $D$ given a particular value of the sensitive feature $s$, and $\text{Err}_{D_{s}}(\cdot)$ is the model error under the conditional distribution $D_{s}$.
% $\text{Err}_{D}(\cdot)=\mathbb{E}_{D}[\mathbf{y}_{u'}-f(\mathcal{G}_{u'})]$ is the model error under the joint distribution $D$.

% \marinka{Chirag, see Slack. $D$ vs $D_s$. }
\textit{Proof Sketch.~} We show that group fairness mismatch of an explanation is bounded by the sum of the model errors $\text{Err}_{D_{s}}(\cdot)$ under distribution $D_{s}$. For a set of $\mathcal{K}$ nodes, the error is computed by taking the expectation of the difference between their true labels, set of model predictions using the original node features and incident edges, and their corresponding predictions using the explanation $\mathbf{E}_{u}$. In Theorem 8, the upper bound is the approximation error and reflects the error due to the prediction differences. The theorem shows that the ability of an explanation to preserve group fairness is quantified by the model error under the distribution $D_s$, i.e., an explanation obtains lower group fairness mismatch for smaller difference in model predictions when using only the important features identified by the explanation. Details are in Appendix~\ref{app:lg_unfairness_theory}.
\hide{
\textit{Proof Sketch.~} We show that group fairness mismatch of an \expmethod is bounded by the sum of the errors $\text{Err}_{D}(\cdot)$ of model $f$ under distribution $D$ conditioned on it's different sensitive attribute values. For a set of $\mathcal{K}$ nodes, the error is computed by taking the expectation of the difference between their true labels, set of model predictions using the original node features and incident edges, and their corresponding predictions using the explanation $\mathbf{E}_{u}$. Details are in Appendix~\ref{app:lg_unfairness_theory}.}

%% file: 050experiment.tex
\looseness=-1
Here, we present empirical analysis of state-of-the-art GNN \expmethods. Firstly, we verify the validity of our theoretical bounds by evaluating the faithfulness, stability, and fairness preservation properties of GNN \expmethods on node classification datasets. Next, we analyze the trade-offs between the aforementioned properties. Lastly, we evaluate the aforementioned properties on other downstream tasks such as link prediction and graph classification. 

% \chirag{use verbs for the xhdr}
% \chirag{Applications to other downstream tasks}
% \chirag{CHECK THE ICML experiments structure}
% and address the following key questions: \textbf{Q1)} Can we empirically verify that theoretical bounds are not violated? 
% \textbf{Q2)} Can we identify commonalities and differences in behaviors of \expmethods?
% \textbf{Q3)} What is the relationship between faithfulness, stability, and fairness?
% % , and how can our findings inform the development of future GNN \expmethods?
% \textbf{Q4)} Do our metrics generalize to different \expmethods, benchmarks, and downstream tasks?

% \hima{we no longer to evaluate the effectivness of metrics!! we need to say we use 9 datasets to empirically analyze the behavior of SOTA GNN explanation methods w.r.t. different key properties that we established, namely, fairness, faithfulness etc.}
\looseness=-1
\emph{Datasets.}~We use 9 real world datasets to empirically analyze the behavior of GNN \expmethods w.r.t.~key properties outlined in Sec.~\ref{sec:properties}. We consider 6 benchmark datasets (Cora, PubMed, Citeseer, Ogb-mag, Ogb-arxiv, MUTAG) and 3 datasets (German credit, Recidivism, Credit defaulter) with sensitive features (e.g., race, gender) from high-stakes domains. See Appendix~\ref{app:sec_experiment} for a detailed overview of  datasets.

\emph{Evaluation metrics.}~
% \chirag{refer back the LHS of the equations}
% We quantify how ``good'' the explanations are using properties from Sec.~\ref{sec:properties}.
We quantify the reliability of an explanation using properties from Sec.~\ref{sec:properties}.
% \hima{Chirag, make sure these metric computations are precisely described. This should exactly say what you compute in your code!}
In particular, we calculate unfaithfulness (Eqn.~\ref{eq:faith}) as: $\frac{1}{|\mathcal{K}|}\sum_{u'\in\mathcal{K}}||f(\mathcal{G}_{u'}){-}f(t(E_{u}, \mathcal{G}_{u'})) ||_{2}$, where the difference is between predictions made using original and masked node features/edges; instability (Eqn.~\ref{eq:stable}) as:
$\mathcal{D}\big(E_{u}, E_{u'}\big),$ where $\mathcal{D}$ is normalized $\ell_{1}$ distance between explanations generated for the original and perturbed node; counterfactual fairness mismatch (Eqn.~\ref{eq:count_fair}) as:
$\mathcal{D}\big(E_{u}, E_{u^{s}}\big),$ where $\mathcal{D}$ is normalized $\ell_{1}$ distance between explanations (Note that an explanation consists of node feature masks and/or edge masks as defined in Section~\ref{sec:properties} (Notation))
%\hima{chirag, check this below}
%(a vector of 0s and 1s corresponding to different node features and edges; 1 if the node feature/edge is important, 0 otherwise)
generated for the original and counterfactual node; and group fairness mismatch (Eqn.~\ref{eq:group_fair}) as: $|~\text{SP}(\mathbf{\hat{y}}_{\mathcal{K}}){-}\text{SP}(\mathbf{\hat{y}}_{\mathcal{K}}^{E_u})~|,$ where $\textrm{SP}$ is the statistical parity metric calculated on the group $\mathcal{K}$ of predictions.
For each node in the test set, we compute the above metrics and report their mean and standard errors for all \expmethods. While our theoretical analysis uses node classification as learning task, the metrics can also be used to evaluate explanations for other downstream tasks. %, including link prediction and graph classification. 

\emph{\Expmethods.} We evaluate 9 \expmethods, including gradient-based: VanillaGrad~\citep{Simonyan14a}, Integrated Gradients~\citep{sundararajan2017axiomatic}; perturbation-based: GNNExplainer~\citep{ying2019gnnexplainer}, PGExplainer~\citep{luo2020parameterized}, GraphMASK~\citep{schlichtkrull2020interpreting}; and surrogate-based methods: GraphLIME~\citep{huang2020graphlime}, PGMExplainer \citep{vu2020pgm}. 
As baselines, we consider two methods which produce random explanations: Random Node Features (a node feature mask drawn from an $M$-dimensional Gaussian vector) and Random Edges (an $N{\times}N$ edge mask drawn from a uniform distribution over $u$'s incident edges).

\emph{Implementation details.~} 
We follow the established approach of generating explanations~\citep{huang2020graphlime,ying2019gnnexplainer} and use reference implementations of \expmethods. We select top-$p$ ($p=25\%$) node features/edges, and use them to generate explanations for all \expmethods. 
% In all node classification experiments, we use GraphSAGE~\cite{hamilton2017inductive} as the GNN predictor. 
Details on hyperparameter selection, training of the GNN predictors, \expmethods, and training details for other downstream tasks are in Appendix~\ref{app:sec_experiment}.
\begin{figure*}[t]
    \centering
    \includegraphics[width=\linewidth]{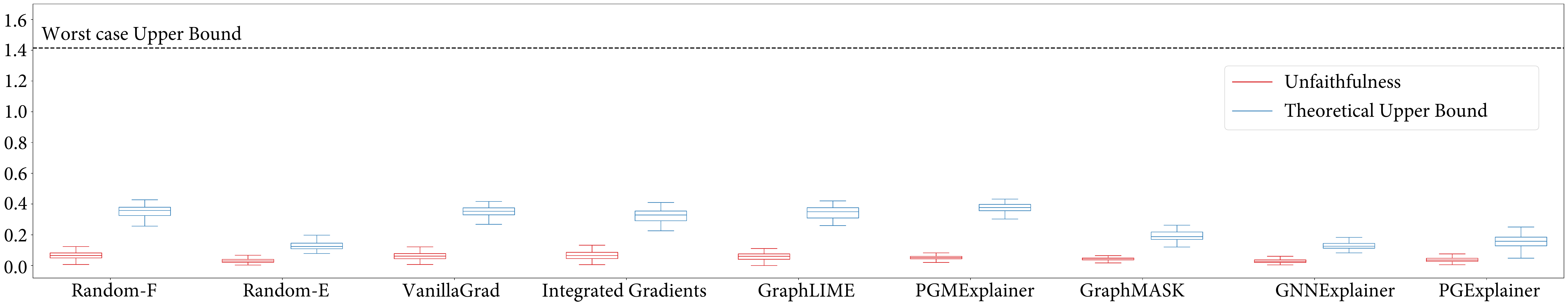}
    \caption{Empirically calculated unfaithfulness (in red) and our theoretical bounds for unfaithfulness (in blue) across nine \expmethods.  
    Results on the German credit graph dataset show no violations of our theoretical bounds. Results for stability, counterfactual fairness mismatch, and group fairness mismatch are shown in Appendix Figs.~\ref{app:fig_empirical_bound_fair}-\ref{app:fig_stab_count_bound}. 
    % Results are shown for the German credit graph dataset, showing that no violations of the bounds were detected in our experiments. Results for group unfairness and instability are shown in Appendix Figs.~\ref{app:fig_empirical_bound_fair}-\ref{app:fig_stab_count_bound}. 
    }
    \label{fig:empirical_bound_faith}
    % \vskip -0.1in
\end{figure*}

\begin{table*}[t]
\centering\small
\caption{
% \looseness=-1
Systematic evaluation of GNN \expmethods (random strategies (in grey), gradient- (in yellow), surrogate- (in purple), and perturbation-based (in red) methods) for node classification. Shown are average values and standard errors of evaluation metrics across all test set nodes. Arrows ($\downarrow$) indicate the direction of better performance. Surrogate-based methods produce most reliable explanations across all datasets. Note that fairness does not apply to some datasets (i.e., N/A) as they do not contain sensitive features. See Table~\ref{app:tab_node_metric_1}-\ref{app:tab_node_metric_2} for results on all 9 datasets.}
\label{tab:metric}
\renewcommand{\arraystretch}{0.9}
\setlength{\tabcolsep}{1.2pt}
% \vspace{-1mm}
{\begin{tabular}{cl|cccc}
 & & \multicolumn{4}{c}{Evaluation metrics} \\
\multirow{2}{*}{Dataset} & \multirow{2}{*}{Method} & \multirow{2}{*}{Unfaithfulness ($\downarrow$)} & \multirow{2}{*}{Instability ($\downarrow$)} & \multicolumn{2}{c}{Fairness Mismatch ($\downarrow$)} \\
 & & &  & \makecell{Counterfactual} & \makecell{Group} \\
\toprule
% \multirow{1}{2cm}{Recidivism graph} &
% \begin{tabular}[l]{@{}l@{}}\ccg{Random Node Features}\\\ccg{Random Edges}\\\ccy{{Gradients}}\\\ccy{Integrated Gradients}\\\ccb{{GraphLIME}}\\\ccb{{PGMExplainer}}\\\ccr{{GraphMASK}}\\\ccr{GNNExplainer}\\\ccr{PGExplainer}\end{tabular}& \begin{tabular}[c]{@{}c@{}}{{0.312}\std{0.004}}\\{{0.040}\std{0.001}}\\{{0.233}\std{0.004}}\\{{0.308}\std{0.005}}\\{{0.191}\std{0.004}}\\{{0.128}\std{0.001}}\\{{0.053}\std{0.002}}\\{{0.042}\std{0.001}}\\{{0.056}\std{0.001}}\end{tabular} &
% \begin{tabular}[c]{@{}c@{}}{{0.403}\std{0.002}}\\{0.376\std{0.000}}\\{{0.285}\std{0.003}}\\{{0.226}\std{0.003}}\\{{0.264}\std{0.004}}\\{{0.226}\std{0.002}}\\{{0.251}\std{0.003}}\\{{0.374}\std{0.000}}\\{{0.371}\std{0.001}}\end{tabular} & 
% \begin{tabular}[c]{@{}c@{}}{{0.403}\std{0.002}}\\{0.376\std{0.000}}\\{{0.173}\std{0.002}}\\{{0.104}\std{0.003}}\\{{0.072}\std{0.003}}\\{{0.223}\std{0.002}}\\{{0.013}\std{0.000}}\\{{0.364}\std{0.001}}\\{{0.355}\std{0.002}}\end{tabular} &
% \begin{tabular}[c]{@{}c@{}}{{0.144}\std{0.003}}\\{0.046\std{0.001}}\\{{0.114}\std{0.002}}\\{{0.139}\std{0.003}}\\{{0.107}\std{0.003}}\\{{0.130}\std{0.002}}\\{{0.060}\std{0.002}}\\{{0.051}\std{0.002}}\\{{0.064}\std{0.002}}\end{tabular} \\
\multirow{1}{2cm}{Credit defaulter graph} & \begin{tabular}[l]{@{}l@{}}\ccg{Random Node Features}\\\ccg{Random Edges}\\\ccy{{VanillaGrad}}\\\ccy{Integrated Gradients}\\\ccb{{GraphLIME}}\\\ccb{{PGMExplainer}}\\\ccr{{GraphMASK}}\\\ccr{GNNExplainer}\\\ccr{PGExplainer}\end{tabular} &
\begin{tabular}[c]{@{}c@{}}{{0.098}\std{0.002}}\\{{0.020}\std{0.001}}\\{{0.092}\std{0.002}}\\{{0.147}\std{0.003}}\\{{0.038}\std{0.002}}\\{{0.283}\std{0.002}}\\{{0.012}\std{0.001}}\\{{0.021}\std{0.001}}\\{{0.028}\std{0.001}}\end{tabular} &
\begin{tabular}[c]{@{}c@{}}{{0.426}\std{0.002}}\\{{0.376}\std{0.000}}\\{{0.333}\std{0.002}}\\{{0.140}\std{0.002}}\\{{0.225}\std{0.004}}\\{{0.156}\std{0.002}}\\{{0.036}\std{0.002}}\\{{0.375}\std{0.000}}\\{{0.364}\std{0.001}}\end{tabular} &
\begin{tabular}[c]{@{}c@{}}{{0.424}\std{0.002}}\\{0.376\std{0.000}}\\{{0.171}\std{0.002}}\\{{0.069}\std{0.001}}\\{{0.063}\std{0.003}}\\{{0.154}\std{0.002}}\\{{0.004}\std{0.000}}\\{{0.366}\std{0.000}}\\{{0.348}\std{0.002}}\end{tabular} & 
\begin{tabular}[c]{@{}c@{}}{{0.045}\std{0.002}}\\{0.017\std{0.001}}\\{{0.042}\std{0.002}}\\{{0.053}\std{0.002}}\\{{0.018}\std{0.001}}\\{{0.161}\std{0.003}}\\{{0.010}\std{0.001}}\\{{0.019}\std{0.001}}\\{{0.022}\std{0.001}}\end{tabular} \\
\midrule
% \multirow{1}{2cm}{Cora} &
% \begin{tabular}[l]{@{}l@{}}\ccg{Random Node Features}\\\ccg{Random Edges}\\\ccy{{Gradients}}\\\ccy{Integrated Gradients}\\\ccb{{GraphLIME}}\\\ccb{{PGMExplainer}}\\\ccr{{GraphMASK}}\\\ccr{GNNExplainer}\\\ccr{PGExplainer}\end{tabular}&
% \begin{tabular}[c]{@{}c@{}}{{0.002}\std{0.000}}\\{{0.004}\std{0.000}}\\{{0.002}\std{0.000}}\\{{0.002}\std{0.000}}\\{{0.001}\std{0.001}}\\{{0.016}\std{0.001}}\\{{0.023}\std{0.005}}\\{{0.003}\std{0.000}}\\{{0.112}\std{0.005}}\end{tabular} &
% \begin{tabular}[c]{@{}c@{}}{{0.181}\std{0.000}}\\{0.196\std{0.006}}\\{{0.154}\std{0.002}}\\{{0.894}\std{0.002}}\\{{0.052}\std{0.015}}\\{{0.224}\std{0.011}}\\{{0.600}\std{0.027}}\\{{0.377}\std{0.009}}\\{{0.372}\std{0.008}}\end{tabular} & 
% \begin{tabular}[c]{@{}c@{}}{\na}\\{\na}\\{\na}\\{\na}\\{\na}\\{\na}\\{\na}\\{\na}\\{\na}\end{tabular} &
% \begin{tabular}[c]{@{}c@{}}{\na}\\{\na}\\{\na}\\{\na}\\{\na}\\{\na}\\{\na}\\{\na}\\{\na}\end{tabular} \\
% \midrule
\multirow{1}{2cm}{Ogbn-arxiv} & \begin{tabular}[l]{@{}l@{}}\ccg{Random Node Features}\\\ccg{Random Edges}\\\ccy{{VanillaGrad}}\\\ccy{Integrated Gradients}\\\ccb{{GraphLIME}}\\\ccb{{PGMExplainer}}\\\ccr{{GraphMASK}}\\\ccr{GNNExplainer}\\\ccr{PGExplainer}\end{tabular} &
\begin{tabular}[c]{@{}c@{}}{{0.529}\std{0.002}}\\{{0.431}\std{0.002}}\\{{0.528}\std{0.002}}\\{{0.528}\std{0.002}}\\{{0.260}\std{0.003}}\\{{0.413}\std{0.002}}\\{{0.586}\std{0.001}}\\{{0.430}\std{0.002}}\\{{0.338}\std{0.002}}\end{tabular} &
\begin{tabular}[c]{@{}c@{}}{{0.375}\std{0.000}}\\{{0.378}\std{0.001}}\\{{0.359}\std{0.001}}\\{{0.372}\std{0.000}}\\{{0.374}\std{0.004}}\\{{0.270}\std{0.002}}\\{{0.125}\std{0.002}}\\{{0.376}\std{0.001}}\\{{0.381}\std{0.001}}\end{tabular} &
\begin{tabular}[c]{@{}c@{}}{\na}\\{\na}\\{\na}\\{\na}\\{\na}\\{\na}\\{\na}\\{\na}\\{\na}\end{tabular} & 
\begin{tabular}[c]{@{}c@{}}{\na}\\{\na}\\{\na}\\{\na}\\{\na}\\{\na}\\{\na}\\{\na}\\{\na}\end{tabular} \\
\bottomrule
\end{tabular}}
\vskip -0.1in
\end{table*}

%\subsection{Results}
%We now discuss our empirical results and insights.
% that answer the questions highlighted above. 
% in the beginning of this section.
%Here, we compare and quantitatively evaluate different families of \expmethods.

% \chirag{highlight the insights more}
% \chirag{across datasets, on average highlight the empirical best \expmethod and theoretical bounds}
% \chirag{can we also note more insights from the theoretical analysis.. give examples for which algorithm are they worst and how they correspond to the empirical results}

\looseness=-1
\xhdr{Empirically verifying our theoretical bounds}
% \xhdr{Q1) Results: \Expmethods satisfy theoretical bounds across all properties}
%
% \hima{the above sentence reads badly. We need to say more clearly exactly what we did before we jump to findings. E.g., "We evaluated our theoretical bounds analytically by computing the LHS of Eqn. 1 across all datasets for all the methods".}
We analytically evaluated our theoretical bounds by computing the LHS of Eqns.~\ref{eq:faith},\ref{eq:stable},\ref{eq:count_fair},\ref{eq:group_fair} for all the nine \expmethods.
Fig.~\ref{fig:empirical_bound_faith} shows the empirical and theoretical bounds for the unfaithfulness of methods, empirically confirming that none of our theoretical bounds were violated. Not only are we introducing theoretical bounds (RHS of the Eqns.~\ref{eq:faith},\ref{eq:stable},\ref{eq:count_fair},\ref{eq:group_fair}) for a broad range of \expmethods, but our bounds are also tight in the sense that their differences with empirical estimates are an order of magnitude smaller than those provided by the worst-case upper bounds (calculated using the maximum difference between the softmax scores of predictions made using original and masked node features/edges). For instance, the empirical estimate and the theoretical bound for the unfaithfulness of GNNExplainer match very closely (Fig.~\ref{fig:empirical_bound_faith}). We also computed the Spearman's rank correlation between the rankings for each of nine methods based on the theoretical bounds vs. empirical estimates of unfaithfulness. The correlation is $0.72$ (p-value=$0.03$) suggesting a strong correspondence between our theoretical bounds and empirical estimates. 
%Our results also indicate that the theoretical bounds have a strong correspondence with empirically calculated unfaithfulness, where the correlation between the ranking of nine explanation methods based on the theoretical vs.~empirical values is $0.72$ (p-value=$0.03$) using Spearman's rank. 
Results for stability and fairness preservation are shown in Figs.~\ref{app:fig_empirical_bound_fair}-\ref{app:fig_stab_count_bound} in the Appendix.

% We observe a difference of $62.2{\pm}4.7$ (the maximum difference being $70.71$) between the worst-case bounds and our theoretical bounds.
% Results for stability and fairness are in Appendix Figs.~\ref{app:fig_empirical_bound_fair}-\ref{app:fig_stab_count_bound}.
% While it is remarkable that theoretical bounds can be derived for such a broad range of \expmethods, we note that the bounds are not tight, yet the gap with empirical calculations remains modest and is an order of magnitude smaller than that provided by the worst case upper bound. Results for stability and fairness are in Appendix Figs.~\ref{app:fig_empirical_bound_fair}-\ref{app:fig_stab_count_bound}.

% e observe that all eight \expmethods preserve their respective upper bounds for unfaithfulness ().

% \xhdr{Q2) Results: Surrogate based methods generate most reliable explanations}
\xhdr{Evaluating the reliability of GNN \expmethods}
%\xhdr{Q2) Surrogate methods generate most reliable explanations} 
%
% \chirag{New evaluation numbers: Faithfulness: perturbation methods (+287\%); Stability: surrogate methods (+39.6\%); Counterfactual: surrogate methods (+54\%); Group: perturbation methods (+178.8\%)}
% \hima{DONT START WRITING ABOUT RESULTS WITHOUT SAYING ONE SENTENCE ABOUT WHAT EXACTLY IS YOUR EXPERIMENT}
We compare the  reliability of \expmethods by computing unfaithfulness (Eqn.~\ref{eq:faith}), instability (Eqn.~\ref{eq:stable}), counterfactual (Eqn.~\ref{eq:count_fair}) and group fairness mismatch (Eqn.~\ref{eq:group_fair}) metrics as described above.
% \hima{It is unclear what exactly are the things we computed, equation numbers or saying something like "we computed faithfulness as outline in Section xx" would be very helpful}.
Results in Table~\ref{tab:metric},\ref{app:tab_node_metric_1},\ref{app:tab_node_metric_2} show that surrogate-based \expmethods produce more reliable explanations than gradient- and perturbation-based methods. We observe that while no \expmethod simultaneously preserves all properties, on average across all node classification datasets (Fig.~\ref{app:fig_category_compare}), surrogate-based methods outperform other methods in instability (+55.1\%) and counterfactual fairness mismatch (+103.7\%), whereas perturbation-based methods outperform other methods in unfaithfulness (+23.8\%) and group fairness mismatch (+116.7\%). Interestingly, we find that the Random Edge baselines, which output explanations that correspond to random sets of edges incident on the target node, achieves the lowest (best possible) unfaithfulness score on most datasets, highlighting the urgent need for further probing of the behavior of all the GNN explainers. Finally, across all datasets (Fig.~\ref{app:fig_node_edge_compare}), explanations based on graph structure are slightly more faithful (+11.6\%), stable (+6.2\%), and counterfactually fair (+3.1\%).

\xhdr{Analyzing the trade-offs between faithfulness, stability, and fairness mismatch} 
%
% Our analyses yield similar results across datasets and families of \expmethods (Table~\ref{tab:metric}). 
% In particular, we find that none of the current \expmethods can consistently produce explanations that are simultaneously faithful, stable, and fair. 
% We find that none of the current \expmethods can consistently produce explanations that are simultaneously faithful, stable, and fair. 
% \hima{DONT START WRITING ABOUT RESULTS WITHOUT SAYING ONE SENTENCE ABOUT WHAT EXACTLY IS YOUR EXPERIMENT}
% \looseness=-1
We explore the trade-offs and possible connections between different properties defined in Sec.~\ref{sec:properties}. 
% \hima{It is unclear what exactly are the things we computed, equation numbers or saying something like "we computed faithfulness as outline in Section xx" would be very helpful}
Results in Table~\ref{tab:metric},\ref{app:tab_node_metric_1},\ref{app:tab_node_metric_2} (e.g., GNNExplainer on Recidivism and Credit defaulter graphs) indicate that methods with lower values of unfaithfulness (Eqn.~\ref{eq:faith}) also exhibit lower values of group fairness (Eqn.~\ref{eq:group_fair}) mismatch (and vice versa). To verify this connection, we compute the Pearson's and Spearman's rank correlation on the unfaithfulness and group fairness mismatch values and observe a strong positive correlation between them (Pearson's $r$=0.87 with p-value=4.82e-09; Spearman's  $\rho$=0.87 with p-value=2.67e-09). We  also observe a connection between instability (Eqn.~\ref{eq:stable}) and counterfactual fairness mismatch (Eqn.~\ref{eq:count_fair}) metrics, where methods with lower values of instability (e.g., GraphMASK on Credit defaulter graph) also exhibit lower values of counterfactual fairness mismatch (and vice versa). We further observe a strong positive correlation between instability and counterfactual fairness mismatch (Pearson's $r$=0.85 with p-value=2.72e-08; Spearman's $\rho$=0.86 with p-value=6.13e-09).
% State-of-the-art \expmethods are optimized to prioritize faithfulness over stable and fair explanations, suggesting that GNN \expmethods do not generate reliable explanations and that there are ample opportunities for algorithmic innovation.
% \chirag{add numbers in the content}
% \chirag{Other downstream tasks}
% \chirag{Add more comments}
% \chirag{MAINTAIN THE FLOW!!!}
% \chirag{remove general statements}
% \hima{This Q4 is raising more doubts about the generalization of our framework honestly! Why don't we roll these sentences into other questions?}

\xhdr{Other downstream tasks}
% Q4) Metrics generalize to 9 \expmethods, 9 datasets, and 3 downstream tasks}
% We experiment with 9 graph \expmethods on 9 real-world benchmark datasets and evaluate their performance using our metrics on key graph ML tasks. 
% Two important properties that enable o
We also apply our framework to link prediction and graph classification. 
% Link-prediction task is defined as predicting “linked” vs. “no-linked” labels for node pairs and graph classification entails predicting labels for a given graph.
% We extend some existing \expmethods that leverage gradient-based optimization and generate explanations for a given link prediction.
% Our framework scales and generalizes to small/large graphs, different \expmethods, and downstream tasks because the metrics focus on local regions and do not require access to ground-truth explanations. 
% Note that the explanations for any graph ML task include a subset of node features, edges, or both. As such, these explanations do not output any new components that are not already defined in our work. Hence, we can generate node features/edge masks from the explanations and calculate the metrics (as described in Sec.~\ref{sec:properties}).
% Finally, to show that the metrics generalize to explanations produced for link-prediction and graph classification tasks, we extend some existing \expmethods and present the results in Table~\ref{tab:link-graph-metric}.
\looseness=-1
We extend some existing methods for these tasks as most GNN \expmethods were developed only for node classification.  Explanations for these tasks also consist of a node feature mask and/or edge mask. % include a subset of node features, edges, or both. % and do not output any new components that are not already considered by our properties.
We generate these node feature/edge masks as described in Sec.~\ref{sec:properties} and evaluate all the properties.
% For example, for gradient \expmethods in link-prediction, we generate explanations for the prediction of the GNN model using its gradient \textit{w.r.t.} the node features of the associated nodes and then obtain the respective explanation masks for evaluations.
Similar to node classification, we observe that random baselines perform at least on par or better than the state-of-the-art GNN explanation methods 
%The generated explanations are highly unstable and Integrated Gradients method, on average across both datasets, is the most unstable with 26.8\% higher instability scores than other methods
(Table~\ref{tab:link-graph-metric}).

%% file: 060conclusion.tex
\looseness=-1

We introduce the first-ever theoretical analysis of the reliability of GNN \expmethods. To this end, we analyze the behavior of nine diverse state-of-the-art GNN explanation methods through the lens of various desirable properties such as faithfulness, stability, and fairness preservation.  Specifically, we establish theoretical upper bounds on the violation of each of these properties. Our theoretical analyses rely on information and probability theory concepts including data processing inequalities and total variation distance, and Lipschitz continuity. 
Further, we carry out extensive empirical analysis with nine real world datasets to verify our theoretical guarantees, and examine trade-offs between faithfulness, stability, and fairness preservation properties. These results yield critical insights on the behavior of state-of-the-art GNN \expmethods which can in turn inform the design and development of future explanation methods.

%% file: 070acknowledgement.tex
We would like to thank the anonymous reviewers for their insightful feedback. M.Z. is supported, in part by NSF under nos. IIS-2030459 and IIS-2033384, Harvard Data Science Initiative, Amazon Research Award, Bayer Early Excellence in Science Award, AstraZeneca Research, Roche Alliance with Distinguished Scientists (ROADS) Award, Department of the Air Force, and MIT Lincoln National Laboratory. H.L. is supported, in part by the NSF awards IIS-2008461 and IIS-2040989, and research awards from the Harvard Data Science Institute, Amazon, Bayer, and Google. H.L. would like to thank Mohan and Sujatha Lakkaraju, and Pracheer Gupta for all their inputs and support. The views expressed are those of the authors and do not reflect the official policy or position of the funding agencies.

%% file: supp_aistats.tex
% If your paper is accepted, change the options for the package
% aistats2022 as follows:
%
%\usepackage[accepted]{aistats2022}
%
% This option will print headings for the title of your paper and
% headings for the authors names, plus a copyright note at the end of
% the first column of the first page.

% If you set papersize explicitly, activate the following three lines:
%\special{papersize = 8.5in, 11in}
%\setlength{\pdfpageheight}{11in}
%\setlength{\pdfpagewidth}{8.5in}

% If you use natbib package, activate the following three lines:
%\usepackage[round]{natbib}
%\renewcommand{\bibname}{References}
%\renewcommand{\bibsection}{\subsubsection*{\bibname}}

% If you use BibTeX in apalike style, activate the following line:
%\bibliographystyle{apalike}

% If your paper is accepted and the title of your paper is very long,
% the style will print as headings an error message. Use the following
% command to supply a shorter title of your paper so that it can be
% used as headings.
%
%\runningtitle{I use this title instead because the last one was very long}

% If your paper is accepted and the number of authors is large, the
% style will print as headings an error message. Use the following
% command to supply a shorter version of the authors names so that
% they can be used as headings (for example, use only the surnames)
%
%\runningauthor{Surname 1, Surname 2, Surname 3, ...., Surname n}

% Supplementary material: To improve readability, you must use a single-column format for the supplementary material.
\onecolumn
\aistatstitle{Supplementary Materials: Probing GNN Explainers: Rigorous Theoretical and Empirical Analysis of GNN Explanation Methods}
\appendix
\input{070appendix}
\label{sec:appendix}

%% file: 070appendix.tex
% \section{Related Work}
% \label{sec:related}
% \input{020related}
\section{Overview of GNN Explanation Methods}\label{app:sec_exp_methods}
We now provide an in-depth overview of the different GNN \expmethods that we analyze in this work. 
As described in the main text (Sec.~\ref{sec:properties}), a $\mathcal{G}{=}(\mathcal{V}, \mathcal{E}, \mathbf{X})$ denote an undirected and unweighted graph comprising of a set of nodes $\mathcal{V}$, a set of edges $\mathcal{E}$, and a set of node feature vectors $\mathbf{X}{=}\{\mathbf{x}_{1}, \dots, \mathbf{x}_{N}\}$ corresponding to nodes in $\mathcal{V}$, where $\mathbf{x}_{u}{\in}\mathbb{R}^{M}$.
The GNN model's $f$ softmax prediction for node $u$ is given by $\mathbf{\hat{y}}_{u}=f(\mathcal{G}_{u})$, where $\mathcal{G}_{u}$ denotes the subgraph associated with node $u$ and $\mathbf{\hat{y}}_{u}\in[0, 1]^C$. 
% The associated adjacency matrix and node attributes for $\mathcal{G}_{u}$.
Finally, the explanation $E_{u}$ consists of a discrete node feature mask $\mathbf{r}_{u} \in \{0, 1\}^{M}$ for $v \in \mathcal{N}_{u}$ and/or a discrete edge mask $\mathbf{R}_{u} \in \{0, 1\}^{N \times N}$, where $1$ indicates that a node attribute/edge is included in the explanation and $0$ indicates otherwise.

\xhdr{Random Explanations} As a control, we consider two methods which produce random explanations: 1) \textit{Random Node Features}--a node feature mask defined by an $M$-dimensional Gaussian distributed vector; and 2) \textit{Random Edges}--an $N\times N$ edge mask drawn from a uniform distribution over $u$'s incident edges.

\xhdr{VanillaGrad} Gradient \citep{Simonyan14a} based explanation generate local explanations for the prediction of a differentiable GNN model $f$ using its gradient with respect to the node features $\mathbf{x}_{u}$: $\triangledown_{\mathbf{x}_{u}}f$. Intuitively, gradient represents how much difference a tiny change in each feature of a node $u$ would make to its corresponding classification score. VanillaGrad output an $M$-dimensional vector that comprises the vanilla gradient of the model, as explanations.

\xhdr{Integrated Gradients} Gradient explanations are often noisy and suffer from saturation problems \citep{sundararajan2017axiomatic}. Integrated gradients addresses the gradient saturation problem by averaging the gradients over a set of interpolated inputs derived using node $u$'s attribute and a baseline. Formally, integrated gradient explanation for a node $u$ is an $M$-dimensional vector given by:
\begin{equation}
    E_{u} = (\mathbf{x}_{u} - \mathbf{\Tilde{x}}) \times \int_{\alpha=0}^{1}\deriv{f(\mathcal{G}_{u'})}{\mathbf{x}_{u}} d\alpha,
\end{equation}
where $\mathbf{\Tilde{x}}$ is the baseline input which can be vector of all zeros/ones and $\mathcal{G}_{u'}$ denotes the graph with the interpolated node attribute $\mathbf{x}_{u'}{=}\mathbf{\Tilde{x}}{+}\alpha(\mathbf{x}_{u}{-}\mathbf{\Tilde{x}})$.

\xhdr{GraphLIME} GraphLIME \citep{huang2020graphlime} is a local interpretable model explanation for GNNs that identifies a nonlinear interpretable model over the neighbors of a node that is locally faithful to the node's prediction. It considers a feature-wise kernelized nonlinear method called Hilbert-Schmidt Independence Criterion Lasso (HSIC Lasso) as an explanation model. For each node prediction, the HSIC Lasso objective function is defined as:
\begin{equation}
    \min_{\beta \in \mathbb{R}^{d}} \frac{1}{2} || \mathbf{L} - \sum_{k=1}^{M} \beta_{k} \mathbf{K}^{(k)} ||^{2}_{F} + \rho||\beta||_{1},
    \label{eq:hsic}
\end{equation}
where $||\cdot||_{F}$ is the Frobenius norm, $\rho \geq 0$ is the regularization parameter, $||\cdot||_{1}$ is the $l_{1}$ norm to enforce sparsity, $\mathbf{L}$ is the centered Gram matrix, $L_{ij} = L(y_{i}, y_{j})$ is the kernel for the output labels of the nodes, $\mathbf{K}^{k}$ is the centered gram matrix for the $k$-th feature, and $K_{ij}=K(x_{i}^{(k)}, x_{j}^{(k)})$ is the kernel for the $k$-th dimensional input node features $\mathbf{x}_{u}$.

\xhdr{PGMExplainer} Probabilistic Graphical models (PGMs) are statistical models that encode complex distributions using graph-based representation and provides a simple interpretation of the dependencies of those underlying random variables. Specifically, Bayesian network, a PGM represents conditional dependencies among variables via a directed acyclic graph. Given a target prediction $\hat{y}_{u}$ to be explained, our proposed PGM explanation is the optimal Bayesian network $\mathcal{B}^{*}$ of the following optimization:
\begin{equation}
    \argmax_{\mathcal{B}\in\mathcal{B}_{E_{u}}} R_{\hat{y}_{u}} (\mathcal{B}),
\end{equation}
where $R_{\hat{y}_{u}}:E_{u} \to \mathbb{R}$ associates each explanation with a score, $\mathcal{B}_{E_{u}}$ is the set of all Bayesian networks, the optimization is subjected to the condition that the number of variables in $\mathcal{B}$ is bounded by a constant to encourage a compact solution and another constraint to ensure that the target prediction is included in the explanation.

\xhdr{GraphMASK}  GraphMASK \citep{schlichtkrull2020interpreting} detect edges at each layer $l$ that can be ignored without affecting the output model predictions.
In general, dropping edges from a given graph is non-trivial, and, hence, for each edge at layer $l$, GraphMASK learns a binary choice $z^{l}_{u, v}$ that indicates whether the edge can be dropped, and then replaces the given edge with a learned baseline. Here, $z^{l}_{u, v}$ indicates an edge connecting node $u$ and $v$.
GraphMASK learns $z^{l}_{u,v}$ for all $(u,v) \in \mathcal{E}$ for the training data points using an erasure function $g_{\pi}$, where $\pi$ denotes the parameters of $g$.
For explaining a given prediction, GraphMASK uses this trained function $g_{\pi}$ and generates a masked representation of the graph using: 
% $\Tilde{h}_{u}^{l}{=} \mathbf{Z}^{l}\circ h_{u}^{l} + \alpha^{l}\circ(1-\mathbf{Z}^{l})$,
\begin{equation}
    \Tilde{h}_{u}^{l}{=} \mathbf{Z}^{l}\circ h_{u}^{l} + \alpha^{l}\circ(1-\mathbf{Z}^{l}),
\end{equation}
where $\mathbf{Z}^{l}$ comprises of all the individual binary scores $z^{l}_{u, v}=g_{\pi}(h_{u}^{l}, h_{v}^{l})$, $\alpha^{l}$ is the learned baseline, and `$\circ$' denotes the element-wise Hadamard product.
% %  which can also be replaced with samples from a Gaussian distribution

\xhdr{GNNExplainer} For a single-instance explanation for node $u$, GNNExplainer \citep{ying2019gnnexplainer} generates an explanation by identifying a subgraph of the computation graph for $u$ and a subset of node features  that are most influential for the model $f$'s prediction. Formally, GNNExplainer determines the importance of individual node attributes and incident edges for node $u$ by leveraging Mutual Information ($MI$) using the following optimization framework:
\begin{equation}
    \max_{\mathcal{G}_{S}} MI (Y, (\mathcal{G}_{S}, \mathbf{X}_{S})),
    \label{eq:gnnex}
\end{equation}
where $\mathcal{G}_{S}\subseteq\mathcal{G}_{u}$ is a subgraph and $\mathbf{X}_{S}$ is the associated node attributes that are important for the GNN's prediction $\hat{y}_{u}$. Intuitively, $MI$ quantifies the change in probability of prediction $\hat{y}_{u}$ when $u$'s computation graph is limited to the explanation graph $\mathcal{G}_{S}$ and its corresponding node attributes $\mathbf{X}_{S}$.

\xhdr{PGExplainer} In contrast to GNNExplainer, PGExplainer \citep{luo2020parameterized} generates explanation only on the graph structure. The direct optimization of the mutual information framework in Eqn.~\ref{eq:gnnex} is intractable \citep{ying2019gnnexplainer,luo2020parameterized}. Thus, PGExplainer consider a relaxation by assuming that the explanatory graph $\mathcal{G}_{S}$ is a Gilbert random graph, where selections of edges from the input graph $\mathcal{G}_{u}$ are conditionally independent to each other. Due to the discrete nature of $\mathcal{G}_{S}$, PGExplainer employs the \textit{reparameterization trick} where they relax the edge weights from binary to continuous variables in the range $(0, 1)$ and then optimize the objective function using gradient-based methods. It approximates the sampling process of $\mathcal{G}_{S}$ with a determinant function of parameters $\Omega$, temperature $\rho$, and an independent random variable $\epsilon$. Specifically, the weight for each edge $\hat{e}(i,j)$ is calculated by:
\begin{equation}
    \epsilon \sim \text{Uniform} (0, 1), ~~\hat{e}(i,j)=\sigma((\text{log}\epsilon - \text{log}(1-\epsilon) + \omega_{ij})/\rho),
\end{equation}
where $\sigma(\cdot)$ is the Sigmoid function, $\omega_{ij} \in \mathbb{R}$ is a trainable parameter. With the reparameterization, the objective function of PGExplainer becomes:
\begin{equation}
    \min_{\Omega} \mathbb{E}_{\epsilon \sim\text{Uniform}(0,1)}~ H(Y|\mathcal{G}_{u}=\mathcal{\hat{G}}_{S}),
\end{equation}
where $H$ is the conditional entropy when the computational graph for $\mathcal{G}_{u}$ is restricted to $\mathcal{G}_{S}$.

Note that recently proposed explanation methods, such as GNN-LRP~\citep{schnake2020higher}, Xgnn~\citep{yuan2020xgnn}, and Gem~\citep{lin2021generative}, can also be empirically analyzed using our framework because all these methods generate explanations using subgraphs around the target node with associated edges and node attributes. Hence, we can select the topk edges/nodes from the subgraph and calculate the metric scores.

\section{Proofs for Theorems in Section \ref{sec:properties}}
\label{app:sec_proof}
\subsection{Analyzing Faithfulness of GNN Explanation Methods}\label{app:local_group_faith}

\xhdr{Theorem 1} \textit{Given a node $u$ and a set $\mathcal{K}$ of node perturbations, the unfaithfulness (Sec.~\ref{sec:def_faith}, Eqn.~\ref{eq:faith}) of its explanation $E_{u}$ can be bounded as follows:}
\begin{equation}
    \frac{1}{|\mathcal{K}|}\sum_{u'\in\mathcal{K}}||f(\mathcal{G}_{u'}) - f(t(E_{u}, \mathcal{G}_{u'})) ||_{2} \leq \gamma~\frac{(1{+}|\mathcal{K}|)}{|\mathcal{K}|}~||\Delta||_{2}, \nonumber
    %\label{eq:unfaithful}
\end{equation}
where $f(\mathcal{G}_{u'}) =\mathbf{\hat{y}}_{u'}$ are softmax predictions that use original attributes and $f(t(E_{u}, \mathcal{G}_{u'}))=\mathbf{\hat{y}}_{u'}^{E}$ are softmax predictions that use attributes marked important by explanation $E_{u}$. Further, $\gamma$ denotes the product of the Lipschitz constants for GNN's activation function and GNN's weight matrices across all layers in the GNN, and $\Delta$ is an \expmethod-specific term.

% % \subsection{Local-Group Unfaithfulness}
% % \label{app:local_group_faith}
% \xhdr{Theorem 1} \textit{Given a node $u$ and a set of $\mathcal{K}$ of node perturbations, the unfaithfulness (Sec.~\ref{sec:def_faith}, Eqn.~\ref{eq:faith}) of its explanation $E_{u}$ can be bounded as follows:}
% % Given node $u$ and its respective set of $\mathcal{K}$ perturbations, the explanation for node $u$ satisfies faithfulness, i.e.,}
% \begin{equation}
%     \frac{1}{|\mathcal{K}|}\sum_{u'\in\mathcal{K}}||f(\mathcal{G}_{u'}) - f(t(E_{u}, \mathcal{G}_{u'})) ||_{2} \leq \gamma~\frac{(1{+}|\mathcal{K}|)}{|\mathcal{K}|}~||\Delta||_{2},
%     % ||(\mathbf{1}-\mathbf{r}_{u})\circ\mathbf{x}_{u}||_{2},
%     % ||\hat{y}_{u} - \hat{y}_{u}^{E}||_{2} + \sum_{k{=}1}^{\mathcal{K}}||\hat{y}_{k}^{'} - \hat{y}_{k}^{E}||_{2} \leq \gamma_{1}~(1 + \mathcal{K})~||(\mathbf{1}-\mathbf{r}_{u})\circ\mathbf{x}_{u}||_{2},
% \end{equation}
% where $f(\mathcal{G}_{u'}){=}\mathbf{\hat{y}}_{u'}$ and $f(t(E_{u}, \mathcal{G}_{u'})){=}\mathbf{\hat{y}}_{u'}^{E}$ are the softmax predictions using the original attributes and the attributes marked important by the explanation $E_{u}$ for the original and perturbed nodes respectively, $\gamma$ is a product of the Lipschitz constants of the activation function and the weight matrices of all message-passing layers in the GNN model, and $\Delta$ term is dependent on the explanations provided by the explainer method.

\textit{Proof.~} Without loss of generality, we use a two-layer GNN model for our proof and show its extension to a GNN model with $L$ layers. The two-layer GNN formulated as a message-passing network is defined as:
\begin{align*}
    % & \hat{y}_{u}{=}\text{softmax}(\mathbf{h}_{u}^{2})\\
    & \mathbf{h}_{u}^{1}{=}\text{sp}(\mathbf{W}_{a}^{1}\mathbf{x}_{u}+\mathbf{W}_{n}^{1}\sum_{v\in \mathcal{N}_{u}}\mathbf{x}_{v}) \\
    & \mathbf{h}_{u}^{2}{=}\mathbf{W}_{\text{fc}}\mathbf{h}_{u}^{1}+\mathbf{b}\\
    & \mathbf{\hat{y}}_{u}{=}\text{softmax}(\mathbf{h}_{u}^{2}),
\end{align*}
where $\mathbf{W}_{n}^{1}$ is the weight matrix associated with the neighbors of node $u$, $\mathbf{W}_{a}^{1}$ is the self-attention weight matrix at layer one, and ``sp'' is the softplus activation function.
For the fully-connected layer, we have $\mathbf{W}_{\text{fc}}$ as the weight matrix and $\mathbf{b}$ as the bias term.
The softplus function is a smooth approximation of the ReLU function.
We generate $|\mathcal{K}|$ perturbations of node $u$ by adding normal Gaussian noise to the node features, i.e., $\mathbf{x}_{u'}{=}\mathbf{x}_{u}{+}\tau$, and rewire edges with some probability $p_{r}$.
For faithfulness, we get the predictions for node $u$ using the weighted node features of node $u$, i.e., the element-wise product between $\mathbf{r}_{u}$ (the feature importance mask generated as an explanation) and $\mathbf{x}_{u}$. Let $\mathbf{\hat{y}}_{u}^{E}$ denote the softmax output for node $u$ using the explanation $E_{u}$, i.e., $f(t(E_{u}, \mathcal{G}_{u}))$. Therefore, the updated equations using the explanations are:
% Therefore, $\mathbf{h}_{u}^{1'} = \text{softplus}(\mathbf{W}_{a}^{1}\mathbf{x}_{u}^{'}+\mathbf{W}_{n}^{1}\sum_{v\in \mathcal{N}_{u}} \mathbf{x}_{v})$.
\begin{align*}
    & (\mathbf{h}_{u}^{1})^{E}=\text{sp}\big(\mathbf{W}_{a}^{1}(\mathbf{r}_{u}\circ\mathbf{x}_{u})+\mathbf{W}_{n}^{1}\sum_{v\in \mathcal{N}'_{u}} \mathbf{x}_{v}\big)\\
    & (\mathbf{h}_{u}^{2})^{E}=\mathbf{W}_{\text{fc}}(\mathbf{h}_{u}^{1})^{E}+\mathbf{b}\\
    & \mathbf{\hat{y}}_{u}^{E}=\text{softmax}((\mathbf{h}_{u}^{2})^{E}),
\end{align*}
where $\mathcal{N}'_{u}$ denotes the new neighborhood for node $u$ due to the adjacency mask matrix $\mathbf{R}_{u}$.
The difference between the predicted labels for the original and the important node features can be given as:
\begin{equation}
    \mathbf{\hat{y}}_{u} - \mathbf{\hat{y}}_{u}^{E} = \text{softmax}(\mathbf{h}_{u}^{2})-\text{softmax}((\mathbf{h}_{u}^{2})^{E})
    \label{eq:faith_1}
\end{equation}

\xhdr{Corollary 1} \textit{For any differentiable function $g: \mathbb{R}^{a} \to \mathbb{R}^{b}$,}
\begin{equation}
    || g(x) - g(y) ||_{2} \leq ||\mathbf{J}||^{*}_{F}||x-y||_{2}~~~~ \forall x, y \in \mathbb{R},
\end{equation}
\looseness=-1where $||\mathbf{J}||^{*}_{F} = \max_{x} ||\mathbf{J}||_{F}$ and $\mathbf{J}$ is the Jacobian matrix of $g(x)$ w.r.t. $x$. This is implied from the mean value theorem, where for any function $g(x)$ and its derivative $\deriv{g(x)}{x}$, we have: $g(x) - g(y)=\deriv{g(\phi)}{\phi}(x-y)$, for some $\phi \in (y, x)$.

Hence, taking the norm on both sides in Eqn.~\ref{eq:faith_1}, we get,
\begin{align*}
    & ||\mathbf{\hat{y}}_{u} - \mathbf{\hat{y}}_{u}^{E}||_{2} = ||\text{softmax}(\mathbf{h}_{u}^{2})-\text{softmax}((\mathbf{h}_{u}^{2})^{E})||_{2}\\
    & \leq \mathcal{C}_{\text{fc}}||\mathbf{h}_{u}^{2}-(\mathbf{h}_{u}^{2})^{E}||_{2} \tag{Using Corollary 1},
\end{align*}
where $\mathcal{C}_{\text{fc}}$ represents the Lipschitz constant for the softmax function.
Substituting the values of $\mathbf{h}_{u}^{2}$ and $(\mathbf{h}_{u}^{2})^{E}$ we get:
\begin{align*}
    & ||\mathbf{\hat{y}}_{u} - \mathbf{\hat{y}}_{u}^{E}||_{2} \leq 
    \mathcal{C}_{\text{fc}}||\mathbf{W}_{\text{fc}}\mathbf{h}_{u}^{1}+\mathbf{b} -\mathbf{W}_{\text{fc}}(\mathbf{h}_{u}^{1})^{E}-\mathbf{b}||_{2} \\
    & \leq 
    \mathcal{C}_{\text{fc}}||\mathbf{W}_{\text{fc}}\mathbf{h}_{u}^{1}-\mathbf{W}_{\text{fc}}(\mathbf{h}_{u}^{1})^{E}||_{2} \\
    & \leq 
    \mathcal{C}_{\text{fc}}~||\mathbf{W}_{\text{fc}}||_{2}~||\mathbf{h}_{u}^{1} -(\mathbf{h}_{u}^{1})^{E}||_{2} \tag{Using Cauchy-Schwartz inequality}
\end{align*}
Substituting the values of $\mathbf{h}_{u}^{1}$ and $(\mathbf{h}_{u}^{1})^{E}$ we get:
\begin{align*}
    & ||\mathbf{\hat{y}}_{u} - \mathbf{\hat{y}}_{u}^{E}||_{2} \leq 
    \mathcal{C}_{\text{fc}}~||\mathbf{W}_{\text{fc}}||_{2}~||\text{sp}\big(\mathbf{W}_{a}^{1}\mathbf{x}_{u}+\mathbf{W}_{n}^{1}\sum_{v\in \mathcal{N}_{u}} \mathbf{x}_{v}\big) - \text{sp}\big(\mathbf{W}_{a}^{1}(\mathbf{r}_{u}\circ\mathbf{x}_{u})+\mathbf{W}_{n}^{1}\sum_{v\in \mathcal{N}'_{u}} \mathbf{x}_{v}\big)||_{2} \\
    & \leq 
    \mathcal{C}_{\text{fc}}~\mathcal{C}_{1}~||\mathbf{W}_{\text{fc}}||_{2}~||\mathbf{W}_{a}^{1}\mathbf{x}_{u}+\mathbf{W}_{n}^{1}\sum_{v\in \mathcal{N}_{u}} \mathbf{x}_{v} - \mathbf{W}_{a}^{1}(\mathbf{r}_{u}\circ\mathbf{x}_{u})-\mathbf{W}_{n}^{1}\sum_{v\in \mathcal{N}'_{u}} \mathbf{x}_{v}||_{2} \tag{Using Corollary 1}\\
    & \leq 
    \mathcal{C}_{\text{fc}}~\mathcal{C}_{1}~||\mathbf{W}_{\text{fc}}||_{2}~\big(||\mathbf{W}_{a}^{1}\big(\mathbf{x}_{u} - (\mathbf{r}_{u}\circ\mathbf{x}_{u})\big)||_{2} + ||\mathbf{W}_{n}^{1}\Delta_{\mathbf{x}_{v}}||_{2}\big), \tag{Using triangle inequality}
    % & \leq \mathcal{C}_{\text{fc}}~\mathcal{C}_{1}~||\mathbf{W}_{\text{fc}}||_{2}~||\mathbf{W}_{a}^{1}||_{2}~||\mathbf{x}_{u} - (\mathbf{r}_{u}\circ\mathbf{x}_{u})||_{2} \tag{Using Cauchy-Schwartz inequality}
\end{align*}
where $\Delta_{\mathbf{x}_{v}}$ is the difference between the representations of the neighbors of $u$ after dropping edges using the edge masks. This difference can be neglected for gradient and GraphLIME methods as they provide explanations in the node feature space. 
% and  the  and is assumed to be very small because the model is smooth function and large restructuring of the graph can lead to different representations for the node \citep{schlichtkrull2020interpreting}. 
Now, using Cauchy-Schwartz inequality, the prediction difference for a node $u$ using its original and just important node features is bounded by:
\begin{equation}
    ||\mathbf{\hat{y}}_{u} - \mathbf{\hat{y}}_{u}^{E}||_{2} \leq \mathcal{C}_{\text{fc}}~\mathcal{C}_{1}~||\mathbf{W}_{\text{fc}}||_{2}~||\mathbf{W}_{a}^{1}||_{2}~||(\mathbf{1}-\mathbf{r}_{u})\circ\mathbf{x}_{u})||_{2},
    \label{eq:faith_2}
\end{equation}
where $\mathcal{C}_{1}$ is the Lipschitz constant for the softplus activation function and $\mathbf{1} \in \mathbb{R}^{M}$ is vector with all ones. For mathematical brevity, let $\gamma_{11}=\mathcal{C}_{\text{fc}}~\mathcal{C}_{1}~||\mathbf{W}_{\text{fc}}||_{2}~||\mathbf{W}_{a}^{1}||_{2}$. Similarly, the prediction difference for GraphMASK which provides an explanation with respect to edges is bounded by:
\begin{equation}
    ||\mathbf{\hat{y}}_{u} - \mathbf{\hat{y}}_{u}^{E}||_{2} \leq 
    \mathcal{C}_{\text{fc}}~\mathcal{C}_{1}~||\mathbf{W}_{\text{fc}}||_{2}~||\mathbf{W}_{n}^{1}||_{2}~||\Delta_{\mathbf{x}_{v}}||_{2},
    \label{eq:faith_edge}
\end{equation}
where $\gamma_{12}=\mathcal{C}_{\text{fc}}~\mathcal{C}_{1}~||\mathbf{W}_{\text{fc}}||_{2}~||\mathbf{W}_{n}^{1}||_{2}$.

\xhdr{Node feature explanations} Since all the perturbed nodes use the same node feature explanation $\mathbf{r}_{u}$, 
% the probability $p_{r}$ of rewiring an edge is very small to maintain the underlying graph structure, 
we obtain the difference between the predictions for a perturbed node $u'$  using the perturbed and the masked node features, i.e.,
\begin{align*}
    & ||\mathbf{\hat{y}}_{u'} - \mathbf{\hat{y}}_{u'}^{E}||_{2} \leq
    \gamma_{11}~||\mathbf{x}_{u'} - (\mathbf{r}_{u}\circ\mathbf{x}_{u'})||_{2},
\end{align*}
where $\mathbf{\hat{y}}_{u'}$ is the softmax prediction using the perturbed node feature $\mathbf{x}_{u'}=\mathbf{x}_{u}+\tau$, and as per the definition of faithfulness we use the explanation mask of node $u$ for node $u'$. Finally, we get:
\begin{equation}
    ||\mathbf{\hat{y}}_{u'} - \mathbf{\hat{y}}_{u'}^{E}||_{2} \leq 
    \gamma_{11}~||(\mathbf{1}-\mathbf{r}_{u})\circ\mathbf{x}_{u'}||_{2}
    \label{eq:faith_3}
\end{equation}
For faithfulness, we generate a set of $\mathcal{K}$ perturbed nodes and get $|\mathcal{K}|$ predictions from the model for each of the corresponding perturbations. Using Eqns.~\ref{eq:faith_2} and \ref{eq:faith_3}, and getting the predictions from all $|\mathcal{K}|$ perturbations, we get:
\begin{align*}
    % & ||\hat{y}_{u} - \hat{y}_{u}^{E}||_{2} + \sum_{k{=}1}^{\mathcal{K}}||\hat{y}_{k}^{'} - \hat{y}_{k}^{E}||_{2} \leq \gamma_{1}~||(\mathbf{1}-\mathbf{r}_{u})\circ\mathbf{x}_{u})||_{2} + \gamma_{1}\sum_{k{=}1}^{\mathcal{K}}~||(\mathbf{1}-\mathbf{r}_{u})\circ(\mathbf{x}_{u}^{'})_{k}||_{2}\\
    & \sum_{u'\in\mathcal{K}}||\mathbf{\hat{y}}_{u'} - \mathbf{\hat{y}}_{u'}^{E}||_{2} \leq \gamma_{11}~||(\mathbf{1}-\mathbf{r}_{u})\circ\mathbf{x}_{u})||_{2} + \gamma_{11}\sum_{u'\in \mathcal{K}}~||(\mathbf{1}-\mathbf{r}_{u})\circ\mathbf{x}_{u'}||_{2}\\
    & \leq \gamma_{11}~||(\mathbf{1}-\mathbf{r}_{u})\circ\mathbf{x}_{u})||_{2} + \gamma_{11}~||(\mathbf{1}-\mathbf{r}_{u})\circ\sum_{u'\in \mathcal{K}}\mathbf{x}_{u'}||_{2},
\end{align*}
% where $\mathbf{\hat{y}}_{u'}$ represents the softmax prediction of node $u'$.
% $(\mathbf{x}_{u}^{'})_{k}$ and $k \in \{1\dots\mathcal{K}\}$. 
Assuming $\tau$ to be drawn from a normal distribution, we get:
$\sum_{u'\in \mathcal{K}}\mathbf{x}_{u'} = |\mathcal{K}|\mathbf{x}_{u} + \sum_{u'\in \mathcal{K}}\tau_{k}$. For sufficiently large $|\mathcal{K}|$, we have: $\sum_{u'\in \mathcal{K}}(\mathbf{x}_{u}^{'})_{k} \approx |\mathcal{K}|\mathbf{x}_{u}$.
Putting everything together and taking the average across $\mathcal{K}$ samples we get, 
\begin{align*}
    & \frac{1}{\mathcal{|K|}}\sum_{u'\in\mathcal{K}}||\mathbf{\hat{y}}_{u'} - \mathbf{\hat{y}}_{u'}^{E}||_{2} \leq \frac{1}{\mathcal{|K|}} \big( \gamma_{11}~||(\mathbf{1}-\mathbf{r}_{u})\circ\mathbf{x}_{u})||_{2} + \gamma_{11}~|\mathcal{K}|~||(\mathbf{1}-\mathbf{r}_{u})\circ\mathbf{x}_{u}||_{2} \big)\\
    & \leq \gamma_{11}~\frac{(1{+}|\mathcal{K}|)}{\mathcal{|K|}}~||(\mathbf{1}-\mathbf{r}_{u})\circ\mathbf{x}_{u}||_{2}
\end{align*}
% the term on the left-side of the inequality represents the local-group faithfulness w.r.t. a single explanation for node $u$ (as defined in Eqn.~\ref{eq:faith}).
For a GNN model with $L$ message-passing layers and one fully-connected layer for node classification, $\gamma_{11}$ takes the general form of:
\begin{equation}
    \gamma_{11} = \mathcal{C}_{\text{fc}}~||\mathbf{W}_{\text{fc}}||_{2}~\prod_{l=1}^{L}\mathcal{C}_{l}||\mathbf{W}_{a}^{l}||_{2},
    \label{eq:gamma_all}
\end{equation}
where $\mathcal{C}_{\text{fc}}$ is the Lipschitz constant for the softmax activation operating on the fully-connected layer, $\mathbf{W}_{\text{fc}}$ is the weight matrix associated with the fully-connected layer, $\mathcal{C}_{l}$ is the Lipschitz constant of the softplus activation of each message-passing layer, and $\mathbf{W}_{a}^{l}$ is the self-attention weight associated with the $l$-th message-passing layer.

\xhdr{Edge explanations} Note, the difference between the predictions for a perturbed node $u'$ using the perturbed and the masked node features will be similar to Eqn.~\ref{eq:faith_edge} as for faithfulness, the perturbations are made only in node $u$, i.e.,
\begin{align*}
    ||\mathbf{\hat{y}}_{u'} - \mathbf{\hat{y}}_{u'}^{E}||_{2} \leq 
    \gamma_{12}~||\Delta_{\mathbf{x}_{v}}||_{2},
    \label{eq:faith_edge}
\end{align*}
Also, since we use the same explanation for all the nodes in set $\mathcal{K}$, the bound for faithfulness using edge explanations is given by:
\begin{align*}
    & \frac{1}{\mathcal{|K|}}\sum_{u'\in\mathcal{K}}||\mathbf{\hat{y}}_{u'} - \mathbf{\hat{y}}_{u'}^{E}||_{2} \leq \gamma_{12}~\frac{(1{+}|\mathcal{K}|)}{|\mathcal{K}|}~||\Delta_{\mathbf{x}_{v}}||_{2},
\end{align*}
where as in Eqn.~\ref{eq:gamma_all}, $\gamma_{12}$ can take the general form for $L$ message-passing layers as: $\gamma_{12} = \mathcal{C}_{\text{fc}}~||\mathbf{W}_{\text{fc}}||_{2}~\prod_{l=1}^{L}\mathcal{C}_{l}||\mathbf{W}_{n}^{l}||_{2}$.
\subsection{Analyzing Stability of GNN Explanation Methods}
\label{app:instability-theory}
\subsubsection{VanillaGrad Explanation}
\label{app:gradient_instability}
\xhdr{Theorem 2} \textit{Given a non-linear activation function $\sigma$ that is Lipschitz continuous, the instability (Sec.~\ref{sec:def_stable}, Eqn.~\ref{eq:stable}) of explanation $E_{u}$ returned by VanillaGrad method can be bounded as follows:}
\begin{equation}
    ||\triangledown_{\mathbf{x}_{u'}}f-\triangledown_{\mathbf{x}_{u}}f||_{p} \leq \gamma_{3}||\mathbf{x}_{u'}-\mathbf{x}_{u}||_{p},
    \label{app:grad_stable}
\end{equation}
% \vskip -0.1in
where $\gamma_{3}$ is a constant, $\mathbf{x}_{u}$ is node $u$'s feature vector, and $\mathbf{x}_{u'}$ is the perturbed node feature vector.

% \xhdr{Theorem 2} \textit{Given a non-linear activation function $\sigma$ that is Lipschitz continuous, the instability (Sec.~\ref{sec:def_stable}, Eqn.~\ref{eq:stable}) of explanation $E_{u}$ returned by VanillaGrad method can be bounded as follows:}
% % \textit{Given a non-linear activation function $\sigma$ that is Lipschitz continuous, the gradient explanation for an original and perturbed graph is stable, i.e.,}
% \begin{equation}
%     ||\triangledown_{\mathbf{x}_{u'}}f-\triangledown_{\mathbf{x}_{u}}f||_{p} \leq \gamma_{3}||\mathbf{x}_{u'}-\mathbf{x}_{u}||_{p},
%     \label{app:eq_grad_stable}
% \end{equation}
% where $\gamma_{3}$ is a constant, $\mathbf{x}_{u}$ is the original node feature, and $\mathbf{x}_{u'}$ is the perturbed node feature.

\textit{Proof.~} Similar to Sec.~\ref{app:local_group_faith}, let us consider a two-layer GNN model trained on a node classification task using softmax cross-entropy loss function with the first layer a message-passing GNN layer and the second layer as a fully-connected layer. The cross-entropy (CE) loss is given as:
\begin{equation}
    \text{CE} = -\sum_{i} y_{i}\log \hat{y}_{u},  % \hat{y}_{i},
\end{equation}
where $\mathbf{y}$ is a vector with one one non-zero element (which is 1), $\mathbf{\hat{y}}_{u}{=}\text{softmax}(\mathbf{h}_{u}^{2})$, $\mathbf{h}_{u}^{2}{=}\mathbf{W}_{\text{fc}}\mathbf{h}_{u}^{1} + \mathbf{b}$, and $\mathbf{h}_{u}^{1}{=}\text{sp}(\mathbf{W}_{a}^{1}\mathbf{x}_{u} + \mathbf{W}_{n}^{1}\sum_{v\in \mathcal{N}_{u}} \mathbf{x}_{v})$.
$\mathbf{W}_{n}^{1}$ is the weight matrix associated with the neighbors of node $u$ and $\mathbf{W}_{a}^{1}$ is the self-attention weight matrix at layer one.
For the fully-connected layer, we have $\mathbf{W}_{\text{fc}}$ as the weight matrix and $\mathbf{b}$ as the bias term.
``sp'' is the softplus activation function which is a smooth approximation of the ReLU function.
For stability, we generate $\mathbf{x}_{u'}$ by adding noise to the node features of node $u$ and keep everything else constant. Therefore, $\mathbf{h}_{u'}^{1} = \text{sp}(\mathbf{W}_{a}^{1}\mathbf{x}_{u'}+\mathbf{W}_{n}^{1}\sum_{v\in \mathcal{N}'_{u}} \mathbf{x}_{v})$.
Now, the differentiation of the model \textit{w.r.t.} the node features can be given as:
\begin{align*}
    \triangledown_{\mathbf{x}_{u}}f = \deriv{(\text{CE})}{\mathbf{x}_{u}} = \deriv{(\text{CE})}{\mathbf{h}_{u}^{2}}\deriv{\mathbf{h}_{u}^{2}}{\mathbf{h}_{u}^{1}}\deriv{\mathbf{h}_{u}^{1}}{\mathbf{x}_{u}}, \tag{By chain rule}
\end{align*}
Note, the advantage of using \textit{softplus} activation function is that it is differentiable for all $x$, i.e.,
\begin{align*}
    & \text{sp}(x)=\textit{ln}(1+\exp^{x}) \\
    & \deriv{(\text{sp}(x))}{\text{x}} = \frac{\exp^{x}}{1 + \exp^{x}} \cdot \frac{(1/\exp^{x})}{(1/\exp^{x})}\\
    & = \frac{1}{1 + \exp^{-x}}\\
    & = \sigma(x),
\end{align*}
where $\sigma(\cdot)$ is the sigmoid activation function. Putting it all together we get,
\begin{eqnarray}
    \triangledown_{\mathbf{x}_{u}}f = (\mathbf{y}_{u}-\mathbf{\hat{y}}_{u})(\mathbf{W}_{\text{fc}})^{T}\sigma(\mathbf{W}_{a}^{1}\mathbf{x}_{u} + \mathbf{W}_{n}^{1}\sum_{v\in \mathcal{N}_{u}} \mathbf{x}_{v})(\mathbf{W}_{a}^{1})^{T}\\
    \triangledown_{\mathbf{x}_{u'}}f = (\mathbf{y}_{u}-\mathbf{\hat{y}}_{u})(\mathbf{W}_{\text{fc}})^{T}\sigma(\mathbf{W}_{a}^{1}\mathbf{x}_{u'} + \mathbf{W}_{n}^{1}\sum_{v\in \mathcal{N}'_{u}} \mathbf{x}_{v})(\mathbf{W}_{a}^{1})^{T}
\end{eqnarray}
Note, $\mathbf{\hat{y}}_{u}$ is same for both  original and perturbed node according to the Definition 2 in Sec.~\ref{sec:properties} and we drop the second neighborhood term since the probability ($p_{r}$) of rewiring the edges is very small to maintain the original graph structure.
Hence, subtracting the explanations (model gradients) for the original and perturbed node features and taking the norm on both sides, we get:
\begin{align*}
    & ||\triangledown_{\mathbf{x}_{u'}}f-\triangledown_{\mathbf{x}_{u}}f||_{p} = ||(\mathbf{y}_{u}-\mathbf{\hat{y}}_{u})(\mathbf{W}_{\text{fc}})^{T}\big(\sigma(\mathbf{W}_{a}^{1}\mathbf{x}_{u'})-\sigma(\mathbf{W}_{a}^{1}\mathbf{x}_{u})\big)(\mathbf{W}_{a}^{1})^{T}||_{p},
\end{align*}
Using Cauchy-Schwartz inequality, we get:
\begin{align*}
    & ||\triangledown_{\mathbf{x}_{u'}}f-\triangledown_{\mathbf{x}_{u}}f||_{p} \leq ||\mathbf{y}_{u}-\mathbf{\hat{y}}_{u}||_{p}~||(\mathbf{W}_{\text{fc}})^{T}||_{p}~||\sigma(\mathbf{W}_{a}^{1}\mathbf{x}_{u'})-\sigma(\mathbf{W}_{a}^{1}\mathbf{x}_{u})||_{p}~||(\mathbf{W}_{a}^{1})^{T}||_{p}
\end{align*}
Assuming that $\sigma(\cdot)$ is normalized Lipschitz, i.e., $||\sigma{(b)}-\sigma{(a)}||_{p} \leq ||b-a||_{p}$, we get,
\begin{align*}
    & ||\triangledown_{\mathbf{x}_{u'}}f-\triangledown_{\mathbf{x}_{u}}f||_{p} \leq ||\mathbf{y}_{u}-\mathbf{\hat{y}}_{u}||_{p}~||(\mathbf{W}_{\text{fc}})^{T}||_{p}~||\mathbf{W}_{a}^{1}\mathbf{x}_{u'}-\mathbf{W}_{a}^{1}\mathbf{x}_{u}||_{p}~||(\mathbf{W}_{a}^{1})^{T}||_{p}\\
    & \leq ||\mathbf{y}_{u}-\mathbf{\hat{y}}_{u}||_{p}~||(\mathbf{W}_{\text{fc}})^{T}||_{p}~||\mathbf{W}_{a}^{1}(\mathbf{x}_{u'}-\mathbf{x}_{u})||_{p}~||(\mathbf{W}_{a}^{1})^{T}||_{p}\\
    & \leq ||\mathbf{y}_{u}-\mathbf{\hat{y}}_{u}||_{p}~||(\mathbf{W}_{\text{fc}})^{T}||_{p}~||\mathbf{W}_{a}^{1}||_{p}~||\mathbf{x}_{u'}-\mathbf{x}_{u}||_{p}~||(\mathbf{W}_{a}^{1})^{T}||_{p} \tag{Using Cauchy-Schwartz inequality}\\
    & \leq \gamma_{3}~||\mathbf{x}_{u'}-\mathbf{x}_{u}||_{p},
\end{align*}
where $\gamma_{3} = ||\mathbf{y}_{u}-\mathbf{\hat{y}}_{u}||_{p}~||(\mathbf{W}_{\text{fc}})^{T}||_{p}~||\mathbf{W}_{a}^{1}||_{p}~||(\mathbf{W}_{a}^{1})^{T}||_{p}$.

For a GNN model with $L$ message-passing layers and one fully-connected layer for node classification, $\gamma_{3}$ takes the general form of:
\begin{equation}
    \gamma_{3} = ||\mathbf{y}_{u}-\mathbf{\hat{y}}_{u}||_{p}~||(\mathbf{W}_{\text{fc}})^{T}||_{p}~\prod_{l=1}^{L}||\mathbf{W}_{a}^{l}||_{p}~||(\mathbf{W}_{a}^{1})^{T}||_{p},
    % \mathcal{C}_{\text{fc}}~||\mathbf{W}_{\text{fc}}||_{2}~\prod_{l=1}^{L}\mathcal{C}_{l}||\mathbf{W}_{a}^{l}||_{2},
\end{equation}
where $\mathbf{W}_{a}^{l}$ is the self-attention weight associated with the $l$-th message-passing layer.
\subsubsection{GraphMASK}
\label{app:graphmask_instability}
\xhdr{Setup} GraphMASK computes the parameters $\pi$ for the erasure function using fully-connected layers with non-linearity and layer-wise normalization.
The scalar location parameter $z^{l}_{u, v}$ is given as:
\begin{equation}
    z^{l}_{u, v} = \mathbf{W}_{2}^{l}~\text{sp}(\text{LN}^{l}(\mathbf{W}_{1}^{l}\mathbf{q}^{l}_{u,v})),
    \label{eq:gpi}
\end{equation}
where sp is the softplus activation function, LN is the layer normalization function, and $\mathbf{q}^{l}_{u,v}$ represents the concatenated representations of $\mathbf{h}_{u}^{l}$ and $\mathbf{h}_{v}^{l}$. Note, the representations at $l{=}0$ are the the node features in the original graph $\mathbf{x}_{u}$ and $\mathbf{x}_{v}$. Since we are considering the task of node-classification, there is no relation-specific representation with respect to each edge. 
% Further, as before, we use ``sp'' which is the differentiable approximation of ReLU activation.
For explaining node $u$'s prediction, GraphMASK generates $z^{l}_{u, v}$ for all its incident edges. Finally, the parameters $\pi$ of the erasure function are trained on multiple datapoints, and then used for explaining predictions \citep{schlichtkrull2020interpreting}. 
For deriving the instability and counterfactual fairness mismatch of GraphMASK, we first state a lemma that helps us prove that a layer normalization function is Lipschitz.

\xhdr{Lemma 1} \textit{A normalization function \text{LN} for layer $l$ is Lipschitz continuous, i.e.,}
\begin{equation}
    ||\text{LN}^{l}(\mathbf{h}_{u'}^{l}) - \text{LN}^{l}(\mathbf{h}_{u}^{l})||_{2} \leq C_{\text{LN}}^{l}~||(\mathbf{h}_{u'}^{l}-\mathbf{h}_{u}^{l})||_{2},
    \label{app:eq_lip_ln}
\end{equation}
where $C_{\text{LN}}^{l}$ is the Lipschitz constant of the normalization function for layer $l$.

\textit{Proof.~} The layer normalization function is a reparametrization trick that significantly reduces the problem of coordinating updates across different layers.
A given representation $\mathbf{h}_{u}^{l}$ is normalized using mean and standard deviation parameters that are learned during the training stage, i.e.,
\begin{equation}
    \text{LN}^{l}(\mathbf{h}_{u}^{l}) = \frac{(\mathbf{h}_{u}^{l}-\mu^{l})}{\varsigma^{l}},
    \label{app:eq_layer_norm}
\end{equation}
where $\mu^{l}$ is the mean and $\varsigma^{l}$ is the standard deviation of the representations at layer $l$, and are fixed after the training completes. Using Eqn.~\ref{app:eq_layer_norm}, the difference between the layer normalized output at layer $l$ of a perturbed and original representation can be given as:
\begin{align*}
    & \text{LN}^{l}(\mathbf{h}_{u'}^{l}) - \text{LN}^{l}(\mathbf{h}_{u}^{l}) = \frac{(\mathbf{h}_{u'}^{l}-\mu^{l})}{\varsigma^{l}} - \frac{(\mathbf{h}_{u}^{l}-\mu^{l})}{\varsigma^{l}} \\
    & \text{LN}^{l}(\mathbf{h}_{u'}^{l}) - \text{LN}^{l}(\mathbf{h}_{u}^{l}) = \frac{(\mathbf{h}_{u'}^{l}-\mathbf{h}_{u}^{l})}{\varsigma^{l}}
\end{align*}
Taking $L_{2}$-norm on both sides and applying Cauchy-Schwartz inequality, we get:
\begin{align*}
    & ||\text{LN}^{l}(\mathbf{h}_{u'}^{l}) - \text{LN}^{l}(\mathbf{h}_{u}^{l})||_{2} \leq ||\frac{1}{\varsigma^{l}}||_{2}~||(\mathbf{h}_{u'}^{l}-\mathbf{h}_{u}^{l})||_{2}
\end{align*}
For consistency, we define $C_{\text{LN}}^{l}{=}||\frac{1}{\varsigma^{l}}||_{2}$ as the Lipschitz constant for the $l^{\text{th}}$ normalization layer.

\xhdr{Theorem 3} \textit{Given concatenated embeddings of node $u$ and $v$, the instability (Sec.~\ref{sec:def_stable}, Eqn.~\ref{eq:stable}) of explanation $E_{u}$ returned by GraphMASK method can be bounded as follows:}
\begin{equation}
    ||\mathbf{z}^{l}_{u', v} - \mathbf{z}^{l}_{u, v}||_{2} \leq \gamma_{4}^{l}~||\mathbf{q}^{l}_{u',v} - \mathbf{q}^{l}_{u,v}||_{2},
\end{equation}
where $\mathbf{z}^{l}_{u, v}$ is the explanation output by GraphMASK indicating whether an edge connecting node $u$ and $v\in\mathcal{N}_{u}$ in layer $l$ can be dropped or not, $\mathbf{q}^{l}_{u,v}$ is the concatenated embeddings for node $u$ and $v\in\mathcal{N}_{u}$ at layer $l$, and $\gamma_{4}^{l}$ denotes the Lipschitz constant which is a product of the weights of the $l$-th fully-connected layer, and the Lipschitz constants for the layer normalization and softplus activation function.
% \xhdr{Theorem 3}
% \textit{Given concatenated embeddings of node $u$ and $v$, the instability (Sec.~\ref{sec:def_stable}, Eqn.~\ref{eq:stable}) of explanation $E_{u}$ returned by GraphMASK method can be bounded as follows:}
% % \textit{Given the concatenated representation of the original and perturbed graph, the scalar location explanation for an edge in layer $l$ is stable, i.e.,}
% \begin{equation}
%     ||z^{l}_{u', v} - z^{l}_{u, v}||_{2} \leq \gamma_{4}^{l}~||\mathbf{q}^{l}_{u',v} - \mathbf{q}^{l}_{u,v}||_{2},
%     \label{app:eq_gmask_stable}
% \end{equation}
% where $\gamma_{4}^{l}$ denotes the Lipschitz constant which is a product of the weights of the $l$-th fully-connected layer, and the Lipschitz constants for the layer normalization function and softplus activation function.

\textit{Proof.~} Using Eqn.~\ref{eq:gpi}, the scalar location parameter for a perturbed node $u'$ can be written as:
\begin{align*}
    z^{l}_{u', v} = \mathbf{W}_{2}^{l}~\text{sp}(\text{LN}^{l}(\mathbf{W}_{1}^{l}\mathbf{q}^{l}_{u',v}))
\end{align*}
Note, for explanation all the parameters of the fully-connected layers are fixed as they are trained initially using a set of training data points.
\begin{align*}
    & z^{l}_{u', v} - z^{l}_{u, v}= \mathbf{W}_{2}^{l}~\text{sp}(\text{LN}^{l}(\mathbf{W}_{1}^{l}\mathbf{q}^{l}_{u',v})) - \mathbf{W}_{2}^{l}~\text{sp}(\text{LN}^{l}(\mathbf{W}_{1}^{l}\mathbf{q}^{l}_{u,v})) \\
    & z^{l}_{u', v} - z^{l}_{u, v}= \mathbf{W}_{2}^{l}~\big(\text{sp}(\text{LN}^{l}(\mathbf{W}_{1}^{l}\mathbf{q}^{l}_{u',v})) - \text{sp}(\text{LN}^{l}(\mathbf{W}_{1}^{l}\mathbf{q}^{l}_{u,v}))\big)
\end{align*}
Taking $L_{2}$-norm on both sides and applying Cauchy-Schwartz inequality, we get:
\begin{align*}
    & ||z^{l}_{u', v} - z^{l}_{u, v}||_{2} \leq ||\mathbf{W}_{2}^{l}||_{2}~||\text{sp}(\text{LN}^{l}(\mathbf{W}_{1}^{l}\mathbf{q}^{l}_{u',v})) - \text{sp}(\text{LN}^{l}(\mathbf{W}_{1}^{l}\mathbf{q}^{l}_{u,v}))||_{2}\\
    & \leq C_{\text{SP}}||\mathbf{W}_{2}^{l}||_{2}~||\text{LN}^{l}(\mathbf{W}_{1}^{l}\mathbf{q}^{l}_{u',v}) - \text{LN}^{l}(\mathbf{W}_{1}^{l}\mathbf{q}^{l}_{u,v})||_{2}, \tag{Using Corollary 1}
\end{align*}
where $C_{\text{SP}}$ is the Lipschitz constant for the softplus activation function. Simplifying further we get:
\begin{align*}
    & ||z^{l}_{u', v} - z^{l}_{u, v}||_{2} \leq C_{\text{SP}}~C_{\text{LN}}^{l}||\mathbf{W}_{2}^{l}||_{2}~||\mathbf{W}_{1}^{l}\mathbf{q}^{l}_{u',v} - \mathbf{W}_{1}^{l}\mathbf{q}^{l}_{u,v}||_{2} \tag{Using Lemma 1}\\
    & \leq C_{\text{SP}}~C_{\text{LN}}^{l}||\mathbf{W}_{2}^{l}||_{2}~||\mathbf{W}_{1}^{l}||_{2}~||\mathbf{q}^{l}_{u',v} - \mathbf{q}^{l}_{u,v}||_{2} \tag{Using Cauchy-Schwartz inequality}
\end{align*}
Hence, for a given layer $l$ the difference between the scalar location parameter of a perturbed and original node $u$ is given by:
\begin{equation}
    ||z^{l}_{u', v} - z^{l}_{u, v}||_{2} \leq \gamma_{4}^{l}~||\mathbf{q}^{l}_{u',v} - \mathbf{q}^{l}_{u,v}||_{2},
    \label{app:eq_gmask_stability}
\end{equation}
where $\gamma_{4}^{l}=C_{\text{SP}}~C_{\text{LN}}^{l}||\mathbf{W}_{2}^{l}||_{2}~||\mathbf{W}_{1}^{l}||_{2}$. Now, for explaining the prediction for node $u$, we can repeat this process for all edges $(u,v) \in \mathcal{E}$ in the neighborhood $\mathcal{N}_{u}$ of node $u$ and generate the matrix $\mathbf{Z}^{l}$. It is to be noted, that the values of the $z^{l}_{u, v}$ elements represent whether a given edge can be dropped or not---an explanation. Further, the composition of multiple Lipschitz continuous functions with Lipschitz constants $\{\mathcal{L}_1, \dots, \mathcal{L}_{\text{L}}\}$ is a new Lipschitz continuous function with $\mathcal{L}_1 \times \dots \times \mathcal{L}_{\text{L}}$ as the Lipschitz constant \citep{gouk2021regularisation}. Using the formulation for a single layer $l$ (Eqn.~\ref{app:eq_gmask_stability}), we can generate a bound for all $L$ layers of the model $f$, where the Lipschitz constant will be: $\prod_{l=1}^{L}\gamma_{4}^{l}$.
\subsubsection{GraphLIME}
\label{app:graphlime_instability}
\xhdr{Setup} 
% For each node prediction, the HSIC Lasso objective function is defined as:
% \begin{equation}
%     \min_{\beta \in \mathbb{R}^{d}} \frac{1}{2} || \mathbf{L} - \sum_{k=1}^{M} \beta_{k} \mathbf{K}^{(k)} ||^{2}_{F} + \rho||\beta||_{1},
%     \label{eq:hsic}
% \end{equation}
% where $||\cdot||_{F}$ is the Frobenius norm, $\rho \geq 0$ is the regularization parameter, $||\cdot||_{1}$ is the $l_{1}$ norm to enforce sparsity, $\mathbf{L}$ is the centered Gram matrix, $L_{ij} = L(y_{i}, y_{j})$ is the kernel for the output labels of the nodes, $\mathbf{K}^{k}$ is the centered gram matrix for the $k$-th feature, and $K_{ij}=K(x_{i}^{(k)}, x_{j}^{(k)})$ is the kernel for the $k$-th dimensional input node features $\mathbf{x}_{u}$.
We use Gaussian kernel for both input and the predictions of all neighbors of node $u$.
\begin{equation}
    \begin{split}
        K(x_{i}^{(k)}, x_{j}^{(k)}) = \exp \big( - \frac{(x_{i}^{(k)}-x_{j}^{(k)})^{2}}{2\sigma_{x}^{2}} \big); 
        L(y_{i}^{(k)}, y_{j}^{(k)}) = \exp \big( - \frac{||y_{i}^{(k)}-y_{j}^{(k)}||^{2}_{2}}{2\sigma_{y}^{2}} \big),
    \end{split}
\end{equation}
The HSIC Lasso objective can be regarded as a minimum redundancy maximum relevancy (mRMR) based feature selection method \citep{peng2005minimum}. Eqn.~\ref{eq:hsic} can be rewritten as:
\begin{equation}
    || \mathbf{L} - \sum_{\mathclap{k=1}}^{\mathclap{M}} \beta_{k} \mathbf{K}^{(k)} ||^{2}_{F} = \text{HSIC}(\mathbf{y}, \mathbf{y}) + \sum_{\mathclap{i=1}}^{\mathclap{M}}\beta_{i}\text{HSIC}(\mathbf{x}_{i}, \mathbf{y}) + \sum_{\mathclap{i,j=1}}^{\mathclap{M}}\beta_{i}\beta_{j}\text{HSIC}(\mathbf{x}_{i}, \mathbf{x}_{j}),
\end{equation}
where HSIC$(\mathbf{x}_{i}, \mathbf{y}) = \text{tr}(\mathbf{K}^{(k)}, \mathbf{L})$ is a kernel-based independence measure called the (empirical) Hilbert-Schmidt independence criterion (HSIC) \citep{gretton2005measuring}.
% The empirical HSIC converges to the true HSIC with $O(1/\sqrt{N})$ (see Theorem 3 in \citep{gretton2005measuring}). 
Further, the HSIC lasso is a convex optimization problem \citep{yamada2014high} and hence given a set of features it will learn feature importance that fit to the predicted labels. We now derive the upper bound for the explanation generated by GraphLIME (or simply GLIME) for the $k$-th node feature. We exclude the sparsity regularizer in our analysis as GLIME enforces sparsity by selecting the top-$P$ features after the optimization.

\xhdr{Theorem 4 (GraphLIME)} \textit{Given the centered Gram matrices for the original and perturbed node features, the instability (Sec.~\ref{sec:def_stable}, Eqn.~\ref{eq:stable}) of explanation $E_{u}$ returned by GraphLIME method can be bounded as:}
\begin{equation}
    ||\beta^{'}_{k} - \beta_{k}||_{F} \leq \gamma_{2}~\cdot~ \text{tr}((\frac{1}{\mathbf{e}^{T}\mathbf{W}^{-1}\mathbf{e}})^{-1}-\mathbf{I}),
\end{equation}
where $\beta^{'}_{k}$ and $\beta_{k}$ are attribute importance generated by GraphLIME for the perturbed and original node features, $\gamma_{2}$ is a noise-independent constant, $\mathbf{e}$ is an all-one vector, and $\mathbf{W}$ is a matrix of the noise terms.

% \xhdr{Theorem 4} \textit{Given the centered Gram matrices for the original and perturbed node attributes, the instability (Sec.~\ref{sec:def_stable}, Eqn.~\ref{eq:stable}) of explanation $E_{u}$ returned by GraphLIME method can be bounded as follows:}
% % \textit{Given the centered Gram matrix for the original and perturbed set of node features, the explanation $\beta_{k}$ for the $k$-th feature of GraphLIME is stable, i.e.,}
% \begin{equation}
%     ||\beta^{'}_{k} - \beta_{k}||_{F} \leq \gamma_{2}~\cdot~ \text{tr}((\frac{1}{\mathbf{e}^{T}\mathbf{W}^{-1}\mathbf{e}})^{-1}-\mathbf{I}),
%     % ||\text{GLIME}(\mathbf{K}^{(k)^{'}}) - \text{GLIME}(\mathbf{K}^{(k)})||_{F} \leq \gamma_{2}~\cdot~ \text{tr}((\frac{1}{\mathbf{e}^{T}\mathbf{W}^{-1}\mathbf{e}})^{-1}-\mathbf{I}),
% \end{equation}
% where $\beta^{'}_{k}$ and $\beta_{k}$ are the feature importance generated by GraphLIME % the $\mathbf{K}^{(k)}$ and $\mathbf{K}^{(k)^{'}}$ are the centered Gram matrix 
% for the original and perturbed node features, $\gamma_{2}$ is a constant independent of the added noise, $\mathbf{e}$ is a vector of all ones, and $\mathbf{W}$ is a matrix comprising of the added noise terms.

\textit{Proof.~} 
% We exclude the sparsity regularizer in our analysis.
% The paper GraphLIME enforces sparsity by selecting the top-$P$ features after the optimization. 
Note, $||Q||_{F}^{2} = \text{tr}(QQ^{T}) = \text{tr}(QQ)$, where $Q$ is a symmetric matrix and tr($\cdot$) is the trace of the matrix. Using this, the objective function can be simplified as:
\begin{align*}
    & \frac{1}{2} || \mathbf{L} - \sum_{i=1}^{M} \beta_{i} \mathbf{K}^{(i)} ||^{2}_{F} \\
    & = \frac{1}{2} \text{tr}\big( (\mathbf{L}-\sum_{i=1}^{M} \beta_{i} \mathbf{K}^{(i)}) \cdot (\mathbf{L}^{T}-\sum_{i=1}^{M} \beta_{i} \mathbf{K}^{(i)^{T}}) \big) \\
    & = \frac{1}{2} \text{tr}\big( (\mathbf{L}-\sum_{i=1}^{M} \beta_{i} \mathbf{K}^{(i)}) \cdot (\mathbf{L}^{T}-\sum_{i=1}^{M} \beta_{i} \mathbf{K}^{(i)}) \big) \\
    & = \frac{1}{2} \text{tr}\big( \mathbf{L}\mathbf{L}^{T}-2\sum_{i=1}^{M} \beta_{i} \mathbf{L}\mathbf{K}^{(i)} + \sum_{i,j=1}^{M} \beta_{i}\beta_{j}\mathbf{K}^{(i)}\mathbf{K}^{(j)} \big)
\end{align*}
We want to minimize the above objective function $O$. We take the partial derivative of $O$ $w.r.t.$ $\beta_{k}$:
% Both $\mathbf{L}$ and $\mathbf{K}^{(k)}$ are symmetric matrices and hence,
% \begin{equation}
%     \frac{\partial{O}}{\partial{\beta_{k}}} = -\mathbf{L}\mathbf{K}^{(k)} + \sum_{j=1}^{d}\beta_{j}\mathbf{K}^{(j)}\mathbf{K}^{(k)} = 0
% \end{equation}
\begin{align*}
    & \frac{\partial{O}}{\partial{\beta_{k}}} = -\mathbf{L}\mathbf{K}^{(k)} + \sum_{j=1}^{M}\beta_{j}\mathbf{K}^{(j)}\mathbf{K}^{(k)} = 0\\
    & \implies -\mathbf{L}\mathbf{K}^{(k)} + \sum_{j=1}^{M}\beta_{j}\mathbf{K}^{(j)}\mathbf{K}^{(k)} = 0\\
    & \implies (-\mathbf{L} + \sum_{j=1}^{M}\beta_{j}\mathbf{K}^{(j)})\mathbf{K}^{(k)} = 0 \\
    & \implies (-\mathbf{L} + \sum_{j=1;j\neq k}^{M}\beta_{j}\mathbf{K}^{(j)} + \beta_{k}\mathbf{K}^{(k)})\mathbf{K}^{(k)} = 0
\end{align*}
We finally get,
\begin{equation}
    \beta_{k}\mathbf{K}^{(k)} = \mathbf{L} - \sum_{j=1;j\neq k}^{M}\beta_{j}\mathbf{K}^{(j)}
\end{equation}
The gram matrix is invertible as it is a full rank symmetric matrix with all the diagonal elements as $1$ (since $K(x_{i},x_{i}) = 1$) and so:
\begin{equation}
    \beta_{k}=\mathbf{\bar{L}}(\mathbf{K}^{(k)})^{-1},
\end{equation}
where $\mathbf{\bar{L}}=\mathbf{L}-\sum_{j=1;j\neq k}^{M}\beta_{j}\mathbf{K}^{(j)}$.

On adding infinitesimal noise $\eta$ to the $k$-th feature, we obtain a new gram matrix given by: $\mathbf{K}^{(k)^{'}} = \mathbf{K}^{(k)}\circ \mathbf{W}$, where $\mathbf{W}\in \mathbb{R}^{M\times M}$ is a function of $\eta$ and $x_{i}$. 
For instance, adding noise $\eta$ to $x_{i}^{(k)}$, we get $K^{'}(x_{i}^{(k)}, x_{j}^{(k)})$ as:
\begin{align*}
    & K^{'}(x_{i}^{(k)}, x_{j}^{(k)}) = \exp \big( - \frac{(x_{i}^{(k)}+\eta-x_{j}^{(k)})^{2}}{2\sigma_{x}^{2}} \big)\\
    & = \exp \big( - \frac{(x_{i}^{(k)}-x_{j}^{(k)})^{2} + \eta^{2} + 2\eta(x_{i}^{(k)}-x_{j}^{(k)})}{2\sigma_{x}^{2}} \big)\\
    & = \exp \big( -\frac{(x_{i}^{(k)}-x_{j}^{(k)})^{2} + 2\eta(x_{i}^{(k)}-x_{j}^{(k)})}{2\sigma_{x}^{2}} \big) \tag{$\eta^{2}=0$ as $\eta$ is infinitesimal noise} \\
    & = \exp \big( -\frac{(x_{i}^{(k)}-x_{j}^{(k)})^{2}}{2\sigma_{x}^{2}} \big)\cdot \exp \big( -\frac{2\eta(x_{i}^{(k)}-x_{j}^{(k)})}{2\sigma_{x}^{2}} \big)
\end{align*}
The importance for the $k$-th node feature is generated by GraphLIME as $\beta_{k}$.
We can now represent the Frobenius norm of the difference between the explanations from GLIME for the original and noisy graph as:
% \begin{eqnarray}
%     ||\text{GLIME}(\mathbf{K}^{(k)^{'}}) - \text{GLIME}(\mathbf{K}^{(k)})||_{F} = ||\mathbf{L}(\mathbf{K}^{(k)^{'}})^{-1} - \mathbf{L}(\mathbf{K}^{(k)})^{-1}||_{F}
% \end{eqnarray}
\begin{align*}
    & ||\beta^{'}_{k} - \beta_{k}||_{F} = ||\mathbf{\bar{L}}(\mathbf{K}^{(k)^{'}})^{-1} - \mathbf{\bar{L}}(\mathbf{K}^{(k)})^{-1}||_{F}\\
    & = ||\mathbf{\bar{L}}\big((\mathbf{K}^{(k)^{'}})^{-1} - (\mathbf{K}^{(k)})^{-1}\big)||_{F}\\
    & \leq ||\mathbf{\bar{L}}||_{F}~||(\mathbf{K}^{(k)^{'}})^{-1}-(\mathbf{K}^{(k)})^{-1}||_{F} \tag{Using Cauchy-Schwartz inequality}\\
    & \leq ||\mathbf{\bar{L}}||_{F}~||(\mathbf{K}^{(k)}\circ \mathbf{W})^{-1}-(\mathbf{K}^{(k)})^{-1}||_{F}\\
    & \leq ||\mathbf{\bar{L}}||_{F}~||(\frac{1}{\mathbf{e}^{T}\mathbf{W}^{-1}\mathbf{e}}\mathbf{K}^{(k)})^{-1}-(\mathbf{K}^{(k)})^{-1}||_{F}, \tag{Using Theorem 3.1 from \citep{reams1999hadamard}}
\end{align*}
where $\mathbf{e}$ is a $m \times 1$ vector of all ones. Note that for using Theorem 3.1 from \citep{reams1999hadamard} we need: (1) $\mathbf{K}^{(k)}$ is positive semidefinite; and (2) $\mathbf{W}$ is positive definite or is
almost positive definite and invertible.
(1) is true by definition as $\mathbf{K}^{(k)}$ is a gram matrix and it is positive semidefinite.
For (2), let us consider a case where $\mathbf{W}\in\mathbb{R}^{2 \times 2}$. Using the Gaussian kernel $\mathbf{K}^{(k)}$, the perturbed gram matrix is:
\begin{align*}
    & \mathbf{K}^{(k)^{'}} = \begin{bmatrix} 1 & \exp \big({-}\frac{(x_{1}^{(k)}{-}x_{2}^{(k)})^{2}}{2\sigma_{x}^{2}} \big) \cdot \exp \big( {-}\frac{2\eta(x_{1}^{(k)}{-}x_{2}^{(k)})}{2\sigma_{x}^{2}} \big)\\
    \exp \big({-}\frac{(x_{1}^{(k)}-x_{2}^{(k)})^{2}}{2\sigma_{x}^{2}} \big) \cdot \exp \big( {-}\frac{2\eta(x_{1}^{(k)}{-}x_{2}^{(k)})}{2\sigma_{x}^{2}} \big) & 1\end{bmatrix}\\
    & = \begin{bmatrix} 1 & \exp \big( {-}\frac{(x_{1}^{(k)}{-}x_{2}^{(k)})^{2}}{2\sigma_{x}^{2}} \big)\\
    \exp \big({-}\frac{(x_{1}^{(k)}{-}x_{2}^{(k)})^{2}}{2\sigma_{x}^{2}} \big) & 1\end{bmatrix} \circ \begin{bmatrix} 1 & \exp \big( {-}\frac{2\eta(x_{1}^{(k)}{-}x_{2}^{(k)})}{2\sigma_{x}^{2}} \big)\\
    \exp \big({-}\frac{2\eta(x_{1}^{(k)}{-}x_{2}^{(k)})}{2\sigma_{x}^{2}} \big) & 1\end{bmatrix}
\end{align*}
Hence, $\mathbf{W}$ is:
\begin{equation}
    \mathbf{W} = \begin{bmatrix} 1 & \exp \big( {-}\frac{2\eta(x_{1}^{(k)}{-}x_{2}^{(k)})}{2\sigma_{x}^{2}} \big)\\
    \exp \big( {-}\frac{2\eta(x_{1}^{(k)}{-}x_{2}^{(k)})}{2\sigma_{x}^{2}} \big) & 1\end{bmatrix}
\end{equation}
For positive definite, we need to show $b^{T}\mathbf{W}b > 0$ for any non-zero vector $b$.
All the elements of $\mathbf{W}$ are positive and using a $b$ vector with all ones will result in $b^{T}\mathbf{W}b > 0$. Hence, $\mathbf{W}$ is positive definite. 
Now, both $\mathbf{W}$ and $\mathbf{K}^{(k)}$ are invertible matrix so we can use the property $(AB)^{-1}=B^{-1}A^{-1}$. Putting everything together we get:
\begin{align*}
        & ||\beta^{'}_{k} - \beta_{k}||_{F} \leq ||\mathbf{\bar{L}}||_{F}~||(\mathbf{K}^{(k)})^{-1}(\frac{1}{\mathbf{e}^{T}\mathbf{W}^{-1}\mathbf{e}})^{-1}-(\mathbf{K}^{(k)})^{-1}||_{F}\\
        & \leq ||\mathbf{\bar{L}}||_{F}~||(\mathbf{K}^{(k)})^{-1}\big((\frac{1}{\mathbf{e}^{T}\mathbf{W}^{-1}\mathbf{e}})^{-1}-\mathbf{I}\big)||_{F}\\
        & \leq ||\mathbf{\bar{L}}||_{F}~||(\mathbf{K}^{(k)})^{-1}||_{F}~||(\frac{1}{\mathbf{e}^{T}\mathbf{W}^{-1}\mathbf{e}})^{-1}-\mathbf{I}||_{F} \tag{Using Cauchy-Schwartz inequality}\\
        & \leq \text{tr}(\mathbf{\bar{L}})~\text{tr}((\mathbf{K}^{(k)})^{-1})~\text{tr}((\frac{1}{\mathbf{e}^{T}\mathbf{W}^{-1}\mathbf{e}})^{-1}-\mathbf{I})\\
        & \leq \gamma_{2}~\cdot~ \text{tr}((\frac{1}{\mathbf{e}^{T}\mathbf{W}^{-1}\mathbf{e}})^{-1}-\mathbf{I}),
        % & \leq \sqrt{\text{rk}(\mathbf{\bar{L}})}~||\mathbf{\bar{L}}||_{2}~\sqrt{\text{rk}((\mathbf{K}^{(k)})^{-1})}~||(\mathbf{K}^{(k)})^{-1}||_{2}
\end{align*}
where $\gamma_{2}=\text{tr}(\mathbf{\bar{L}})~\text{tr}((\mathbf{K}^{(k)})^{-1})$ is a constant independent of the added noise.
\subsection{Analyzing Fairness of GNN Explanation Methods}
\label{app:lg_unfairness_theory}
% \subsubsection{Local-Group Unfairness}
% \label{app:lg_unfairness_theory}
\xhdr{Theorem 8} \textit{Given a node $u$, a sensitive feature $s$, and a set $\mathcal{K}$ comprising of $u$ and its perturbations, the group fairness mismatch (Sec.~\ref{sec:def_fair}, Eqn.~\ref{eq:group_fair}) of an explanation $E_{u}$ can be bounded as follows:}
\begin{equation}
    |~\text{SP}(\mathbf{\hat{y}}_{\mathcal{K}}){-}\text{SP}(\mathbf{\hat{y}}_{\mathcal{K}}^{E_u})~| \leq \sum_{\mathclap{s \in \{0, 1\}}}|\text{Err}_{D_{s}}(f(t(E_{u},\mathcal{G}_{u'})){-}f(\mathcal{G}_{u'}))|, \nonumber
    %\label{eq:lg_unfairness}
\end{equation}
% \marinka{Chirag, see Slack. $D$ vs $D_s$. }
where $\text{SP}(\mathbf{\hat{y}}_{\mathcal{K}})$ and $\text{SP}(\mathbf{\hat{y}}^{E_{u}}_{\mathcal{K}})$ are statistical parity estimates, $D$ is the joint distribution over node features $\mathbf{x}_{u'}$ in $\mathcal{G}_{u'}$ and their respective labels $\mathbf{y}_{u'}$ for $\forall u' \in \mathcal{K}$, $D_{s}$ is $D$ conditioned on the value of the sensitive feature $s$, and $\text{Err}_{D_{s}}(\cdot)$ is the model error under $D_{s}$.

% \xhdr{Theorem 8} 
% \textit{Given a node $u$, a sensitive attribute $s$, and a set $\mathcal{K}$ comprising of node $u$ and its perturbations, the group fairness mismatch (Sec.~\ref{sec:def_fair}, Eqn.~\ref{eq:group_fair}) of an explanation $E_{u}$ corresponding to node $u$ can be bounded as follows:}
% % \textit{Given node $u$, its sensitive attribute $s$, and a set of $\mathcal{K}$ perturbations, the explanation for node $u$ satisfies group fairness, i.e.,}
% \begin{equation}
%     |~\text{SP}(\mathbf{\hat{y}}_{\mathcal{K}}) - \text{SP}(\mathbf{\hat{y}}_{\mathcal{K}}^{E_u})~| \leq \sum_{\mathclap{s \in \{0, 1\}}}|\text{Err}_{D_{s}}(f(t(E_{u},\mathcal{G}_{u'})){-}f(\mathcal{G}_{u'}))|,
% \end{equation}
% % \begin{equation}
% %     |\text{SP}(\mathbf{\hat{y}}^{E}_{u'}){-}\text{SP}(\mathbf{\hat{y}}_{u'})| \leq \sum_{s \in \{0, 1\}}|\text{Err}_{D_{s}}(f(t(E_{u},\mathcal{G}_{u'})){-}f(\mathcal{G}_{u'}))|
% % \end{equation}
% where $\text{SP}(\mathbf{\hat{y}}_{\mathcal{K}})$ and $\text{SP}(\mathbf{\hat{y}}^{E_{u}}_{\mathcal{K}})$ are statistical parity estimates as defined in Sec.~\ref{sec:def_fair}, $D$ is the joint distribution over the node attributes $\mathbf{x}_{u'}$ in graph $\mathcal{G}_{u'}$ and their respective labels $\mathbf{y}_{u'}$ for all nodes $u' \in \mathcal{K}$, $D_{s}$ is the conditional distribution of $D$ given a particular value of the sensitive attribute $s$, and $\text{Err}_{D}(\cdot)=\mathbb{E}_{D}[\mathbf{y}_{u'}-f(\mathcal{G}_{u'})]$ is the error of the model $f$ under the joint distribution $D$.

\textit{Proof.~} For group fairness, we define the total variation divergence $d_{\text{TV}}$ to measure the difference between two probability distributions, i.e., $\mathbf{\hat{y}}_{\mathcal{K}}$ and $\mathbf{\hat{y}}^{E_{u}}_{\mathcal{K}}$. For a given binary sensitive attribute $s \in \{0, 1\}$, we can write:~ $d_{\text{TV}}(D_{s}(\mathbf{y}_{u'}), D_{s}(\mathbf{\hat{y}}_{\mathcal{K}})) \leq \mathbb{E}_{D_{s}}[|\mathbf{y}_{u'}-f(\mathcal{G}_{u'})|]$ \citep{zhao2019inherent}.
The total variation divergence is symmetrical and satisfies triangle inequality. Hence, we have:
\begin{align*}
    \begin{split}
        d_{\text{TV}}(D_{0}(\mathbf{y}_{u'}), D_{1}(\mathbf{y}_{u'})) \leq d_{\text{TV}}(D_{0}(\mathbf{y}_{u'}), D_{0}(\mathbf{\hat{y}}_{\mathcal{K}})){+}d_{\text{TV}}(D_{0}(\mathbf{\hat{y}}_{\mathcal{K}}), D_{1}(\mathbf{\hat{y}}_{\mathcal{K}})){+}\\d_{\text{TV}}(D_{1}(\mathbf{y}_{u'}), D_{1}(\mathbf{\hat{y}}_{\mathcal{K}}))
    \end{split}
\end{align*}
% \mathbf{\hat{y}}_{\mathcal{K}}^{E_u}
Now, the middle term in the right side of the inequality is the statistical parity (SP) for the binary sensitive attribute and therefore we can simplify the equation as:
\begin{equation}
    d_{\text{TV}}(D_{0}(\mathbf{y}_{u'}), D_{1}(\mathbf{y}_{u'})) \leq d_{\text{TV}}(D_{0}(\mathbf{y}_{u'}), D_{0}(\mathbf{\hat{y}}_{\mathcal{K}})){+}\text{SP}(\mathbf{\hat{y}}_{\mathcal{K}}){+}\\d_{\text{TV}}(D_{1}(\mathbf{y}_{u'}), D_{1}(\mathbf{\hat{y}}_{\mathcal{K}}))
    \label{app:eq_dtv_original}
\end{equation}
Similarly, using the explanation $E$, we can write a similar inequality of the group predictions $\mathbf{\hat{y}}_{\mathcal{K}}^{E_u}$ and equate the left hand side terms to get:
\begin{equation}
    \begin{split}
        d_{\text{TV}}(D_{0}(\mathbf{y}_{u'}), D_{0}(\mathbf{\hat{y}}_{\mathcal{K}})){+}\text{SP}(\mathbf{\hat{y}}_{\mathcal{K}}){+}d_{\text{TV}}(D_{1}(\mathbf{y}_{u'}), D_{1}(\mathbf{\hat{y}}_{\mathcal{K}})) =\\ d_{\text{TV}}(D_{0}(\mathbf{y}_{u'}), D_{0}(\mathbf{\hat{y}}_{\mathcal{K}}^{E_u})){+}\text{SP}(\mathbf{\hat{y}}_{\mathcal{K}}^{E_u}){+}d_{\text{TV}}(D_{1}(\mathbf{y}_{u'}), D_{1}(\mathbf{\hat{y}}_{\mathcal{K}}^{E_u}))
        % d_{\text{TV}}(D_{0}(\mathbf{y}_{u'}), D_{0}(\mathbf{\hat{y}}_{u'})){+}\text{SP}(\mathbf{\hat{y}}_{u'}){+}d_{\text{TV}}(D_{1}(\mathbf{y}_{u'}), D_{1}(\mathbf{\hat{y}}_{u'})) = d_{\text{TV}}(D_{0}(\mathbf{y}_{u'}), D_{0}(\mathbf{\hat{y}}^{E}_{u'}))\\{+}\text{SP}(\mathbf{\hat{y}}^{E}_{u'}){+}d_{\text{TV}}(D_{1}(\mathbf{y}_{u'}), D_{1}(\mathbf{\hat{y}}^{E}_{u'}))
    \end{split}
    \label{app:eq_lg_fairness_equate}
\end{equation}
Using Lemma 3.1 of \cite{zhao2019inherent}, we can write $d_{\text{TV}}(D_{s}(\mathbf{y}_{u'}), D_{s}(\mathbf{\hat{y}}_{\mathcal{K}})) \leq \text{Err}_{D_{s}}(f(\mathcal{G}_{u'}))$. Further simplification of Eqn.~\ref{app:eq_lg_fairness_equate} and plugging the lemma, we get:
\begin{eqnarray}
    \begin{split}
        \text{SP}(\mathbf{\hat{y}}_{\mathcal{K}}^{E_u}){-}\text{SP}(\mathbf{\hat{y}}_{\mathcal{K}}){=} d_{\text{TV}}(D_{0}(\mathbf{y}_{u'}), D_{0}(\mathbf{\hat{y}}_{\mathcal{K}}^{E_u})){-}d_{\text{TV}}(D_{0}(\mathbf{y}_{u'}), D_{0}(\mathbf{\hat{y}}_{\mathcal{K}}))\\{+}d_{\text{TV}}(D_{1}(\mathbf{y}_{u'}), D_{1}(\mathbf{\hat{y}}_{\mathcal{K}}^{E_u})){-}d_{\text{TV}}(D_{1}(\mathbf{y}_{u'}), D_{1}(\mathbf{\hat{y}}_{\mathcal{K}}))
        % \text{SP}(\mathbf{\hat{y}}^{E}_{u'}){-}\text{SP}(\mathbf{\hat{y}}_{u'}){=} d_{\text{TV}}(D_{0}(\mathbf{y}_{u'}), D_{0}(\mathbf{\hat{y}}^{E}_{u'})){-}d_{\text{TV}}(D_{0}(\mathbf{y}_{u'}), D_{0}(\mathbf{\hat{y}}_{u'}))\\{+}d_{\text{TV}}(D_{1}(\mathbf{y}_{u'}), D_{1}(\mathbf{\hat{y}}^{E}_{u'})){-}d_{\text{TV}}(D_{1}(\mathbf{y}_{u'}), D_{1}(\mathbf{\hat{y}}_{u'}))
    \end{split}\\
    \begin{split}
        \text{SP}(\mathbf{\hat{y}}_{\mathcal{K}}^{E_u}){-}\text{SP}(\mathbf{\hat{y}}_{\mathcal{K}})\leq \text{Err}_{D_{0}}(f(t(E_{u},\mathcal{G}_{u'}))){-}\text{Err}_{D_{0}}(f(\mathcal{G}_{u'})){+}\text{Err}_{D_{1}}(f(t(E_{u},\mathcal{G}_{u'}))){-}\\\text{Err}_{D_{1}}(f(\mathcal{G}_{u'}))
        % \text{SP}(\mathbf{\hat{y}}^{E}_{u'}){-}\text{SP}(\mathbf{\hat{y}}_{u'})\leq \text{Err}_{D_{0}}(f(t(E_{u},\mathcal{G}_{u'}))){-}\text{Err}_{D_{0}}(f(\mathcal{G}_{u'})){+}\text{Err}_{D_{1}}(f(t(E_{u},\mathcal{G}_{u'}))){-}\\\text{Err}_{D_{1}}(f(\mathcal{G}_{u'}))
    \end{split}\\
    \text{SP}(\mathbf{\hat{y}}_{\mathcal{K}}^{E_u}){-}\text{SP}(\mathbf{\hat{y}}_{\mathcal{K}})\leq \text{Err}_{D_{0}}(f(t(E_{u},\mathcal{G}_{u'})){-}f(\mathcal{G}_{u'})){+}\text{Err}_{D_{1}}(f(t(E_{u},\mathcal{G}_{u'})){-}f(\mathcal{G}_{u'}))
\end{eqnarray}
Taking the absolute value as norms on both sides, we get:
\begin{align*}
    & |\text{SP}(\mathbf{\hat{y}}_{\mathcal{K}}^{E_u}){-}\text{SP}(\mathbf{\hat{y}}_{\mathcal{K}})|\leq |\text{Err}_{D_{0}}(f(t(E_{u},\mathcal{G}_{u'})){-}f(\mathcal{G}_{u'})){+}\text{Err}_{D_{1}}(f(t(E_{u},\mathcal{G}_{u'})){-}f(\mathcal{G}_{u'}))|\\
    & |\text{SP}(\mathbf{\hat{y}}_{\mathcal{K}}^{E_u}){-}\text{SP}(\mathbf{\hat{y}}_{\mathcal{K}})| \leq |\text{Err}_{D_{0}}(f(t(E_{u},\mathcal{G}_{u'})){-}f(\mathcal{G}_{u'}))|+|\text{Err}_{D_{1}}(f(t(E_{u},\mathcal{G}_{u'})){-}f(\mathcal{G}_{u'}))| \tag{Using triangle inequality} \\
    & |\text{SP}(\mathbf{\hat{y}}_{\mathcal{K}}){-}\text{SP}(\mathbf{\hat{y}}_{\mathcal{K}}^{E_u})| \leq \sum_{\mathclap{s \in \{0, 1\}}}|\text{Err}_{D_{s}}(f(t(E_{u},\mathcal{G}_{u'})){-}f(\mathcal{G}_{u'}))|
\end{align*}
\begin{figure*}[t]
    \centering
    \includegraphics[width=\linewidth]{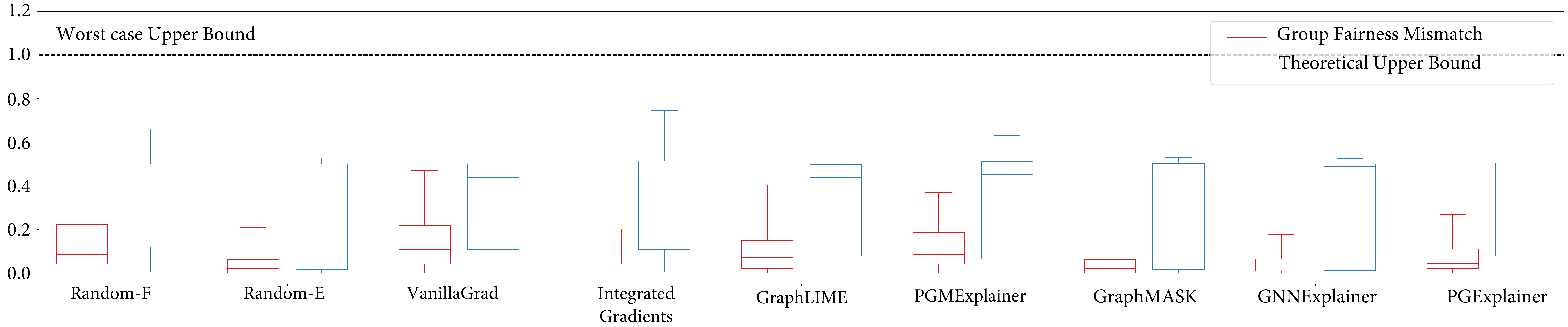}
    \caption{
    The empirically calculated group fairness mismatch measure (in red) and our theoretical upper bounds for group fairness mismatch (in blue) for nine \expmethods.
    Results show that no \expmethod violate the group fairness bounds for the German credit graph dataset.
    % Theoretical bounds for local-group unfairness for eight \expmethods. 
    % The local-group unfairness (left inequality term in Eqn.~\ref{eq:lg_unfairness}) is always lower than the upper bound (right inequality term in Eqn.~\ref{eq:lg_unfairness}).
    }
    \label{app:fig_empirical_bound_fair}
\end{figure*}
\section{Experiments}
\label{app:sec_experiment}
\xhdr{Datasets} 
We experiment with nine datasets in this work:

1) \textit{German credit graph}~\citep{agarwal2021towards} has 1,000 nodes representing customers in a German bank, connected based on similarity of their credit applications.
The task is to classify clients into good vs.~bad credit risks considering clients' gender as the sensitive attribute.

2) \textit{Recidivism graph}~\citep{agarwal2021towards} has 18,876 nodes representing defendants released on bail that are connected based on similarity of their past criminal records and demographics. 
The task is to classify defendants into bail (i.e., unlikely to commit a violent crime if released) vs.~no bail (i.e., likely to commit a violent crime) considering race information as the protected attribute.

3) \textit{Credit defaulter graph}~\citep{agarwal2021towards} has 30,000 nodes representing individuals connected based on payment behaviors and demographics. 
The task is to classify individuals into credit card payment vs.~no payment on time considering age as the sensitive attribute.

4) \textit{Cora} dataset \citep{mccallum2000automating} comprises of 2708 nodes representing scientific publications  classified into one of seven classes. The data contains bibliographic records of machine learning papers that have been manually clustered into classes that refer to the same publication. The citation network consists of 5429 links. Each publication in the dataset is described by a 0/1-valued word vector indicating the absence/presence of the corresponding word from the dictionary. The dictionary consists of 1433 unique words.

5) \textit{PubMed} dataset \citep{sen2008collective} consists of 19717 nodes representing scientific publications from PubMed database pertaining to diabetes classified into one of three classes. The citation network consists of 44338 links. Each publication in the dataset is described by a TF/IDF weighted word vector from a dictionary which consists of 500 unique words.

6) \textit{Citeseer} dataset \citep{giles1998citeseer} consists of 3312 nodes representing scientific publications classified into one of six classes. The citation network consists of 4732 links. Each publication in the dataset is described by a 0/1-valued word vector indicating the absence/presence of the corresponding word from the dictionary. The dictionary consists of 3703 unique words.

7) \textit{Ogbn-mag} dataset~\citep{hu2020open} is a heterogeneous graph composed of a subset of the Microsoft Academic Graph (MAG)~\citep{wang2020microsoft}. It contains four types of entities---papers (736,389 nodes), authors (1,134,649 nodes), institutions (8,740 nodes), and fields of study (59,965 nodes)---as well as four types of directed relations connecting two types of entities—an author is “affiliated with” an institution, an author “writes” a paper, a paper “cites” a paper, and a paper “has a topic of” a field of study. The task is to classify papers into 349 venues (conference or journal), given its content, references, authors, and authors’ affiliations.

\looseness=-1
8) \textit{Ogbn-arxiv} citation graph~\citep{hu2020open} has 169,343 nodes representing CS arXiv papers linked based on who cites whom patterns. The task is to classify papers into 40 thematic categories, \eg cs.AI, cs.LG, and cs.OS.

9) \textit{MUTAG} dataset~\citep{debnath1991structure} contains 4,337 graphs representing chemical compounds where nodes represent different atoms and edges represent chemical bonds. The graphs are labeled into two different classes according to their mutagenic effect on the Gram-negative bacterium \textit{S. typhimuriuma}.

The training and testing splits for the \textit{German credit graph}, \textit{Recidivism graph}, and \textit{Credit defaulter graph} dataset is setup following the codes released by \citep{agarwal2021towards}. For the \textit{Cora}, \textit{PubMed}, \textit{Citeseer}, \textit{Ogbn-mag}, and \textit{Ogbn-arxiv}, we use the training and testing data loader\footnote{\url{https://ogb.stanford.edu/docs/home/}} provided by \cite{hu2020open}. Finally, for \textit{MUTAG} dataset we used the training and testing splits following \citep{Fey/Lenssen/2019}. All datasets used in this work are publicly available and are accordingly cited.

\looseness=-1
\xhdr{Implementation details} All codes and datasets are available at \textcolor{blue!50}{\url{https://anonymous.4open.science/r/GNNExEval-CC00/}}. We use a mutli-layer GraphSAGE model as our GNN predictor $f$ for all node classification tasks.
% encoder and set the hidden dimensionality to 16. The encoder is followed by a two-layer MLP projection head [Chen et al., 2020]. We only use ReLU and BatchNormalization (BN) layers after the first hidden layer in the MLP. For both the MLP layers, we set the hidden dimensionality to 16.
For the \textit{German credit graph}, \textit{Recidivism graph}, and \textit{Credit defaulter graph} datasets, we follow \citep{agarwal2021towards} and design a model comprising of two GraphSAGE convolution layers with ReLU non-linear activation function and a fully-connected linear classification layer with Softmax activations. The hidden dimensionality of the layers is set to $16$. The same configuration was also used for \textit{Cora}, \textit{PubMed} and \textit{Citeseer} datasets.

For the \textit{Ogbn-arxiv} dataset, we follow \citet{hu2020open} and design a model comprising of three GraphSAGE convolution layers with ReLU non-linear activation function and a fully-connected linear classification layer with Softmax activations. The hidden dimensionality of the layers is set to $256$. Similarly, for \textit{Ogbn-mag} dataset, we design a model comprising of three GraphSAGE convolution layers with ReLU non-linear activation function for first two layers and Softmax for the final layer. The hidden dimensionality of the layers is set to $192$.

For \textit{MUTAG} dataset, we follow \cite{Fey/Lenssen/2019} and design a model comprising of three GCN convolution layers with ReLU non-linear activation function, global mean pooling layer and a fully-connected linear classification layer with Softmax activations. The hidden dimensionality of the layers is set to $16$.

Finally, for the link prediction task using \textit{Cora} dataset, we use a Graph AutoEncoder model with two GraphSAGE convolution layers with ReLU activation as the encoder model and an InnerProduct layer~\citep{torchgeo} as the decoder. The hidden dimensionality of the layers is set to 16. Table~\ref{app:tab_dataset} details the performance of the above models on their respective tasks.
% \chirag{WILL ADD A TABLE WITH TEST ACCURACY FOR ALL DATASETS}
\begin{table*}[ht]
\centering\small
\caption{Statistics of all nine graph datasets used for node, link, and graph prediction with the GNN's testing accuracy measured using all nodes in the test split. For the MUTAG graph classification dataset, we detail the mean nodes, edges, and degrees across all the molecules in the test split, and for the Cora-link dataset we report the AUROC for the link prediction task on the test split.}
\label{app:tab_dataset}
\setlength{\tabcolsep}{2pt}
% \renewcommand{\arraystretch}{0.9}
% \vspace{-1mm}
% {\begin{tabular}{cl|cccc}
{\begin{tabular}{lcccccccc}
Datasets & Nodes & Edges & Node features & Sensitive Attribute & Classes & Accuracy \\
 \toprule
German credit & 1,000 & 22,242 & 27 & Gender & 2 & 70.43\% \\
Recidivism & 18,876 & 321,308 & 18 & Race & 2 & 92.68\% \\
Credit defaulter & 30,000 & 1,436,858 & 13 & Age & 2 & 70.69\% \\
Cora & 2,708 & 5,429 & 1,433 & \naapp & 7 & 85.78\% \\
PubMed & 19,717 & 44,338 & 500 & \naapp & 3 & 93.77\%\\
Citeseer & 3,327 & 4,732 & 3,703 & \naapp & 6 & 82.69\% \\
Ogbn-mag & 1,939,743 & 21,111,007 & 128 & \naapp & 349 & 37.03\% \\
Ogbn-arxiv & 169,343 & 1,166,243 & 128 & \naapp & 40 & 66.13\% \\
MUTAG & 19.79 & 17.93 & 7 & \naapp & 2 & 76.32\% \\
Cora-link & 2,708 & 5,429 & 1,433 & \naapp & 7 & 97.00\% \\
%  & & \multicolumn{4}{c}{Evaluation metrics} \\
% \multirow{2}{*}{} & \multirow{2}{*}{Method} & \multirow{2}{*}{Unfaithfulness ($\downarrow$)} & \multirow{2}{*}{Instability ($\downarrow$)} & \multicolumn{2}{c}{Fairness Mismatch ($\downarrow$)} \\
%  & & &  & \makecell{Counterfactual} & \makecell{Group} \\
% \multirow{1}{2cm}{MUTAG} &
% \begin{tabular}[l]{@{}l@{}}\ccg{Random Node Features}\\\ccg{Random Edges}\\\ccy{{VanillaGrads}}\\\ccy{Integrated Gradients}\\\end{tabular}&
% \begin{tabular}[c]{@{}c@{}}{{0.014}\std{0.001}}\\{{0.018}\std{0.001}}\\{{0.015}\std{0.001}}\\{{0.015}\std{0.001}}\end{tabular} &
% \begin{tabular}[c]{@{}c@{}}{{0.048}\std{0.001}}\\{0.346\std{0.004}}\\{{0.852}\std{0.023}}\\{{0.624}\std{0.026}}\end{tabular} & 
% \begin{tabular}[c]{@{}c@{}}{\na}\\{\na}\\{\na}\\{\na}\end{tabular} &
% \begin{tabular}[c]{@{}c@{}}{\na}\\{\na}\\{\na}\\{\na}\end{tabular} \\
\bottomrule
\end{tabular}}
\end{table*}

\xhdr{Compute details} We use a Intel(R) Xeon(R) CPU E5-2680 with 250Gb RAM and a single NVIDIA Tesla M40 GPU for all our experiments.

\xhdr{Hyperparameters} For all experiments, we use normal Gaussian noise $\mathcal{N}(0,1)$ for perturbing node attributes and set the probability of perturbing an attribute dimension to 0.1. For training GraphSAGE, we use an Adam optimizer with a learning rate of $1\times10^{-3}$, weight decay of $1\times10^{-5}$, and the number of epochs to 1000. For the GNN explanation methods, all hyperparameters are set following the authors’ guidelines.
% Now, we discuss the empirical and theoretical bounds for stability, counterfactual fairness, and group fairness of GNN \expmethods.
\subsection{Results}
\begin{figure*}[t]
    \centering
    \begin{subfigure}[b]{0.32\textwidth}
        \includegraphics[width=\linewidth]{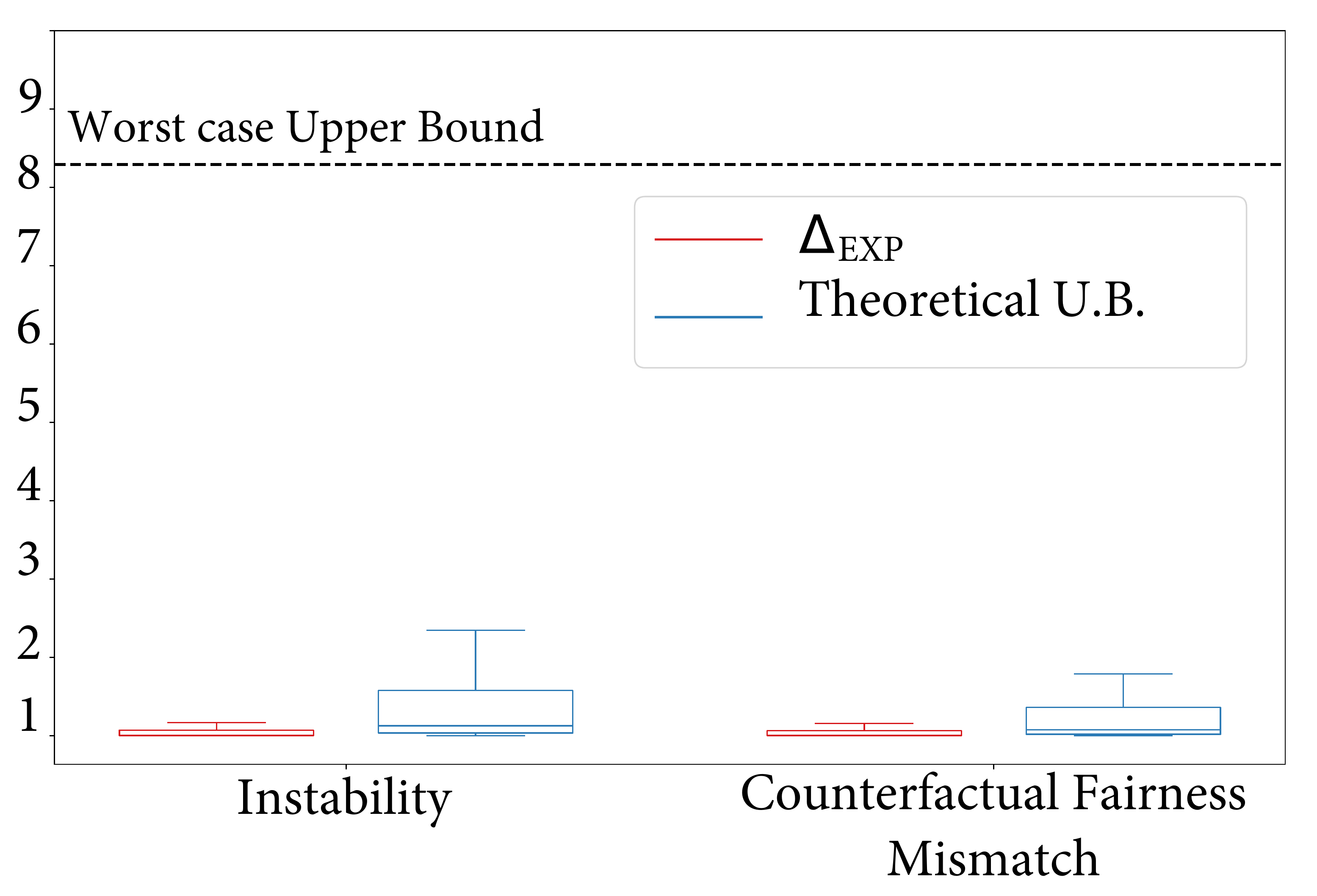}
        \caption{VanillaGrads}
        \label{app:fig_stab_count_grad_bound}
    \end{subfigure}
    \begin{subfigure}[b]{0.32\textwidth}
        \includegraphics[width=\linewidth]{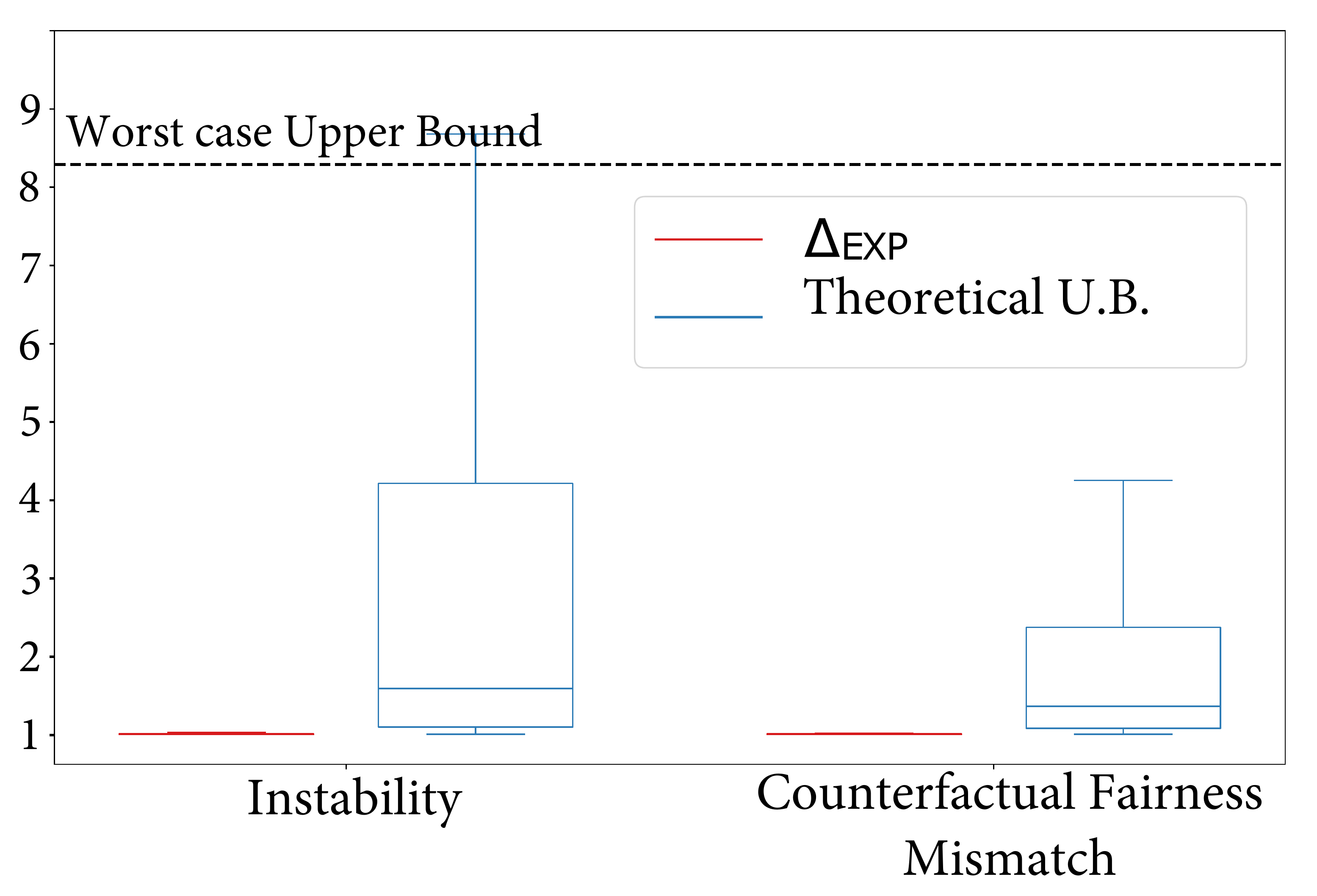}
        \caption{GraphLIME}
        \label{app:fig_stab_count_graphlime_bound}
    \end{subfigure}
    \begin{subfigure}[b]{0.33\textwidth}
        \includegraphics[width=\linewidth]{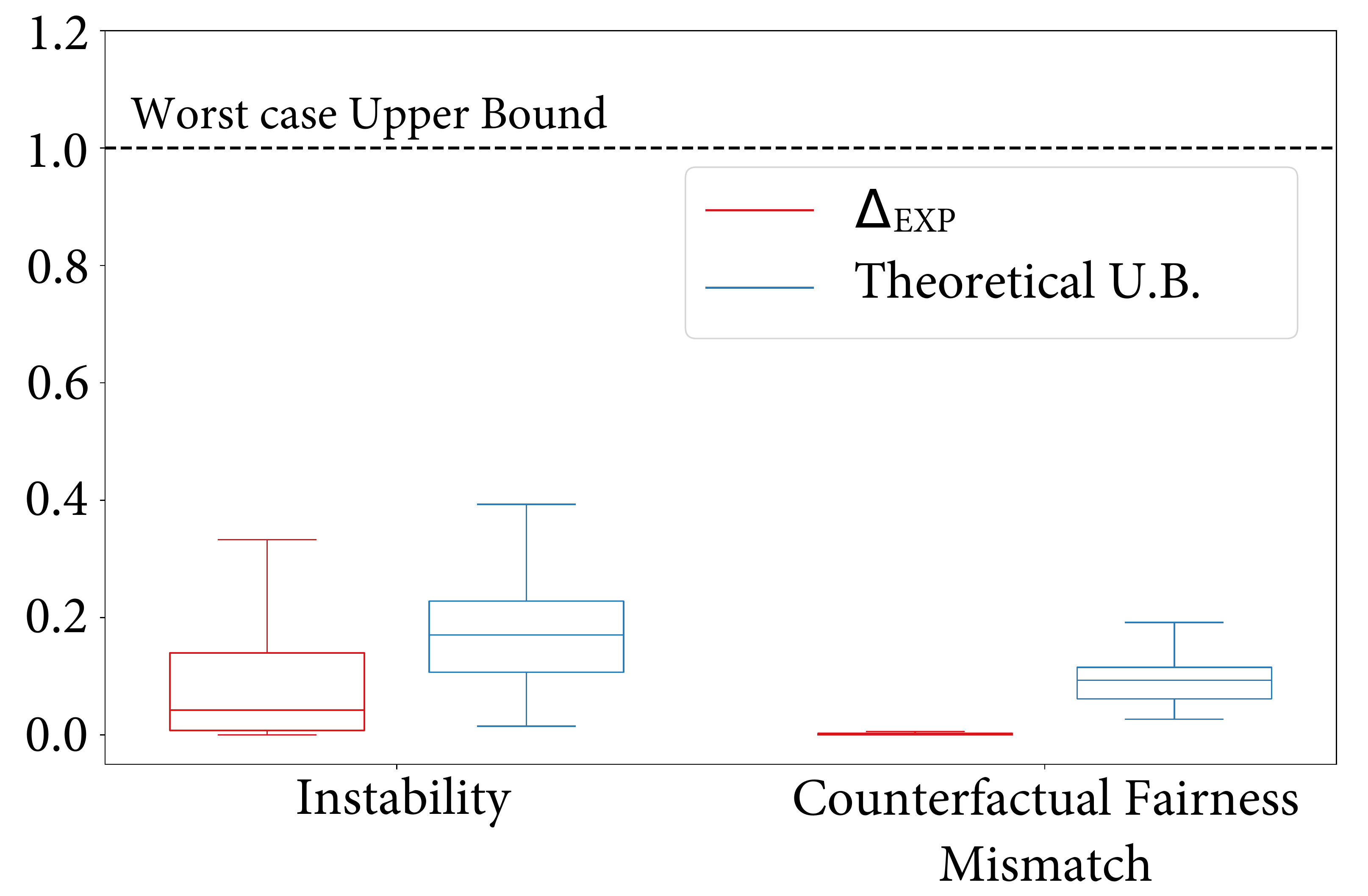}
        \caption{GraphMASK}
        \label{app:fig_stab_count_graphmask_bound}
    \end{subfigure}
    \caption{The theoretical upper bounds (in blue) for the instability and counterfactual fairness mismatch metric for (a) VanillaGrad, (b) GraphLIME, and (c) GraphMASK \expmethod. 
    Results across both properties show that the empirically calculated explanation differences $\Delta_{\text{EXP}}$ (in red) do not violate our theoretical bounds when evaluated on the German credit graph dataset.}
    \label{app:fig_stab_count_bound}
\end{figure*}

\xhdr{Empirically verifying our theoretical bounds} We compare the reliability of \expmethods by computing the faithfulness, stability, counterfactual, and group fairness metrics as described in Sec.~\ref{sec:exp}. The empirical and theoretical bounds for group fairness (Fig.~\ref{app:fig_empirical_bound_fair}) shows that no bounds were violated in our experiments. We observe consistent trend between empirically computed group fairness mismatch and theoretical bounds for all nine \expmethods, with the empirical values always lower than our theoretical upper bounds. Further, we compare the empirical and theoretical bounds for stability and counterfactual fairness for the three representative \expmethods: VanillaGrads (Fig.~\ref{app:fig_stab_count_grad_bound}), GraphLIME (Fig.~\ref{app:fig_stab_count_graphlime_bound}), and GraphMASK (Fig.~\ref{app:fig_stab_count_graphmask_bound}).
Across all \expmethods, the theoretical bounds are well below the worst case upper bound with only some outlier points for stability in GraphLIME. Despite that, the median (horizontal line inside each box in Fig.~\ref{app:fig_stab_count_bound}) of the theoretical bounds are an order of magnitude smaller than that provided by the worst case upper bound.
% \xhdr{Q2) Surrogate model based methods generate most reliable explanations}
% Results in Table~\ref{app:tab_node_metric_1}-\ref{app:tab_node_metric_2} across eight datasets show that surrogate model based \expmethods tend to produce more reliable explanations than gradient and perturbation based methods.
% Specifically, all nine \expmethods are highly unstable across eight datasets suggesting the need for \expmethods that are stable to infinitesimally small noise perturbations. 
% \xhdr{Q4) Metrics generalize to 9 explanation methods, 9 datasets, and 3 downstream tasks} 
% \begin{figure*}[t]
%     \centering\small
% % 	\begin{flushleft}
% % 	    \hspace{1.2cm}Link Prediction on Cora dataset
% % 	    \hspace{2.7cm}Graph Classification on MUTAG dataset
% % 	\end{flushleft}
%     \begin{subfigure}[b]{0.43\textwidth}
%         \includegraphics[width=\linewidth]{iclr2022/FIG/link_faith.pdf}
%         \caption{Unfaithfulness: Link Prediction on Cora}
%         \label{fig:faith_link}
%     \end{subfigure}
%     \begin{subfigure}[b]{0.56\textwidth}
%         \includegraphics[width=\linewidth]{iclr2022/FIG/graph_faith.pdf}
%         \caption{Unfaithfulness: Graph Classification on MUTAG}
%         \label{fig:faith_graph}
%     \end{subfigure}    
%     \caption{Systematic evaluation of GNN \expmethods (random strategies (in grey), gradient methods (in yellow), perturbation method (in red), and surrogate method (in blue)) for link prediction (Column a-b) and graph classification (Column c-d) tasks. Shown are box plots of unfaithfulness scores of test set samples. For both tasks, random strategies outperform other GNN \expmethods on unfaithfulness.}
%     \label{app:fig_link_graph}
% \end{figure*}
\begin{figure}[ht]
    \centering
    \begin{subfigure}[b]{0.24\textwidth}
        \includegraphics[width=\linewidth]{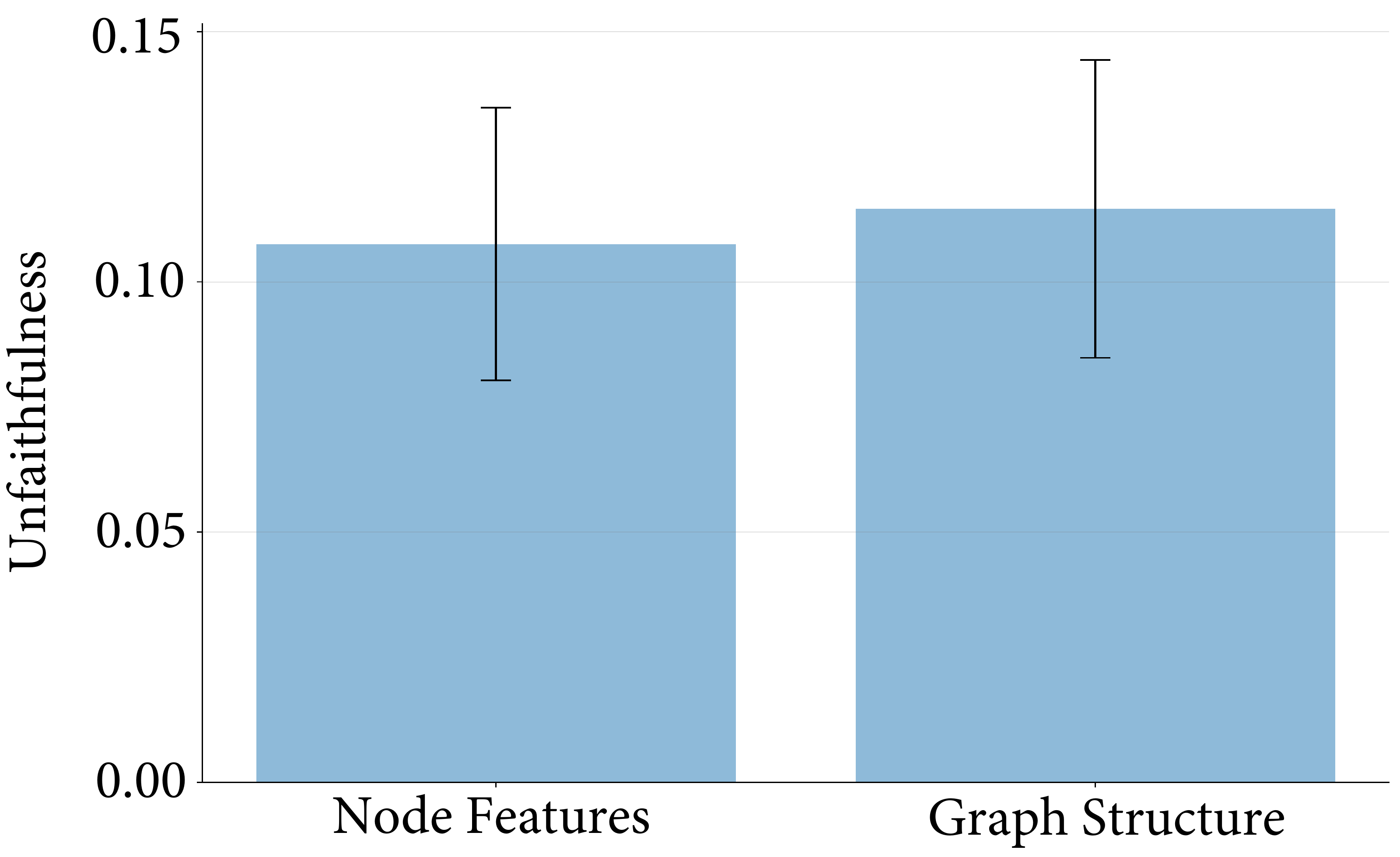}
    \end{subfigure}
    \begin{subfigure}[b]{0.24\textwidth}
        \includegraphics[width=\linewidth]{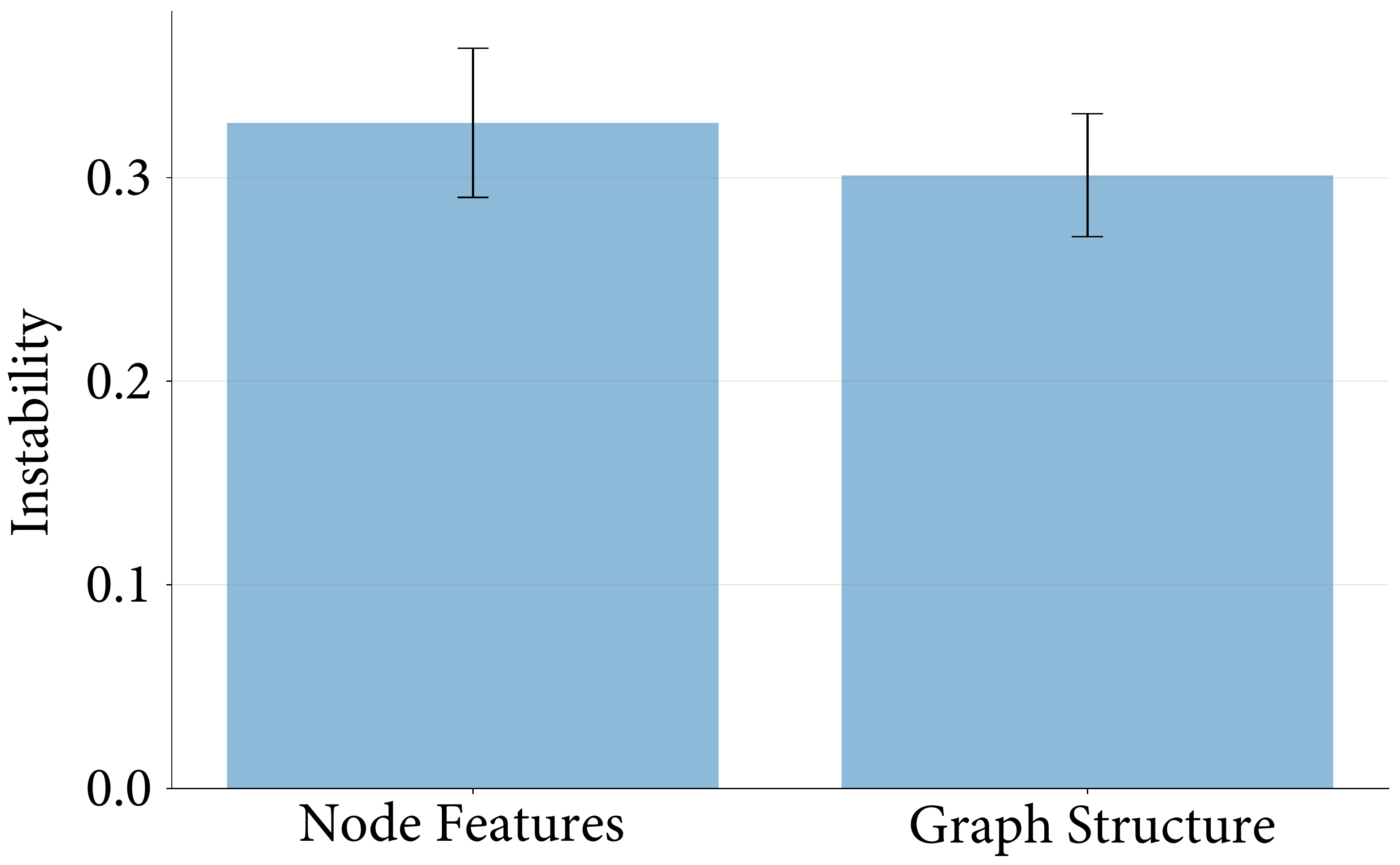}
    \end{subfigure}
    \begin{subfigure}[b]{0.24\textwidth}
        \includegraphics[width=\linewidth]{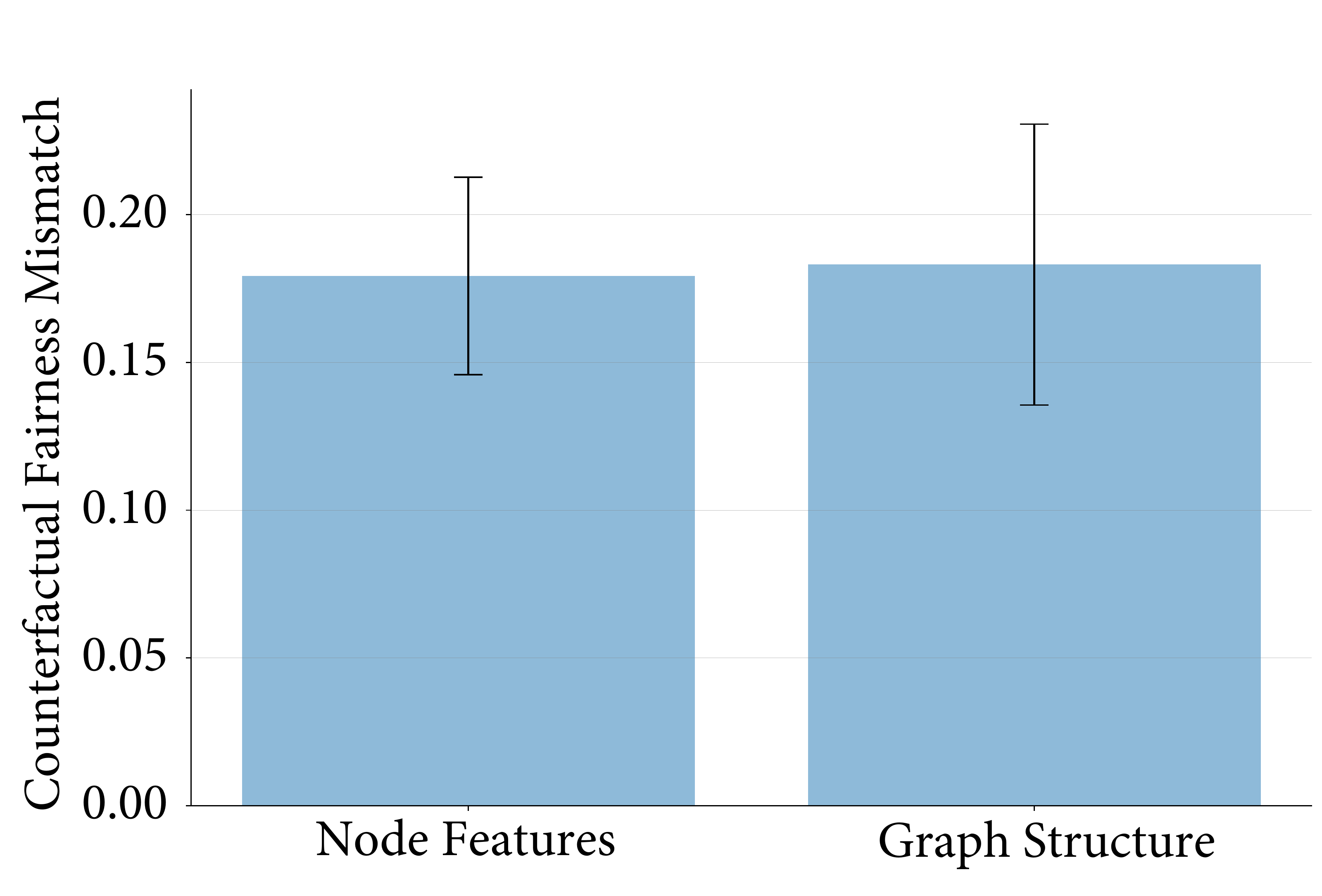}
    \end{subfigure}
    \begin{subfigure}[b]{0.24\textwidth}
        \includegraphics[width=\linewidth]{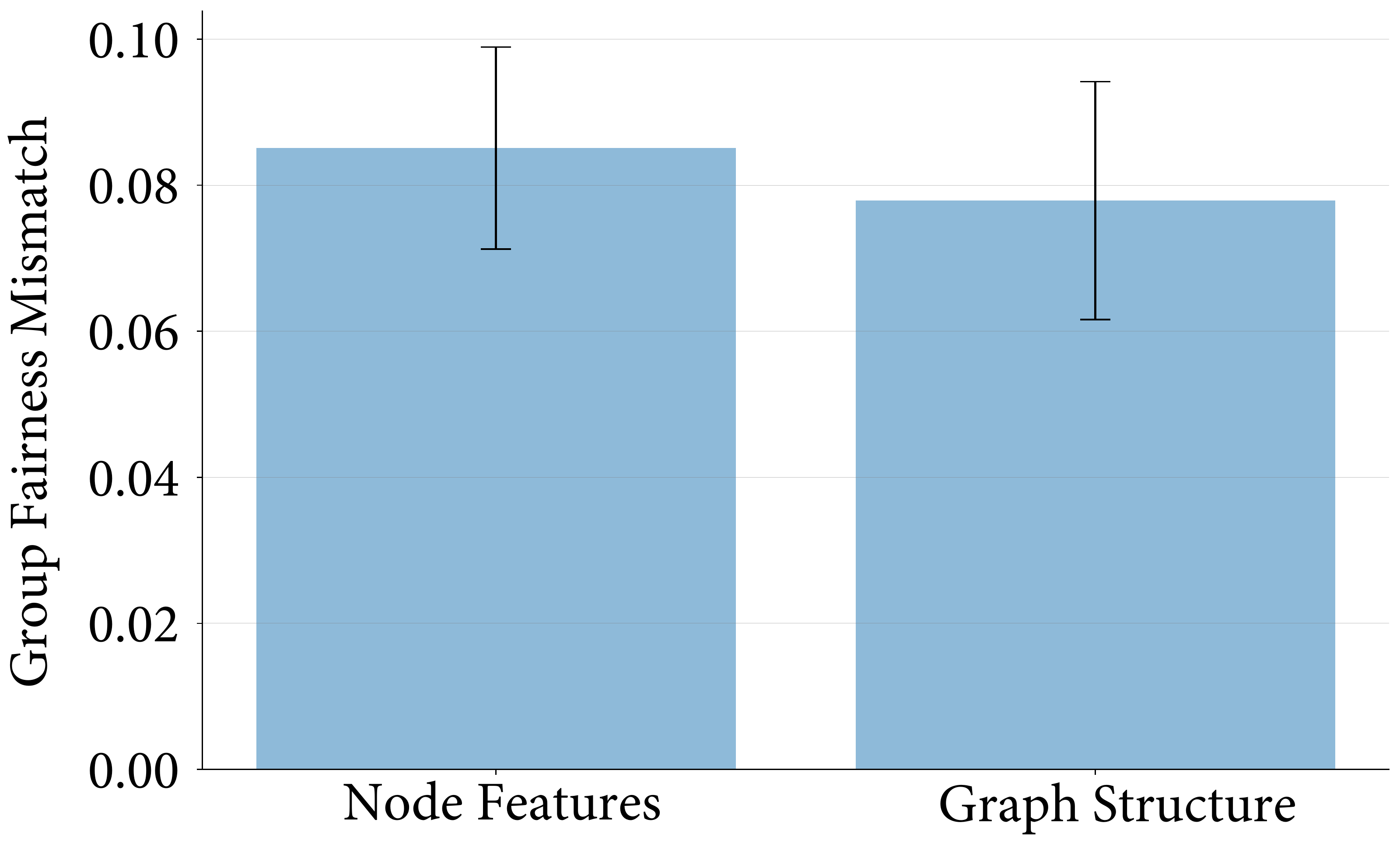}
    \end{subfigure}
    \caption{Shown are bar plots with mean values and standard errors of computed metrics for \expmethods generating node- (VanillaGrad, Integrated Gradients, GraphLIME, GNNExplainer) and graph-level (PGMExplainer, GraphMASK, PGExplainer) explanations across all datasets. We observe that graph structure-based explanations are more stable than node feature explanations, but performs on par in terms of other properties.}
    \label{app:fig_node_edge_compare}
\end{figure}
\begin{table}[ht]
\centering\small
\caption{
\looseness=-1Systematic evaluation of GNN \expmethods (random strategies (in grey), gradient-based methods (in yellow), surrogate-based methods (in purple), and perturbation-based methods (in red)) for node classification tasks. Shown are average values of metrics and standard errors across all nodes in the test set. Arrows ($\downarrow$) indicate the direction of better performance. Note that fairness does not apply to some datasets (i.e., N/A) as they do not contain sensitive attributes.
}
\label{app:tab_node_metric_1}
\setlength{\tabcolsep}{2.7pt}
\vspace{-1mm}
{\begin{tabular}{cl|cccc}
 & & \multicolumn{4}{c}{Evaluation metrics} \\
\multirow{2}{*}{Dataset} & \multirow{2}{*}{Method} & \multirow{2}{*}{Unfaithfulness ($\downarrow$)} & \multirow{2}{*}{Instability ($\downarrow$)} & \multicolumn{2}{c}{Fairness Mismatch ($\downarrow$)} \\
 & & &  & \makecell{Counterfactual} & \makecell{Group} \\
\toprule
\multirow{1}{2cm}{German\\credit graph} &
\begin{tabular}[l]{@{}l@{}}\ccg{Random Node Features}\\\ccg{Random Edges}\\\ccy{{VanillaGrad}}\\\ccy{Integrated Gradients}\\\ccb{{GraphLIME}}\\\ccb{{PGMExplainer}}\\\ccr{{GraphMASK}}\\\ccr{GNNExplainer}\\\ccr{PGExplainer}\end{tabular}&  \begin{tabular}[c]{@{}c@{}}{{0.208}\std{0.011}}\\{{0.049}\std{0.004}}\\{{0.185}\std{0.010}}\\{{0.199}\std{0.011}}\\{{0.158}\std{0.009}}\\{{0.131}\std{0.007}}\\{{0.034}\std{0.003}}\\{{0.046}\std{0.004}}\\{{0.074}\std{0.006}}\end{tabular} & 
\begin{tabular}[c]{@{}c@{}}{{0.386}\std{0.006}}\\{0.375\std{0.001}}\\{{0.222}\std{0.010}}\\{{0.254}\std{0.019}}\\{{0.096}\std{0.013}}\\{{0.183}\std{0.006}}\\{{0.270}\std{0.008}}\\{{0.377}\std{0.001}}\\{{0.367}\std{0.004}}\end{tabular} & 
\begin{tabular}[c]{@{}c@{}}{{0.387}\std{0.006}}\\{0.375\std{0.001}}\\{{0.137}\std{0.007}}\\{{0.210}\std{0.018}}\\{{0.063}\std{0.008}}\\{{0.185}\std{0.006}}\\{{0.006}\std{0.001}}\\{{0.359}\std{0.002}}\\{{0.360}\std{0.009}}\end{tabular} & 
\begin{tabular}[c]{@{}c@{}}{{0.165}\std{0.015}}\\{0.061\std{0.009}}\\{{0.154}\std{0.012}}\\{{0.150}\std{0.012}}\\{{0.114}\std{0.010}}\\{{0.129}\std{0.010}}\\{{0.046}\std{0.006}}\\{{0.060}\std{0.009}}\\{{0.079}\std{0.001}}\end{tabular} \\
\midrule
\multirow{1}{2cm}{Recidivism graph} &
\begin{tabular}[l]{@{}l@{}}\ccg{Random Node Features}\\\ccg{Random Edges}\\\ccy{{VanillaGrad}}\\\ccy{Integrated Gradients}\\\ccb{{GraphLIME}}\\\ccb{{PGMExplainer}}\\\ccr{{GraphMASK}}\\\ccr{GNNExplainer}\\\ccr{PGExplainer}\end{tabular}& \begin{tabular}[c]{@{}c@{}}{{0.312}\std{0.004}}\\{{0.040}\std{0.001}}\\{{0.233}\std{0.004}}\\{{0.308}\std{0.005}}\\{{0.191}\std{0.004}}\\{{0.128}\std{0.001}}\\{{0.053}\std{0.002}}\\{{0.042}\std{0.001}}\\{{0.056}\std{0.001}}\end{tabular} &
\begin{tabular}[c]{@{}c@{}}{{0.403}\std{0.002}}\\{0.376\std{0.000}}\\{{0.285}\std{0.003}}\\{{0.226}\std{0.003}}\\{{0.264}\std{0.004}}\\{{0.226}\std{0.002}}\\{{0.251}\std{0.003}}\\{{0.374}\std{0.000}}\\{{0.371}\std{0.001}}\end{tabular} & 
\begin{tabular}[c]{@{}c@{}}{{0.403}\std{0.002}}\\{0.376\std{0.000}}\\{{0.173}\std{0.002}}\\{{0.104}\std{0.003}}\\{{0.072}\std{0.003}}\\{{0.223}\std{0.002}}\\{{0.013}\std{0.000}}\\{{0.364}\std{0.001}}\\{{0.355}\std{0.002}}\end{tabular} &
\begin{tabular}[c]{@{}c@{}}{{0.144}\std{0.003}}\\{0.046\std{0.001}}\\{{0.114}\std{0.002}}\\{{0.139}\std{0.003}}\\{{0.107}\std{0.003}}\\{{0.130}\std{0.002}}\\{{0.060}\std{0.002}}\\{{0.051}\std{0.002}}\\{{0.064}\std{0.002}}\end{tabular} \\
\midrule
\multirow{1}{2cm}{Credit defaulter graph} & \begin{tabular}[l]{@{}l@{}}\ccg{Random Node Features}\\\ccg{Random Edges}\\\ccy{{VanillaGrad}}\\\ccy{Integrated Gradients}\\\ccb{{GraphLIME}}\\\ccb{{PGMExplainer}}\\\ccr{{GraphMASK}}\\\ccr{GNNExplainer}\\\ccr{PGExplainer}\end{tabular} &
\begin{tabular}[c]{@{}c@{}}{{0.098}\std{0.002}}\\{{0.020}\std{0.001}}\\{{0.092}\std{0.002}}\\{{0.147}\std{0.003}}\\{{0.038}\std{0.002}}\\{{0.283}\std{0.002}}\\{{0.012}\std{0.001}}\\{{0.021}\std{0.001}}\\{{0.028}\std{0.001}}\end{tabular} &
\begin{tabular}[c]{@{}c@{}}{{0.426}\std{0.002}}\\{{0.376}\std{0.000}}\\{{0.333}\std{0.002}}\\{{0.140}\std{0.002}}\\{{0.225}\std{0.004}}\\{{0.156}\std{0.002}}\\{{0.036}\std{0.002}}\\{{0.375}\std{0.000}}\\{{0.364}\std{0.001}}\end{tabular} &
\begin{tabular}[c]{@{}c@{}}{{0.424}\std{0.002}}\\{0.376\std{0.000}}\\{{0.171}\std{0.002}}\\{{0.069}\std{0.001}}\\{{0.063}\std{0.003}}\\{{0.154}\std{0.002}}\\{{0.004}\std{0.000}}\\{{0.366}\std{0.000}}\\{{0.348}\std{0.002}}\end{tabular} & 
\begin{tabular}[c]{@{}c@{}}{{0.045}\std{0.002}}\\{0.017\std{0.001}}\\{{0.042}\std{0.002}}\\{{0.053}\std{0.002}}\\{{0.018}\std{0.001}}\\{{0.161}\std{0.003}}\\{{0.010}\std{0.001}}\\{{0.019}\std{0.001}}\\{{0.022}\std{0.001}}\end{tabular} \\
\bottomrule
\end{tabular}}
\vskip -0.1in
\end{table}

\begin{table}[ht]
\centering\small
\caption{
\looseness=-1Systematic evaluation of GNN \expmethods (random strategies (in grey), gradient-based methods (in yellow), surrogate-based methods (in purple), and perturbation-based methods (in red)) for node classification tasks. Shown are average values of metrics and standard errors across all nodes in the test set. Arrows ($\downarrow$) indicate the direction of better performance. Note that fairness does not apply to some datasets (i.e., N/A) as they do not contain sensitive attributes.
}
\label{app:tab_node_metric_2}
\vspace{-1mm}
{\begin{tabular}{cl|cccc}
 & & \multicolumn{4}{c}{Evaluation metrics} \\
\multirow{2}{*}{Dataset} & \multirow{2}{*}{Method} & \multirow{2}{*}{Unfaithfulness ($\downarrow$)} & \multirow{2}{*}{Instability ($\downarrow$)} & \multicolumn{2}{c}{Fairness Mismatch ($\downarrow$)} \\
 & & &  & \makecell{Counterfactual} & \makecell{Group} \\
\toprule
\multirow{1}{2cm}{Cora} &
\begin{tabular}[l]{@{}l@{}}\ccg{Random Node Features}\\\ccg{Random Edges}\\\ccy{{VanillaGrad}}\\\ccy{Integrated Gradients}\\\ccb{{GraphLIME}}\\\ccb{{PGMExplainer}}\\\ccr{{GraphMASK}}\\\ccr{GNNExplainer}\\\ccr{PGExplainer}\end{tabular}&
\begin{tabular}[c]{@{}c@{}}{{0.002}\std{0.000}}\\{{0.004}\std{0.000}}\\{{0.002}\std{0.000}}\\{{0.002}\std{0.000}}\\{{0.001}\std{0.001}}\\{{0.016}\std{0.001}}\\{{0.023}\std{0.005}}\\{{0.003}\std{0.000}}\\{{0.112}\std{0.005}}\end{tabular} &
\begin{tabular}[c]{@{}c@{}}{{0.181}\std{0.000}}\\{0.196\std{0.006}}\\{{0.154}\std{0.002}}\\{{0.894}\std{0.002}}\\{{0.052}\std{0.015}}\\{{0.224}\std{0.011}}\\{{0.600}\std{0.027}}\\{{0.377}\std{0.009}}\\{{0.372}\std{0.008}}\end{tabular} & 
\begin{tabular}[c]{@{}c@{}}{\na}\\{\na}\\{\na}\\{\na}\\{\na}\\{\na}\\{\na}\\{\na}\\{\na}\end{tabular} &
\begin{tabular}[c]{@{}c@{}}{\na}\\{\na}\\{\na}\\{\na}\\{\na}\\{\na}\\{\na}\\{\na}\\{\na}\end{tabular} \\
\midrule
\multirow{1}{2cm}{PubMed} &
\begin{tabular}[l]{@{}l@{}}\ccg{Random Node Features}\\\ccg{Random Edges}\\\ccy{{VanillaGrad}}\\\ccy{Integrated Gradients}\\\ccb{{GraphLIME}}\\\ccb{{PGMExplainer}}\\\ccr{{GraphMASK}}\\\ccr{GNNExplainer}\\\ccr{PGExplainer}\end{tabular}&
\begin{tabular}[c]{@{}c@{}}{{0.002}\std{0.000}}\\{{0.002}\std{0.000}}\\{{0.003}\std{0.000}}\\{{0.004}\std{0.000}}\\{{0.001}\std{0.001}}\\{{0.045}\std{0.003}}\\{{0.010}\std{0.002}}\\{{0.002}\std{0.000}}\\{{0.094}\std{0.004}}\end{tabular} &
\begin{tabular}[c]{@{}c@{}}{{0.180}\std{0.000}}\\{0.195\std{0.002}}\\{{0.139}\std{0.002}}\\{{0.855}\std{0.003}}\\{{0.440}\std{0.023}}\\{{0.142}\std{0.008}}\\{{0.742}\std{0.018}}\\{{0.192}\std{0.002}}\\{{0.367}\std{0.006}}\end{tabular} & 
\begin{tabular}[c]{@{}c@{}}{\na}\\{\na}\\{\na}\\{\na}\\{\na}\\{\na}\\{\na}\\{\na}\\{\na}\end{tabular} &
\begin{tabular}[c]{@{}c@{}}{\na}\\{\na}\\{\na}\\{\na}\\{\na}\\{\na}\\{\na}\\{\na}\\{\na}\end{tabular} \\
\midrule
\multirow{1}{2cm}{Citeseer} &
\begin{tabular}[l]{@{}l@{}}\ccg{Random Node Features}\\\ccg{Random Edges}\\\ccy{{VanillaGrad}}\\\ccy{Integrated Gradients}\\\ccb{{GraphLIME}}\\\ccb{{PGMExplainer}}\\\ccr{{GraphMASK}}\\\ccr{GNNExplainer}\\\ccr{PGExplainer}\end{tabular}&
\begin{tabular}[c]{@{}c@{}}{{0.003}\std{0.000}}\\{{0.005}\std{0.000}}\\{{0.003}\std{0.000}}\\{{0.004}\std{0.000}}\\{{0.001}\std{0.000}}\\{{0.009}\std{0.001}}\\{{0.170}\std{0.007}}\\{{0.003}\std{0.000}}\\{{0.129}\std{0.010}}\end{tabular} &
\begin{tabular}[c]{@{}c@{}}{{0.180}\std{0.000}}\\{0.263\std{0.019}}\\{{0.142}\std{0.002}}\\{{0.896}\std{0.000}}\\{{0.048}\std{0.015}}\\{{0.262}\std{0.015}}\\{{0.200}\std{0.028}}\\{{0.212}\std{0.015}}\\{{0.400}\std{0.017}}\end{tabular} & 
\begin{tabular}[c]{@{}c@{}}{\na}\\{\na}\\{\na}\\{\na}\\{\na}\\{\na}\\{\na}\\{\na}\\{\na}\end{tabular} &
\begin{tabular}[c]{@{}c@{}}{\na}\\{\na}\\{\na}\\{\na}\\{\na}\\{\na}\\{\na}\\{\na}\\{\na}\end{tabular} \\
\midrule
\multirow{1}{2cm}{Ogbn-mag} &
\begin{tabular}[l]{@{}l@{}}\ccg{Random Node Features}\\\ccg{Random Edges}\\\ccy{{VanillaGrad}}\\\ccy{Integrated Gradients}\\\ccb{{GraphLIME}}\\\ccb{{PGMExplainer}}\\\ccr{{GraphMASK}}\\\ccr{GNNExplainer}\\\ccr{PGExplainer}\end{tabular}&
\begin{tabular}[c]{@{}c@{}}{{0.002}\std{0.000}}\\{{0.002}\std{0.000}}\\{{0.002}\std{0.000}}\\{{0.002}\std{0.000}}\\{{0.001}\std{0.000}}\\{{0.002}\std{0.000}}\\{{0.002}\std{0.000}}\\{{0.002}\std{0.000}}\\{{0.002}\std{0.000}}\end{tabular} &
\begin{tabular}[c]{@{}c@{}}{{0.373}\std{0.002}}\\{0.376\std{0.002}}\\{{0.312}\std{0.005}}\\{{0.368}\std{0.002}}\\{{0.354}\std{0.023}}\\{{0.222}\std{0.006}}\\{{0.323}\std{0.003}}\\{{0.375}\std{0.002}}\\{{0.375}\std{0.006}}\end{tabular} & 
\begin{tabular}[c]{@{}c@{}}{\na}\\{\na}\\{\na}\\{\na}\\{\na}\\{\na}\\{\na}\\{\na}\\{\na}\end{tabular} &
\begin{tabular}[c]{@{}c@{}}{\na}\\{\na}\\{\na}\\{\na}\\{\na}\\{\na}\\{\na}\\{\na}\\{\na}\end{tabular} \\
\midrule
\multirow{1}{2cm}{Ogbn-arxiv} & \begin{tabular}[l]{@{}l@{}}\ccg{Random Node Features}\\\ccg{Random Edges}\\\ccy{{VanillaGrad}}\\\ccy{Integrated Gradients}\\\ccb{{GraphLIME}}\\\ccb{{PGMExplainer}}\\\ccr{{GraphMASK}}\\\ccr{GNNExplainer}\\\ccr{PGExplainer}\end{tabular} &
\begin{tabular}[c]{@{}c@{}}{{0.529}\std{0.002}}\\{{0.431}\std{0.002}}\\{{0.528}\std{0.002}}\\{{0.528}\std{0.002}}\\{{0.260}\std{0.003}}\\{{0.413}\std{0.002}}\\{{0.586}\std{0.001}}\\{{0.430}\std{0.002}}\\{{0.338}\std{0.002}}\end{tabular} &
\begin{tabular}[c]{@{}c@{}}{{0.375}\std{0.000}}\\{{0.378}\std{0.001}}\\{{0.359}\std{0.001}}\\{{0.372}\std{0.000}}\\{{0.374}\std{0.004}}\\{{0.270}\std{0.002}}\\{{0.125}\std{0.002}}\\{{0.376}\std{0.001}}\\{{0.381}\std{0.001}}\end{tabular} &
\begin{tabular}[c]{@{}c@{}}{\na}\\{\na}\\{\na}\\{\na}\\{\na}\\{\na}\\{\na}\\{\na}\\{\na}\end{tabular} & 
\begin{tabular}[c]{@{}c@{}}{\na}\\{\na}\\{\na}\\{\na}\\{\na}\\{\na}\\{\na}\\{\na}\\{\na}\end{tabular} \\
\bottomrule
\end{tabular}}
\vskip -0.1in
\end{table}

\begin{table}[t]
\centering\small
\caption{Systematic evaluation of GNN \expmethods (random strategies (in grey), gradient- (in yellow), perturbation- (in red), and surrogate-based (in purple) method) for link prediction on Cora-link and graph classification on MUTAG dataset. For both tasks, VanillaGrad method perform on par or better than other \expmethods on instability, and random strategies outperform other GNN \expmethods on unfaithfulness.}
% \vspace{-2.5mm}
\label{tab:link-graph-metric}
{\begin{tabular}{cl|cc}
Task & Method & Unfaithfulness ($\downarrow$) & Instability ($\downarrow$) \\
\toprule
\multirow{1}{2.5cm}{Link prediction} & \begin{tabular}[l]{@{}l@{}}\ccg{Random Node Features}\\\ccy{{VanillaGrad}}\\\ccy{Integrated Gradients}\\\ccr{GNNExplainer}\end{tabular} &
\begin{tabular}[c]{@{}c@{}}{{0.037}\std{0.013}}\\{{0.046}\std{0.017}}\\{{0.069}\std{0.028}}\\{{0.040}\std{0.017}}\end{tabular} &
\begin{tabular}[c]{@{}c@{}}{{0.375}\std{0.003}}\\{{0.310}\std{0.062}}\\{{0.747}\std{0.002}}\\{{0.376}\std{0.000}}\end{tabular} \\
\midrule
\multirow{1}{2.5cm}{Graph classification} & \begin{tabular}[l]{@{}l@{}}\ccg{Random Node Features}\\\ccg{Random Edges}\\\ccy{{VanillaGrad}}\\\ccy{Integrated Gradients}\\\ccb{{PGMExplainer}}\\\ccr{GNNExplainer}\end{tabular} &
\begin{tabular}[c]{@{}c@{}}{{0.105}\std{0.056}}\\{{0.022}\std{0.029}}\\{{0.295}\std{0.078}}\\{{0.086}\std{0.046}}\\{{0.154}\std{0.083}}\\{{0.094}\std{0.052}}\end{tabular} &
\begin{tabular}[c]{@{}c@{}}{{0.492}\std{0.046}}\\{{0.366}\std{0.064}}\\{{0.363}\std{0.129}}\\{{0.473}\std{0.087}}\\{{0.385}\std{0.118}}\\{{0.490}\std{0.046}}\end{tabular} \\
\bottomrule
\end{tabular}}
\vskip -0.15in
\end{table}

\begin{figure}[ht]
    \centering
    \begin{subfigure}[b]{0.45\textwidth}
        \includegraphics[width=\linewidth]{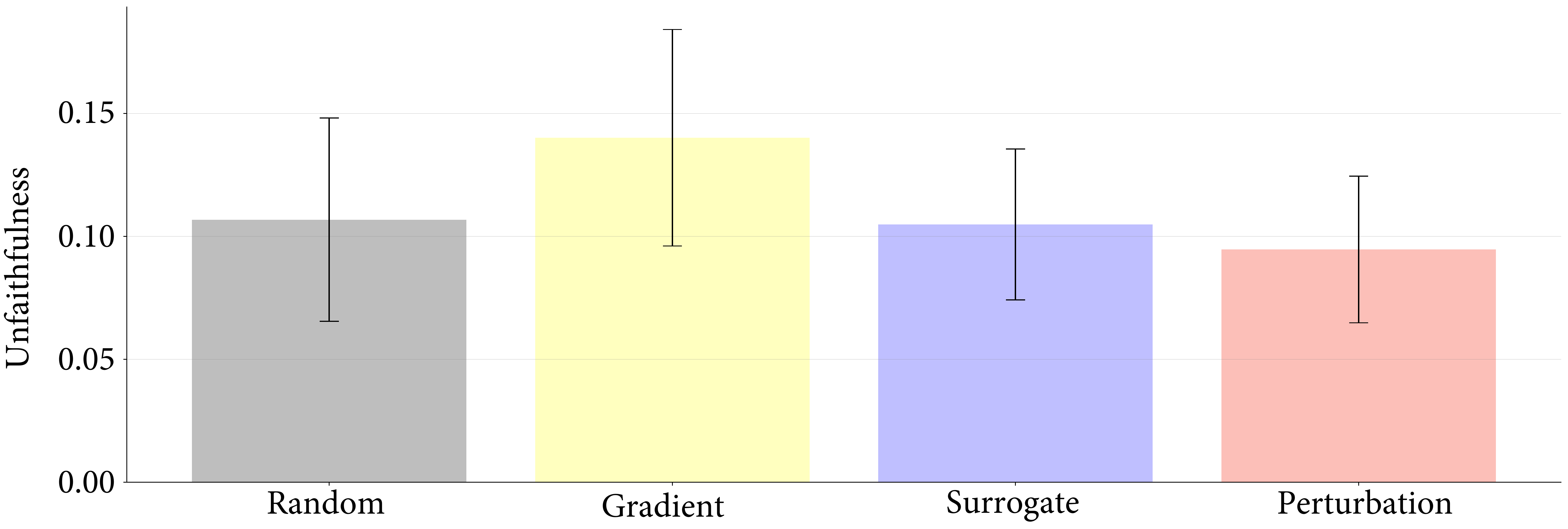}
    \end{subfigure}
    \begin{subfigure}[b]{0.45\textwidth}
        \includegraphics[width=\linewidth]{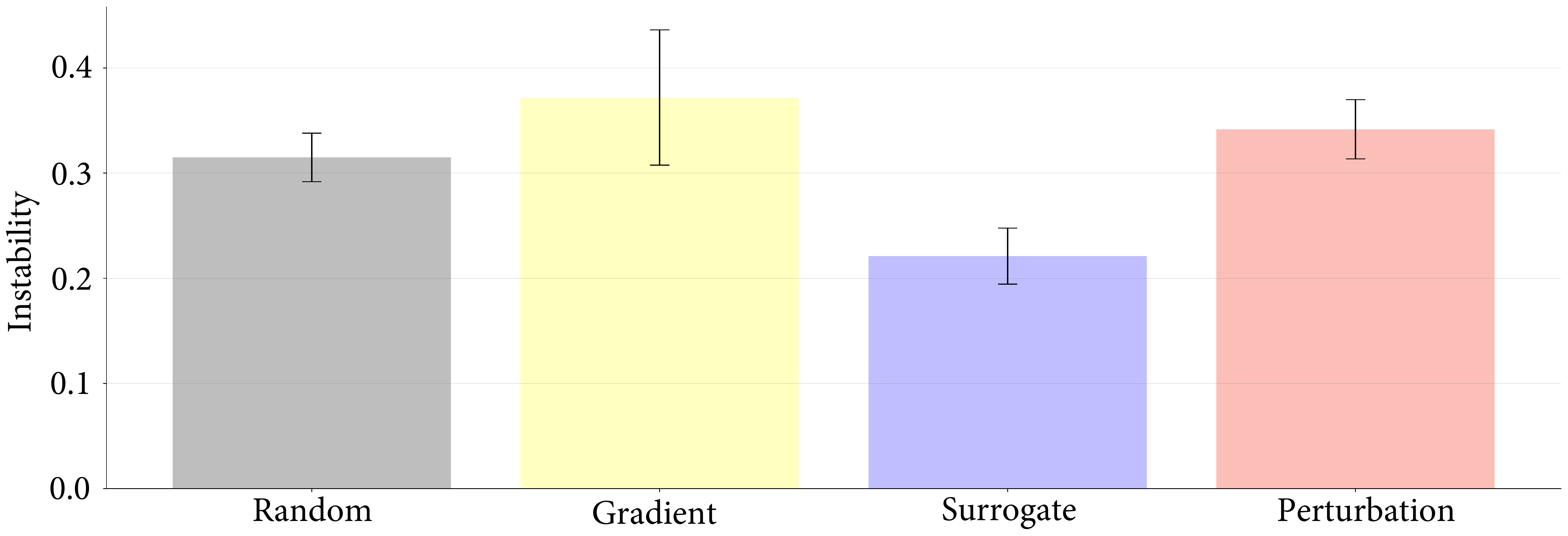}
    \end{subfigure}\\
    \begin{subfigure}[b]{0.45\textwidth}
        \includegraphics[width=\linewidth]{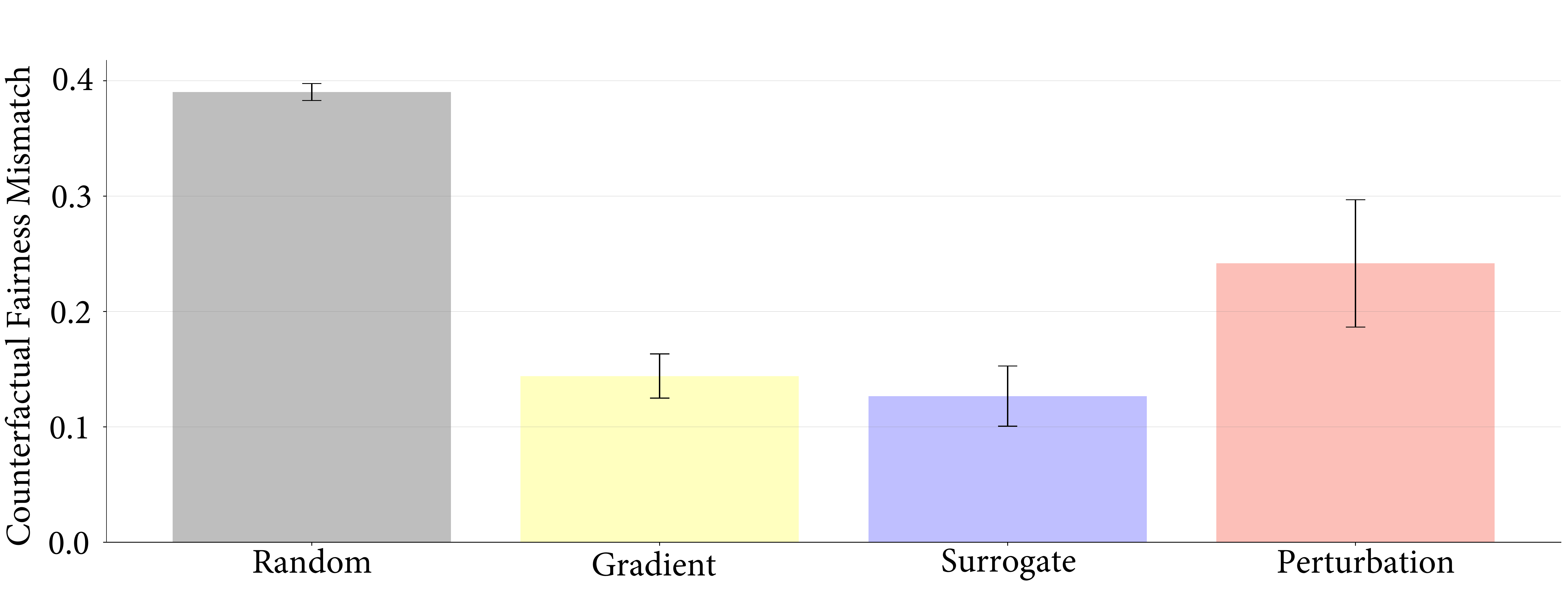}
    \end{subfigure}
    \begin{subfigure}[b]{0.45\textwidth}
        \includegraphics[width=\linewidth]{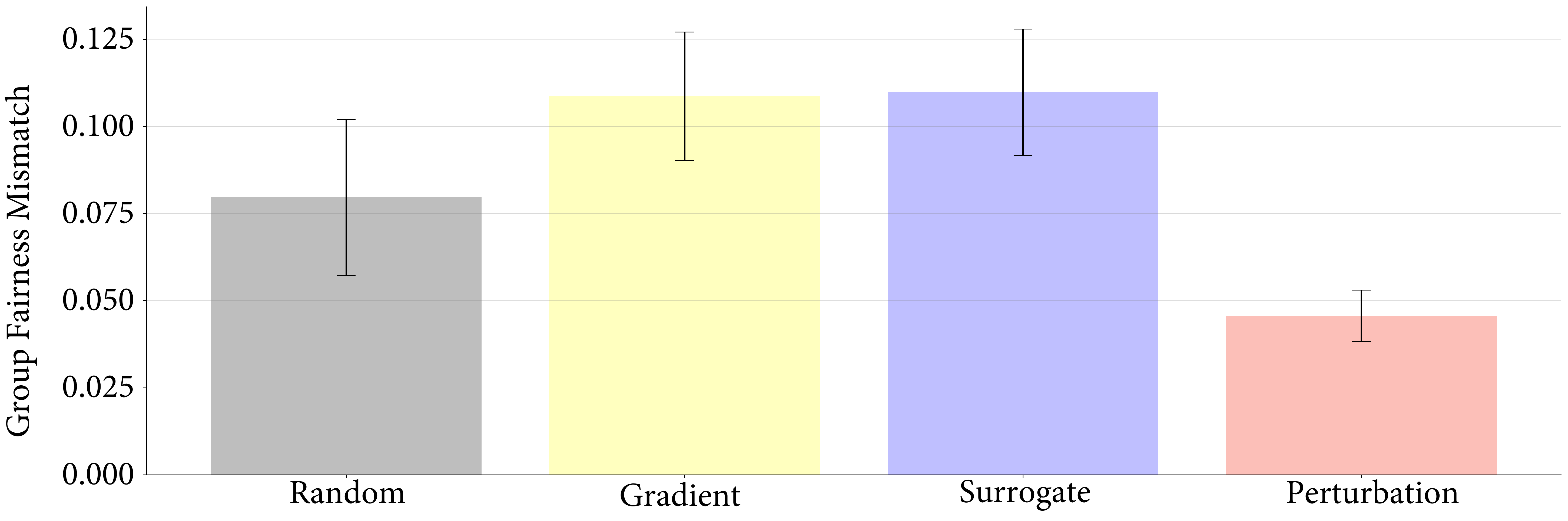}
    \end{subfigure}    
    \caption{Shown are bar plots with mean values and standard errors of computed metrics for four categories of \expmethods (random strategies (in grey), gradient-based methods (in yellow), surrogate-based methods (in purple), and perturbation-based methods (in red)) across all datasets. We observe that surrogate-based \expmethods are more stable and better preserves counterfactual fairness, whereas perturbation-based \expmethods outperform others on unfaithfulness and group fairness mismatch scores.}
    \label{app:fig_category_compare}
\end{figure}
\begin{figure}[ht]
    \centering
    \begin{subfigure}[b]{0.81\textwidth}
        \includegraphics[width=\linewidth]{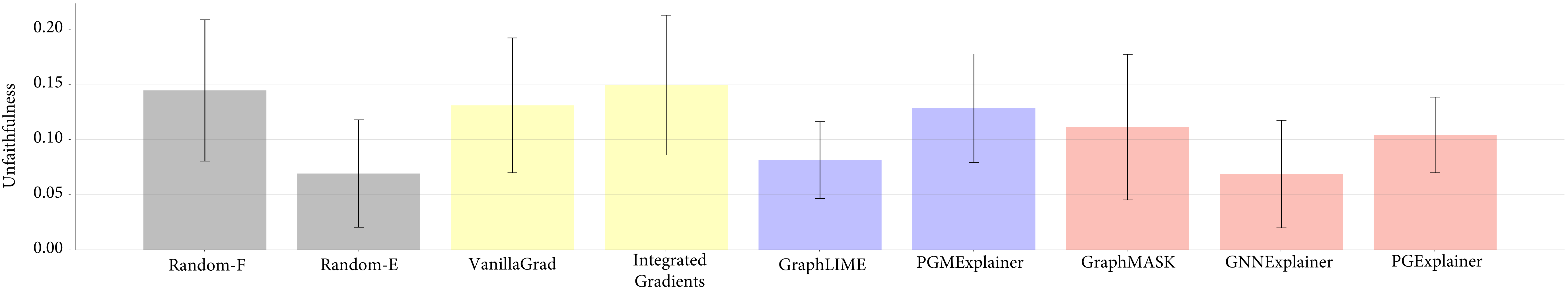}
        % \label{app:fig_stab_count_grad_bound}
    \end{subfigure}
    \begin{subfigure}[b]{0.81\textwidth}
        \includegraphics[width=\linewidth]{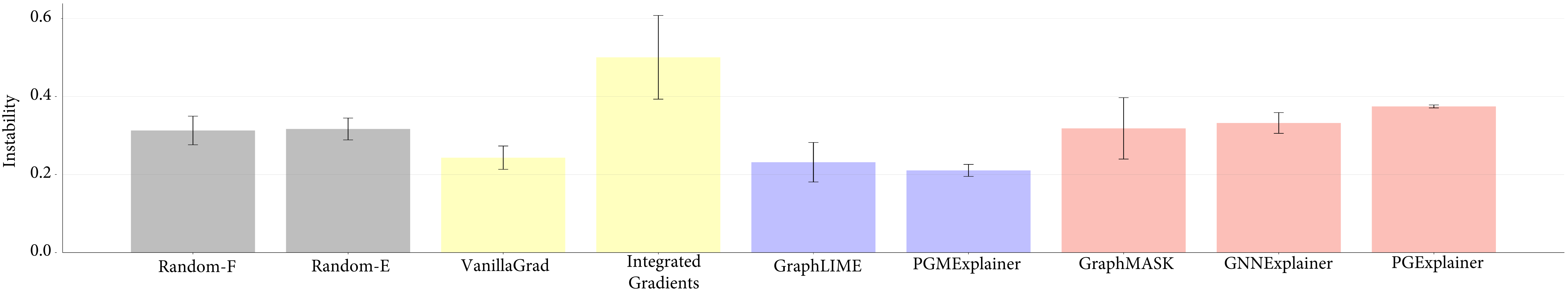}
        % \caption{GraphLIME}
        % \label{app:fig_stab_count_graphlime_bound}
    \end{subfigure}
    \begin{subfigure}[b]{0.81\textwidth}
        \includegraphics[width=\linewidth]{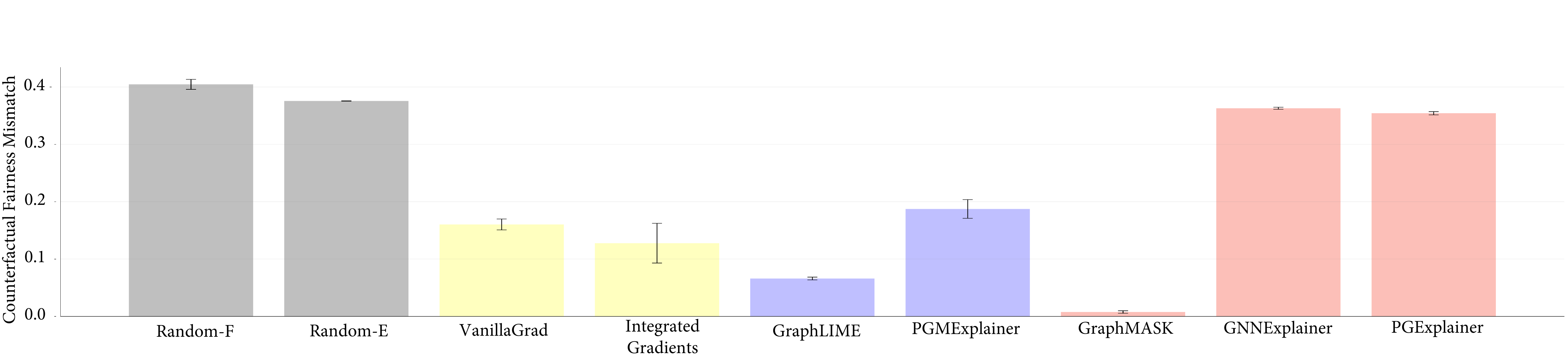}
        % \caption{GraphMASK}
        % \label{app:fig_stab_count_graphmask_bound}
    \end{subfigure}
    \begin{subfigure}[b]{0.81\textwidth}
        \includegraphics[width=\linewidth]{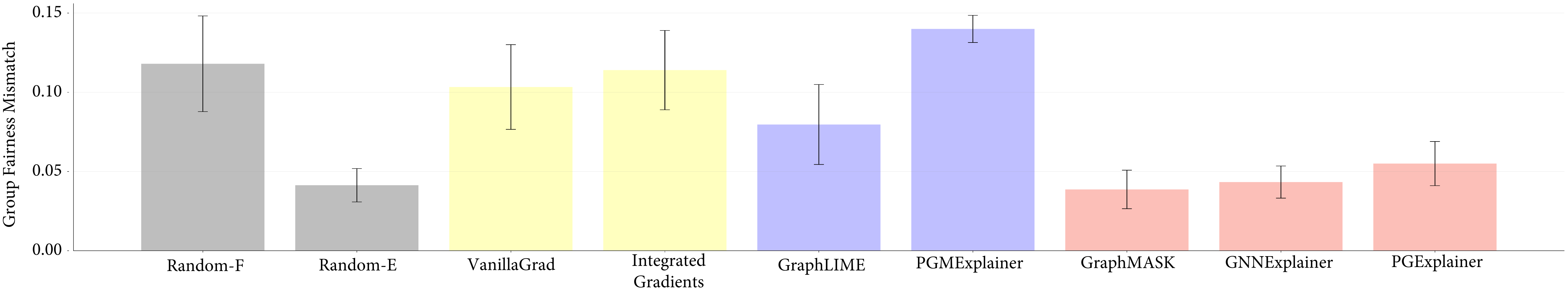}
        % \caption{GraphMASK}
        % \label{app:fig_stab_count_graphmask_bound}
    \end{subfigure}    
    \caption{Shown are bar plots with mean values and standard errors of computed metrics for nine GNN \expmethods (random strategies (in grey), gradient-based methods (in yellow), surrogate-based methods (in purple), and perturbation-based methods (in red)) across all datasets. We observe that Random Edge baseline outperforms all \expmethods in terms of unfaithfulness, GraphMASK outperforms all \expmethods in preserving counterfactual and group fairness, and no \expmethod satisfies all four properties.}
    \label{app:fig_method_compare}
\end{figure}
% \begin{figure}
%     \centering
%     \includegraphics[width=0.45\textwidth]{iclr2022/FIG/node_edge_wo_random_compare.pdf}
%     \caption{Across all datasets, average and standard errors of gradient, surrogate, and perturbation-based methods generating node- and graph-level explanations. INSIGHTS: Graph structures achieve on-par or better instability, counterfactual and group fairness mismatch, but underperforms on unfaithfulness.}
%     \label{fig:my_label}
% \end{figure}
\begin{figure}[ht]
    \centering
    \begin{subfigure}[b]{0.42\textwidth}
        \includegraphics[width=\linewidth]{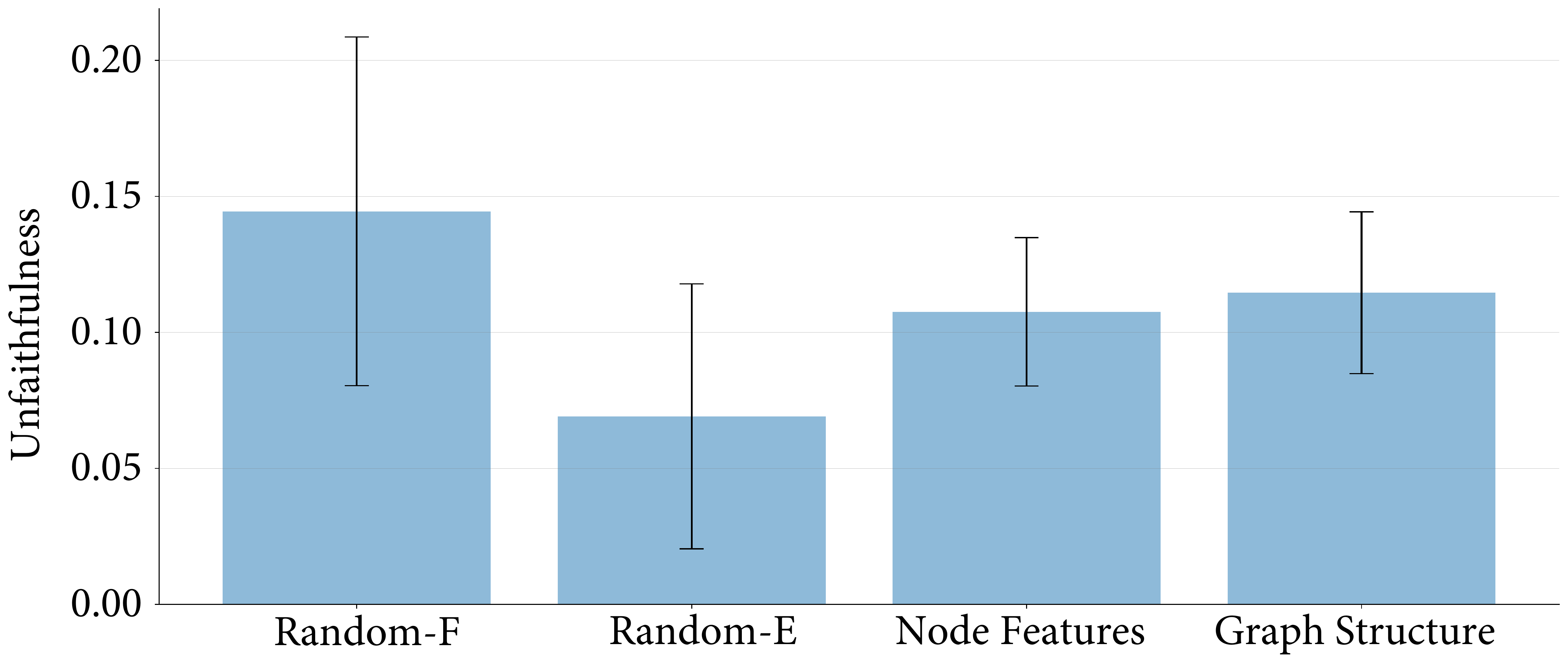}
    \end{subfigure}
    \begin{subfigure}[b]{0.42\textwidth}
        \includegraphics[width=\linewidth]{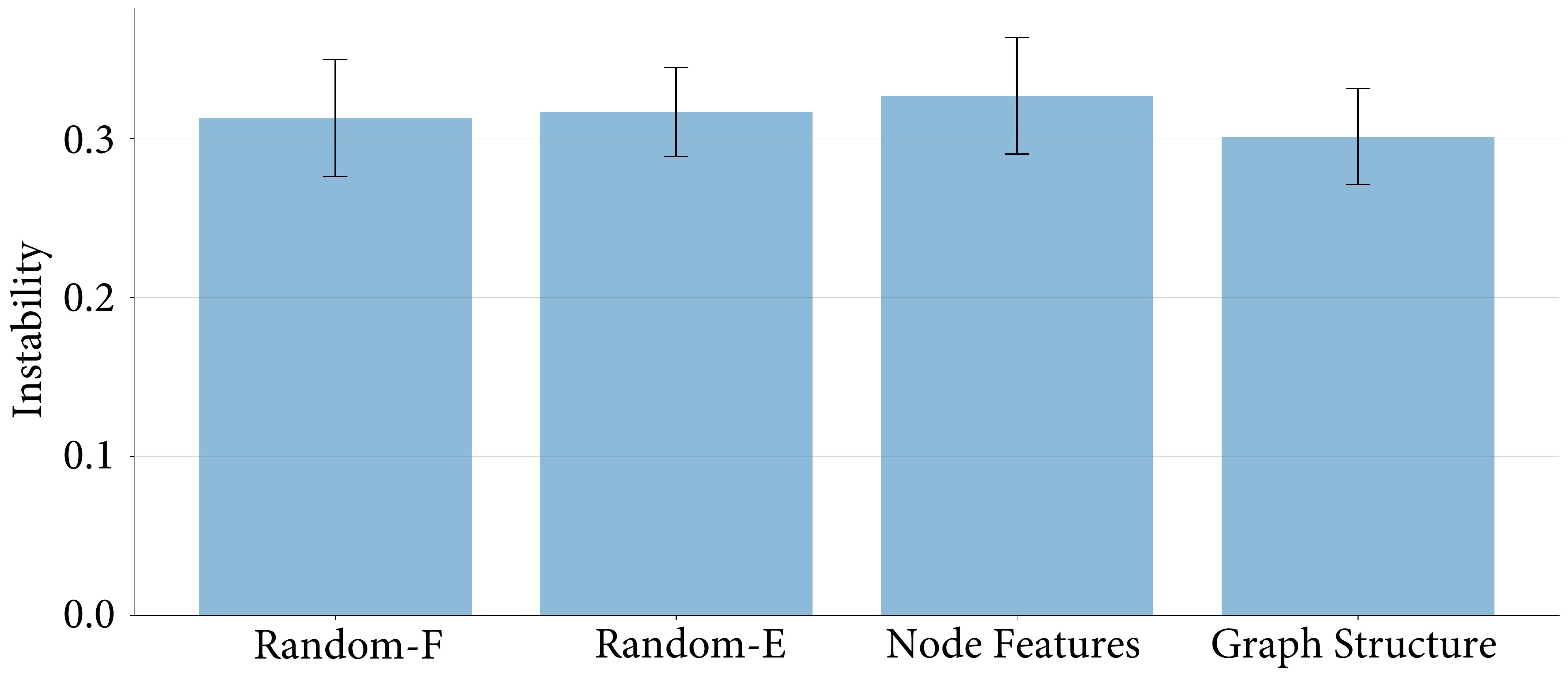}
    \end{subfigure}\\
    \begin{subfigure}[b]{0.42\textwidth}
        \includegraphics[width=\linewidth]{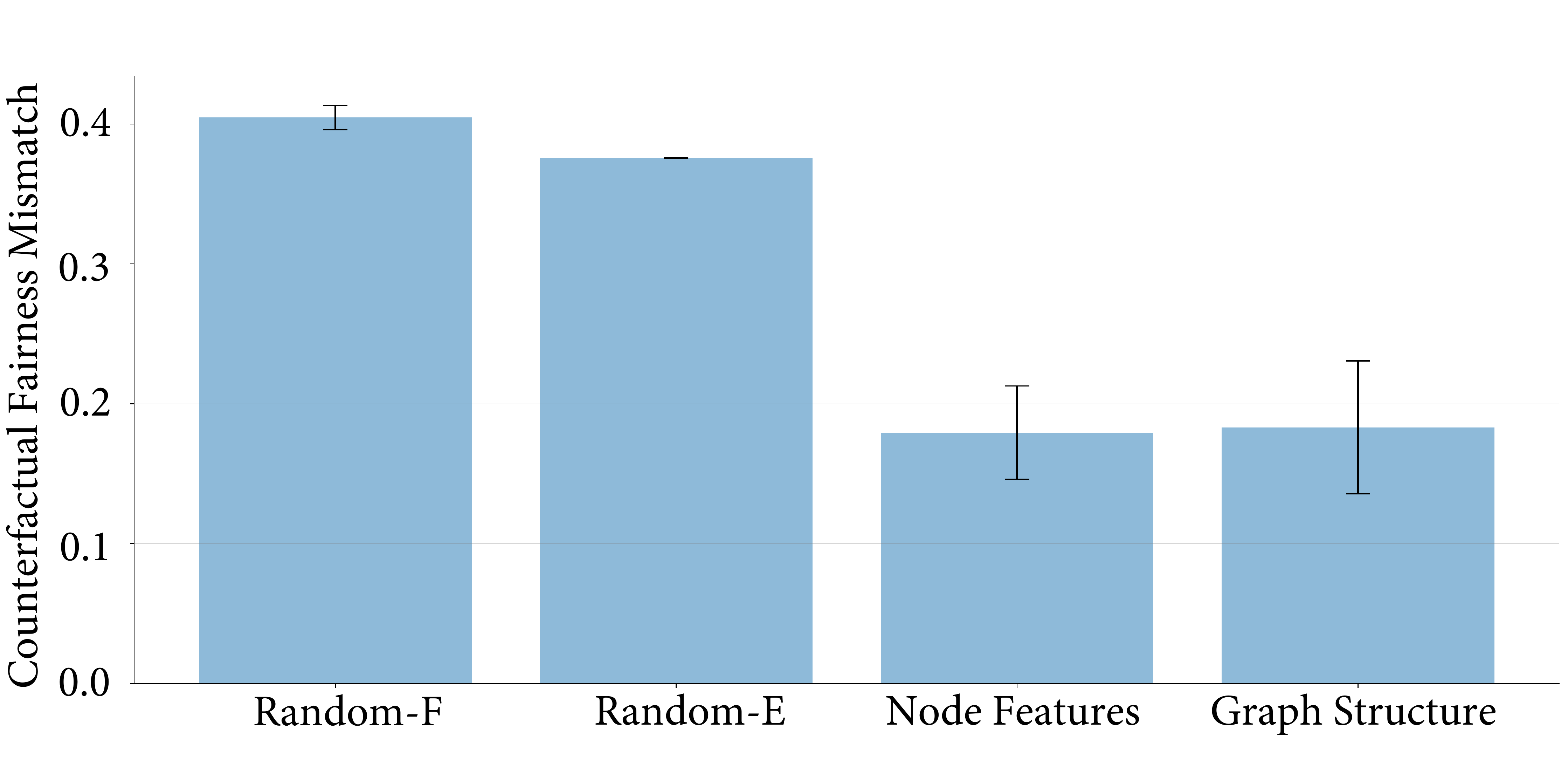}
    \end{subfigure}
    \begin{subfigure}[b]{0.42\textwidth}
        \includegraphics[width=\linewidth]{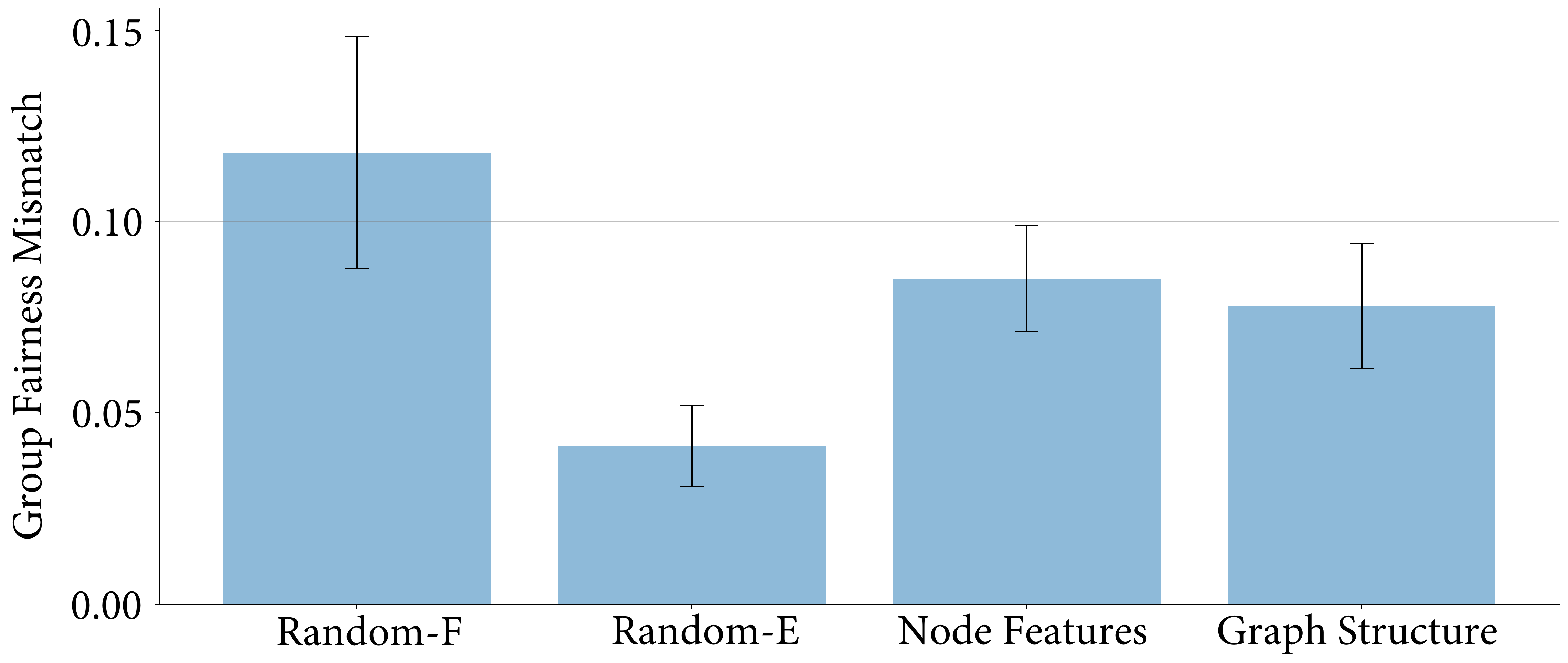}
    \end{subfigure}    
    \caption{Shown are bar plots with mean values and standard errors of computed metrics of random node feature explanations, random edge explanations, and \expmethods generating node- (VanillaGrad, Integrated Gradients, GraphLIME, GNNExplainer) and graph-level (PGMExplainer, GraphMASK, PGExplainer) explanations across all datasets. We observe that Random Edge explanations achieves better unfaithfulness and group fairness mismatch, whereas graph structure-based explanations perform better on stability and preserving counterfactual fairness.}
    \label{app:fig_randomf_e_node_edge_compare}
\end{figure}